\documentclass[10pt,twocolumn,letterpaper,final,pagenumbers]{article}

\usepackage[pagenumbers]{cvpr} 

\usepackage{makecell}
\usepackage{balance}
\usepackage{cuted}

\definecolor{cvprblue}{rgb}{0.21, 0.49, 0.74}
\usepackage[pagebackref,breaklinks,colorlinks,allcolors=cvprblue]{hyperref}
\usepackage{algorithm}
\usepackage{algorithmic}
\usepackage{amsthm}
\newtheorem{theorem}{Theorem}

\usepackage{multirow}

\usepackage{makecell} 
\def\confName{CVPR}
\def\confYear{2027}
\usepackage{graphicx}
\usepackage{xcolor}
\usepackage{graphicx} 

\usepackage{newtxtt} 

\newlength{\suppqualtilesize}

\newcommand{\suppqualsetup}{%
    \centering
    \small
    \setlength{\tabcolsep}{2pt}%
    \renewcommand{\arraystretch}{1}%
    \setlength{\suppqualtilesize}
        {\dimexpr(\textwidth-16pt)/4\relax}%
    \ifdim\suppqualtilesize>0.215\textheight
        \setlength{\suppqualtilesize}{0.215\textheight}%
    \fi
}

\newcommand{\suppqualimage}[1]{%
    \includegraphics[
        width=\suppqualtilesize,
        height=\suppqualtilesize,
        keepaspectratio
    ]{#1}%
}

\newcommand{\suppqualheader}{%
    \makebox[\suppqualtilesize]{\textbf{Original}} &
    \makebox[\suppqualtilesize]{\textbf{Monarch}} &
    \makebox[\suppqualtilesize]{\textbf{DiTFastAttn}} &
    \makebox[\suppqualtilesize]{\textbf{LoSpAttn}} 
    \\[4pt]
}

\newcommand{\suppqualrow}[2]{%
    \suppqualimage{images/supplementary_images/#1/original_pipe/#2.jpg} &
    \suppqualimage{images/supplementary_images/#1/monarch/#2.jpg} &
    \suppqualimage{images/supplementary_images/#1/dfa/#2.jpg} &
    \suppqualimage{images/supplementary_images/#1/losp_pipe/#2.jpg} 
    \\[3pt]
}

\newlength{\storypanel}
\newlength{\storylabelwidth}
\newlength{\storytile}
\newlength{\storynoteheight}

\newcommand{\storylabel}[1]{%
    \parbox[c][\storytile][c]{\storylabelwidth}{%
        \centering
        \rotatebox[origin=c]{90}{\shortstack[c]{#1}}%
    }%
}
\newcommand{\storyimage}[1]{%
    \parbox[c][\storytile][c]{\storytile}{%
        \centering
        \includegraphics[
            width=\storytile,
            height=\storytile,
            keepaspectratio
        ]{#1}%
    }%
}

\newcommand{\storynote}[1]{%
    \parbox[c][\storynoteheight][c]{\storytile}{%
        \centering #1%
    }%
}

\newcommand{\storymap}[1]{%
    \storyimage{images/plots/selected_attention_maps/#1}%
}

\title{LoSpAttn: Fast Training-Free Hybrid Attention for Vision and Diffusion Transformers}

\author{
\textbf{J Karthik Palaniappan \quad Akshay Kiran Jose \quad Anvitha Vinod\textsuperscript{*} \quad Pravin Nair} \\
Department of Electrical Engineering \\
Indian Institute of Technology Madras, Chennai, India \\
{\small \texttt{ee25s010@smail.iitm.ac.in, ee25m009@smail.iitm.ac.in,}} \\
{\small \texttt{anvithavinod24@gmail.com, pravinnair@ee.iitm.ac.in}}
}

\newcommand{\q}{\boldsymbol{q}}
\newcommand{\m}{\boldsymbol{m}}
\newcommand{\x}{\boldsymbol{x}}
\newcommand{\y}{\boldsymbol{y}}
\newcommand{\X}{\mathbf{X}}
\newcommand{\Q}{\mathbf{Q}}
\newcommand{\D}{\mathbf{D}}
\newcommand{\K}{\mathbf{K}}
\newcommand{\V}{\mathbf{V}}
\newcommand{\G}{\mathbf{G}}
\newcommand{\Z}{\mathbf{Z}}

\newcommand{\M}{\mathbf{M}}
\newcommand{\I}{\mathbf{I}}
\newcommand{\boldO}{\mathbf{O}}
\newcommand{\A}{\mathbf{A}}
\newcommand{\B}{\mathbf{B}}
\newcommand{\C}{\mathbf{C}}
\renewcommand{\S}{\mathbf{S}}

\def\x{{\mathbf x}}

\begin{document}
\maketitle

\begingroup
\renewcommand{\thefootnote}{\fnsymbol{footnote}}
\footnotetext[1]{Work done during internship at Indian Institute of Technology Madras.}
\endgroup

\begin{abstract}
Softmax attention is central to modern vision and diffusion
transformers, but its quadratic complexity in tokens remains
a major computational bottleneck. Uniformly replacing softmax
attention with linear or sparse attention in large pretrained models often degrades performance, as attention rank and sparsity vary across heads and layers, and additionally across denoising timesteps in diffusion transformers. We show that selectively exploiting these two complementary structures enables inference acceleration in large pretrained models while retaining competitive task performance, with the largest gains at long token sequences. In particular, we propose Low-Rank and Sparse Attention (LoSpAttn), a training-free hybrid framework that uses fast, one-time calibration to select low-rank or block-sparse replacements, retaining exact softmax when neither is sufficiently accurate. For reliable low-rank replacement, we develop a numerically stable Nystr\"om approximation using symmetric kernel lifting and
preconditioning, with theoretical error guarantees for
normalized attention. Experiments on ViT-Large, DiT-XL/2, and PixArt-$\Sigma$ demonstrate inference acceleration while maintaining classification accuracy and generation quality comparable to the original pretrained models. For example, on PixArt-$\Sigma$ at 2K resolution ($16{,}384$ tokens), LoSpAttn achieves up to $1.4\times$ end-to-end and $11\times$ attention speedup over the FlashAttention-2. Code provided at \url{https://github.com/J-Karthik-palaniappan/LospAttn}
\end{abstract}    
\section{Introduction}
\label{sec:intro}

Transformers have become central to various computer vision tasks, including visual recognition, image restoration, and image generation~\cite{khan2022transformers}. In particular, Vision Transformers (ViTs) \cite{dosovitskiy2021image} and Diffusion Transformers
(DiTs)~\cite{peebles2023scalable} use self-attention
\cite{vaswani2017attention} to model dependencies between visual tokens, typically representing image patches or latent patches.  However, global self-attention has quadratic computational complexity in the number of tokens, making it increasingly expensive at high image resolutions. 
Both ViTs and DiTs comprise multiple transformer blocks with multi-head self-attention, but the computational challenge is further amplified in DiTs, where the entire network is evaluated repeatedly across successive denoising steps.

Let $\X \in \mathbb{R}^{N \times d_i}$ be the input feature map corresponding to the $N$ visual tokens, each with the $d_i$-dimensional features. For every attention head, the input is linearly projected into query, key, and value matrices $\Q_h,\K_h\in\mathbb{R}^{N\times d}$
and $\V_h\in\mathbb{R}^{N\times d_v}$. The self-attention output is
\begin{equation}
    \boldO_h
    =
    \S_h\V_h,\ 
    \S_h=\D_h^{-1}\G_h,
    \ 
    \G_h=
    \exp\!\left(
        \frac{\Q_h\K_h^\top}{\sqrt d}
    \right)
    \label{eq:selfattention}
\end{equation}
where $\G_h\in\mathbb{R}^{N\times N}$ is the matrix of the unnormalized attention kernel, $\D_h\!=\!\operatorname{diag}(\G_h\mathbf{1}_N)$, and $\S_h$ is the softmax attention matrix. Evaluating $\boldO_h$ requires $\mathcal{O}(N^2d)$ complexity in time and $\mathcal{O}(N^2+Nd)$ in storage when $d_v\!=\!\mathcal O(d)$.  FlashAttention~\cite{dao2022flashattention,dao2024flashattention}
avoids storing the full attention matrix, reducing memory use,
but the computation grows quadratically with the number of tokens. Numerous approaches reduce attention complexity through kernel-based
linearization~\cite{katharopoulos2020transformers,choromanski2021rethinking}, low-rank projections~\cite{wang2020linformer}, Nystr\"om approximations~\cite{xiong2021nystromformer,chen2021skyformer},
sparse attention~\cite{liang2021swinir,hassani2023neighborhood,liu2026clear},
and hybrid schemes that combine complementary attention
mechanisms~\cite{dass2023vitality,zhang2026sla}. However, to preserve the performance of softmax attention, these approaches require training or fine-tuning, which can incur substantial computational costs when adapting large pretrained models. 

\begin{figure*}[t]
    \centering
    \begingroup
    \scriptsize

    \setlength{\tabcolsep}{1pt}
    \renewcommand{\arraystretch}{1}
    \setlength{\storypanel}{0.49\textwidth}
    \setlength{\storylabelwidth}{0.18\storypanel}
    \setlength{\storynoteheight}{2.2\baselineskip}
    \setlength{\storytile}{%
        \dimexpr
        (\storypanel-\storylabelwidth-6\tabcolsep)/3
        \relax
    }
    \setlength{\storylabelwidth}{18pt}
    \setlength{\storytile}{%
        \dimexpr
        (\storypanel-\storylabelwidth-6\tabcolsep)/3
        \relax
    }
    \begin{minipage}[t]{\storypanel}
        \vspace{0pt}
        \centering

        \begin{tabular}{@{}cccc@{}}
            &
            \storynote{Softmax\\attention matrix} &
            \storynote{Our low-rank\\approximation} &
            \storynote{Sparse\\approximation} \\[2pt]

            \storylabel{Low-rank} &
            \storymap{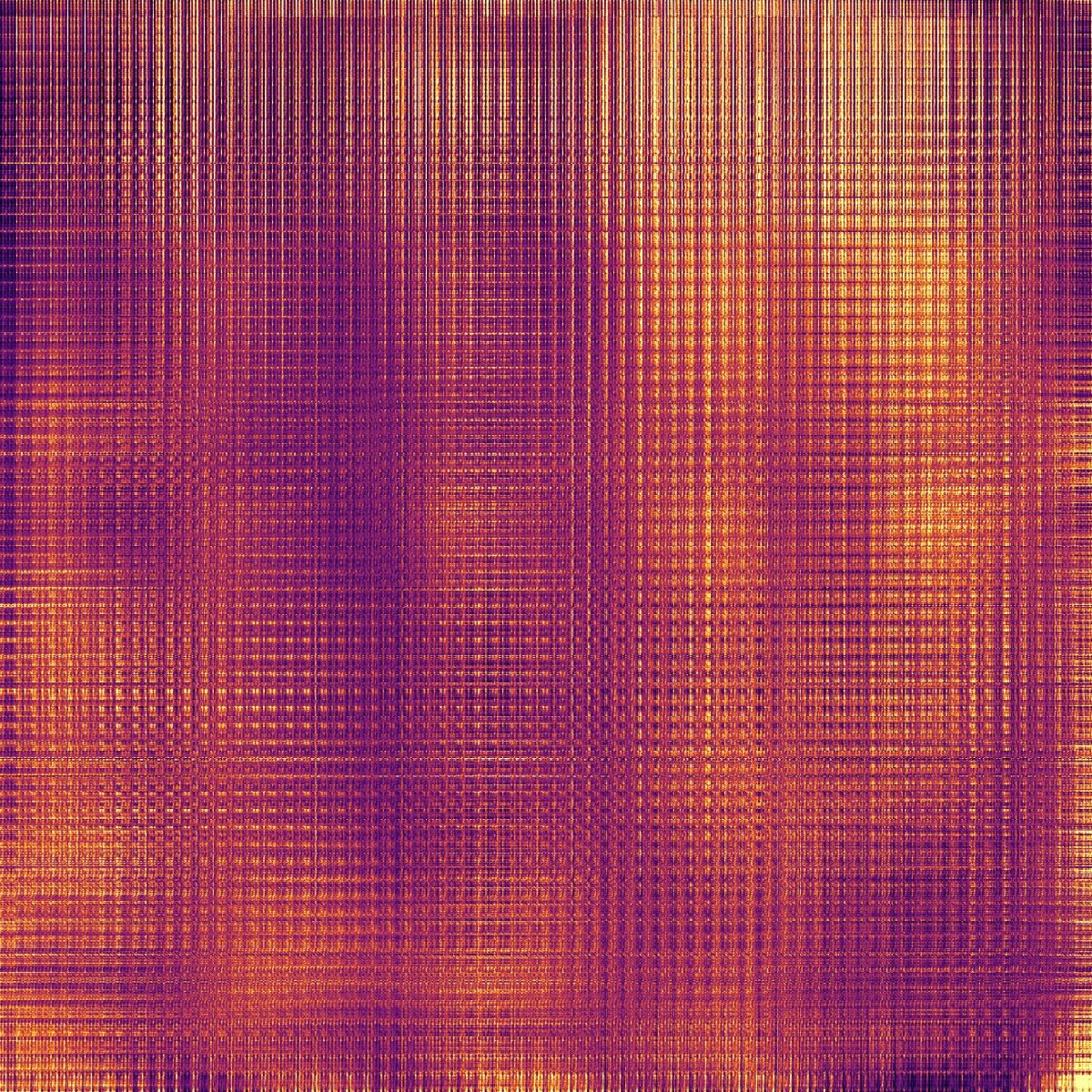} &
            \storymap{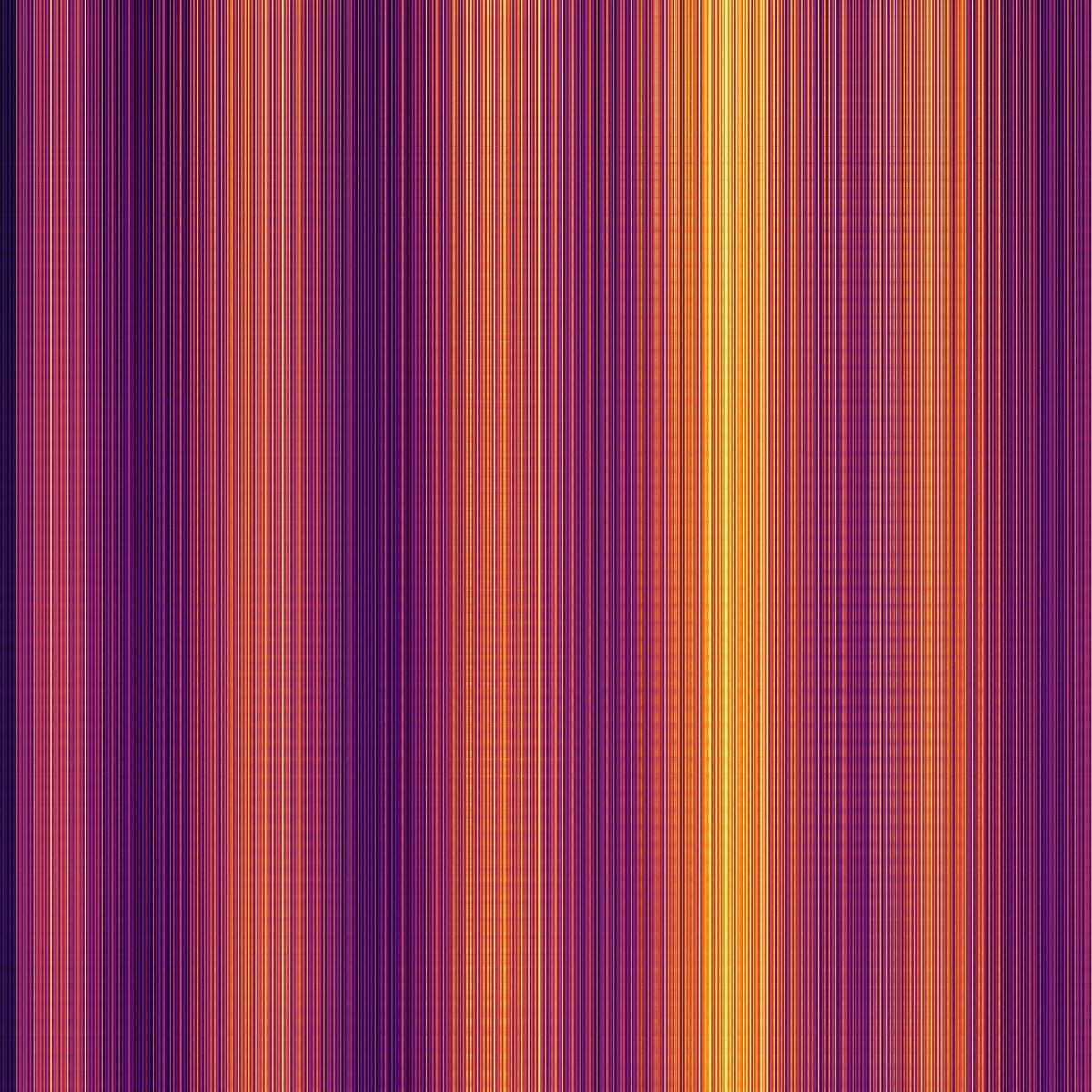} &
            \storymap{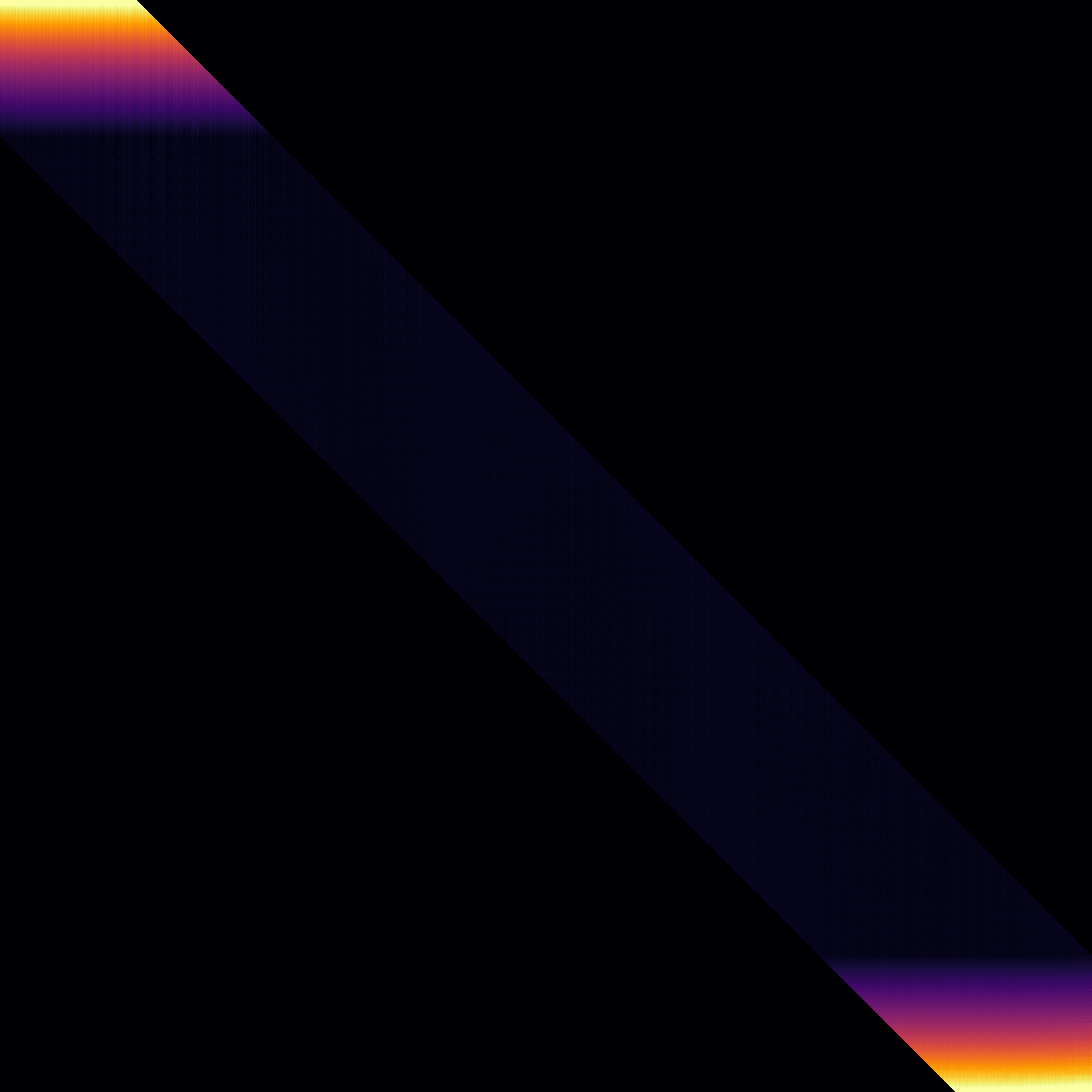} \\

            &
            \storynote{Reference} &
            \storynote{Low error} &
            \storynote{High error} \\[2pt]

            \storylabel{Sparse} &
            \storymap{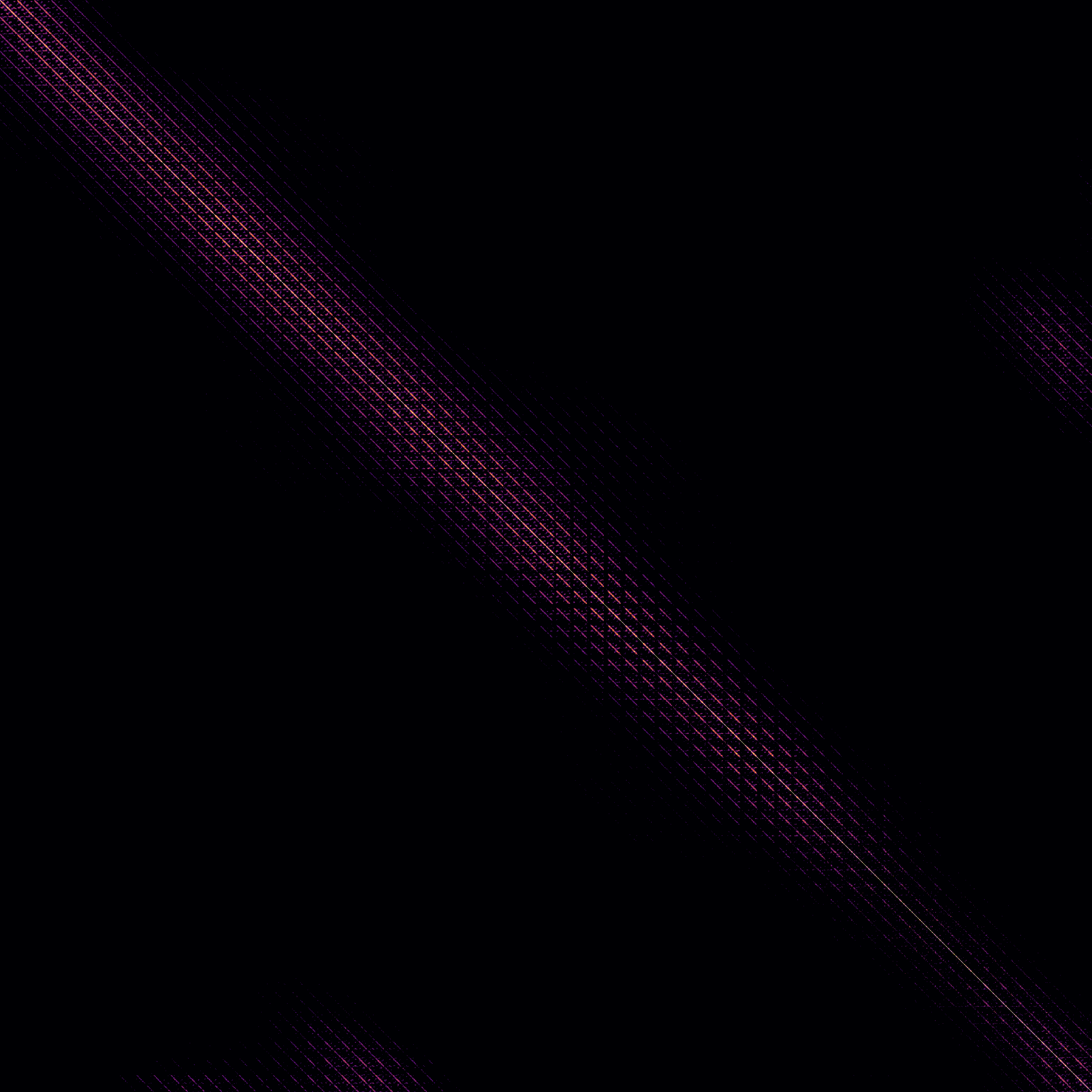} &
            \storymap{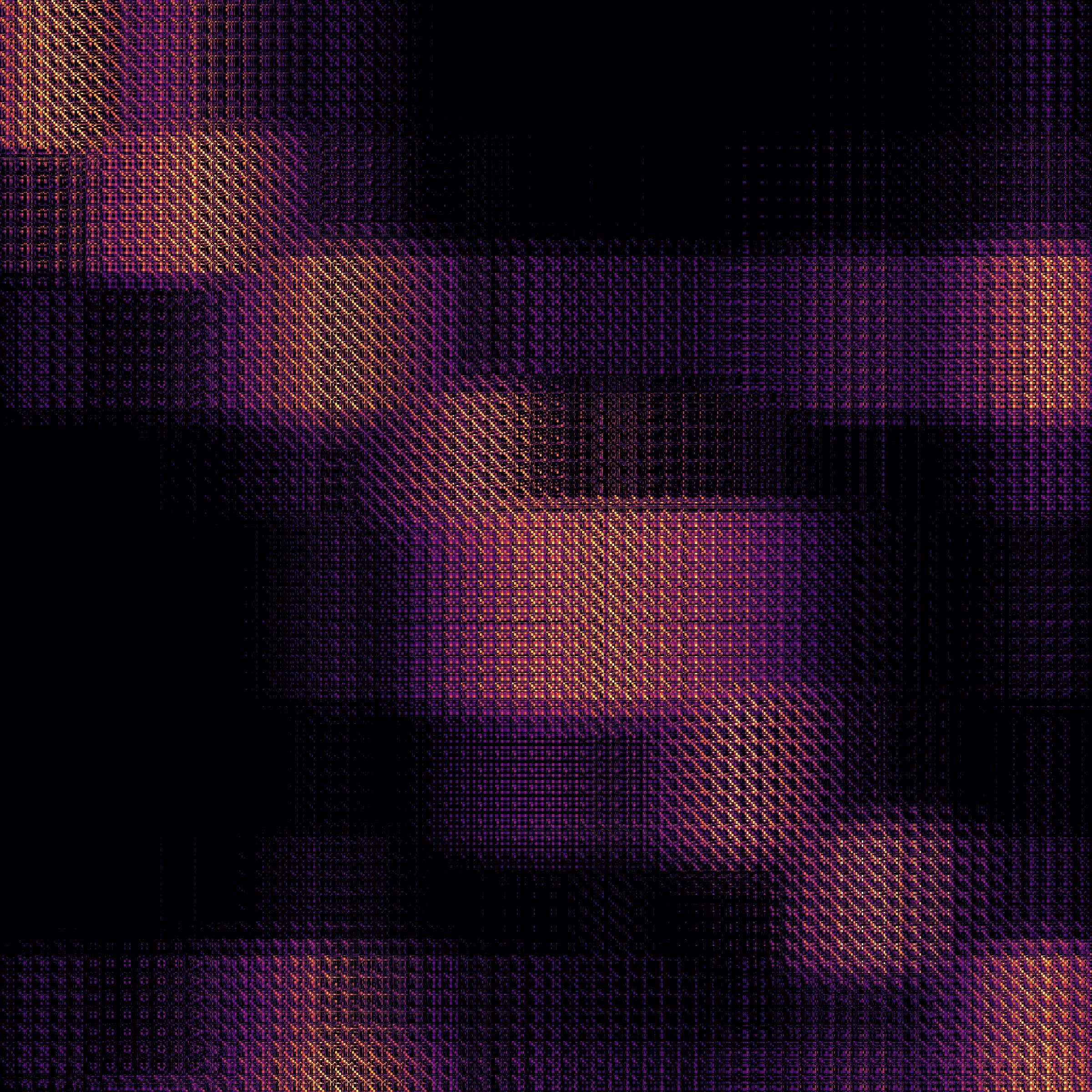} &
            \storymap{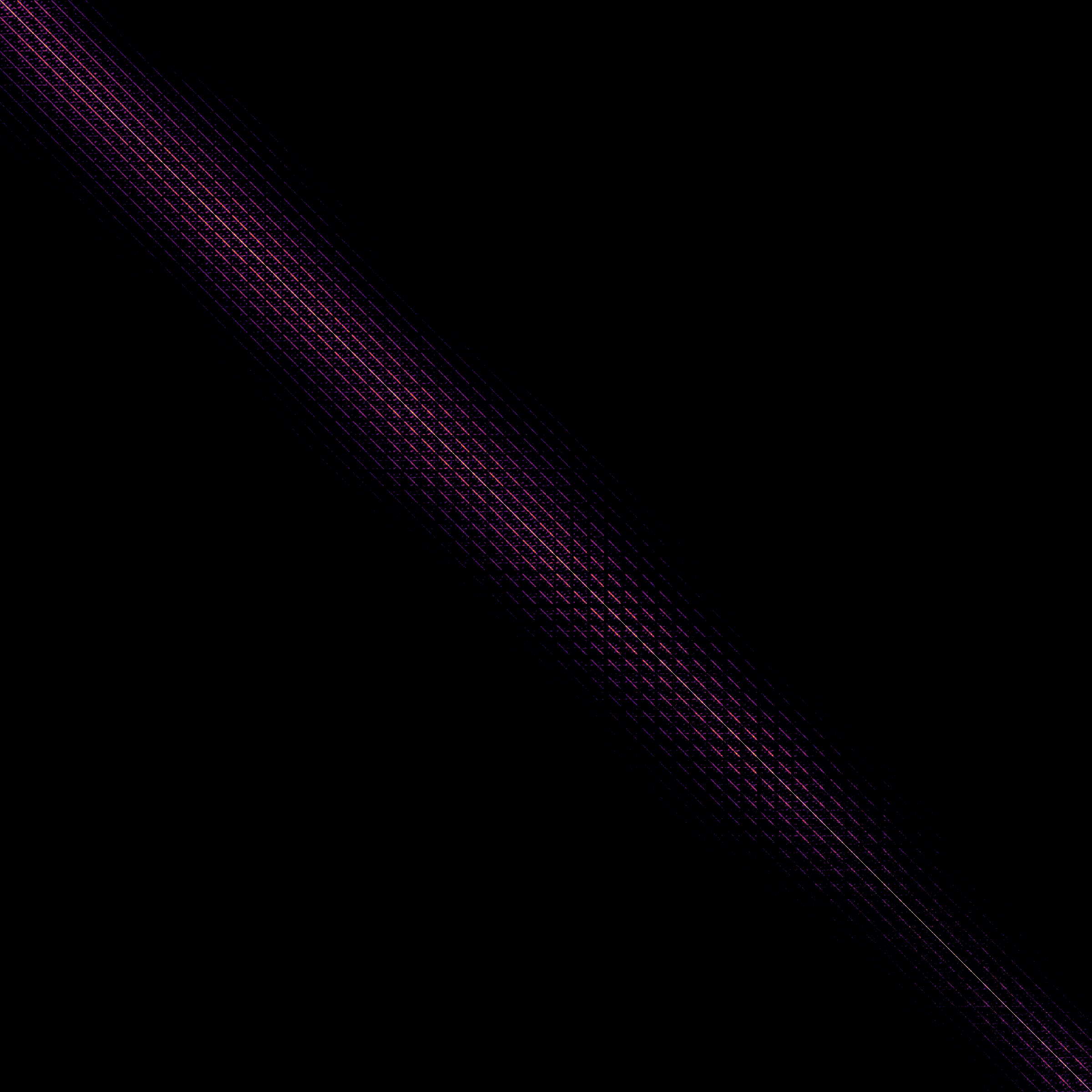} \\

            &
            \storynote{Reference} &
            \storynote{High error} &
            \storynote{Low error} \\[2pt]

            \storylabel{Neither} &
            \storymap{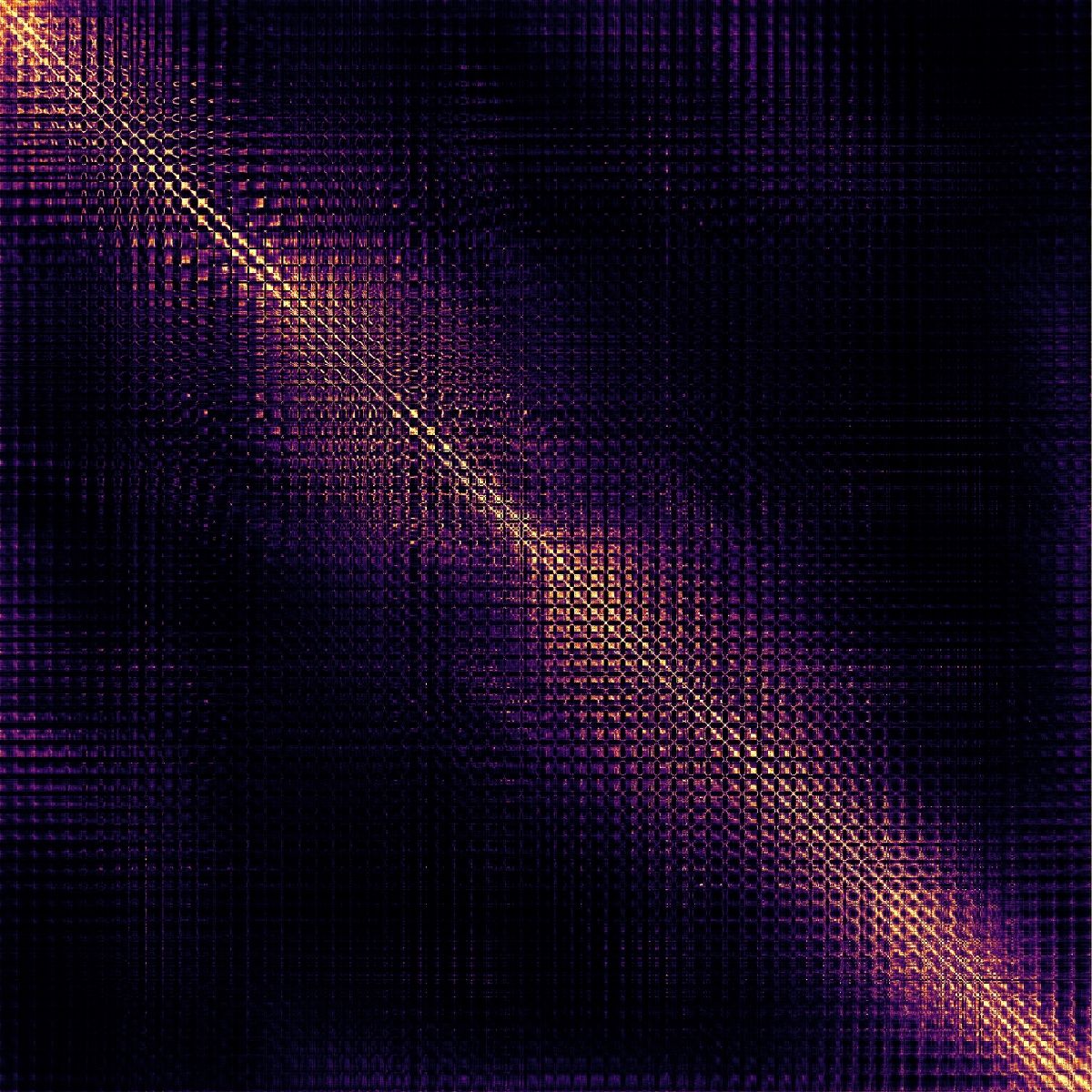} &
            \storymap{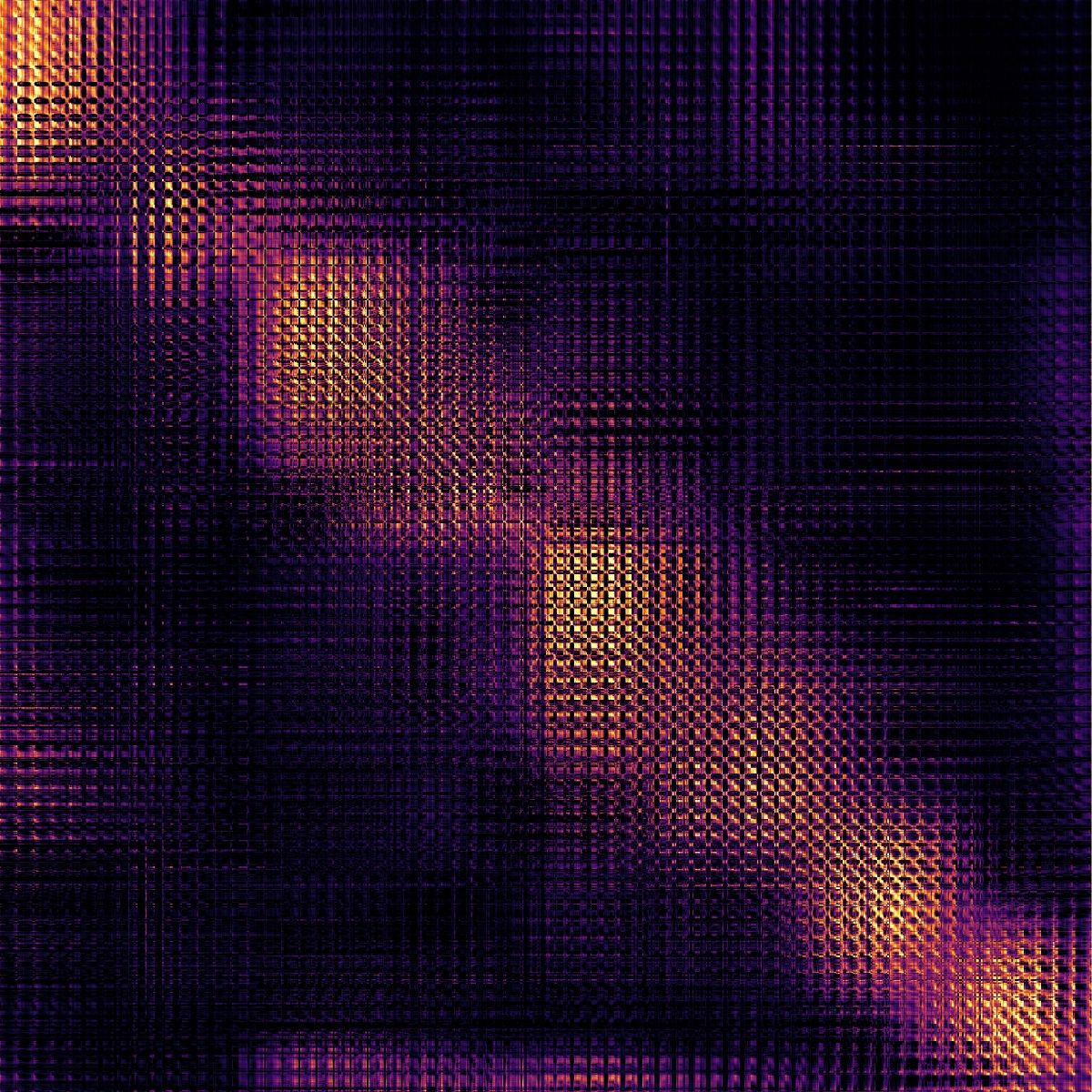} &
            \storymap{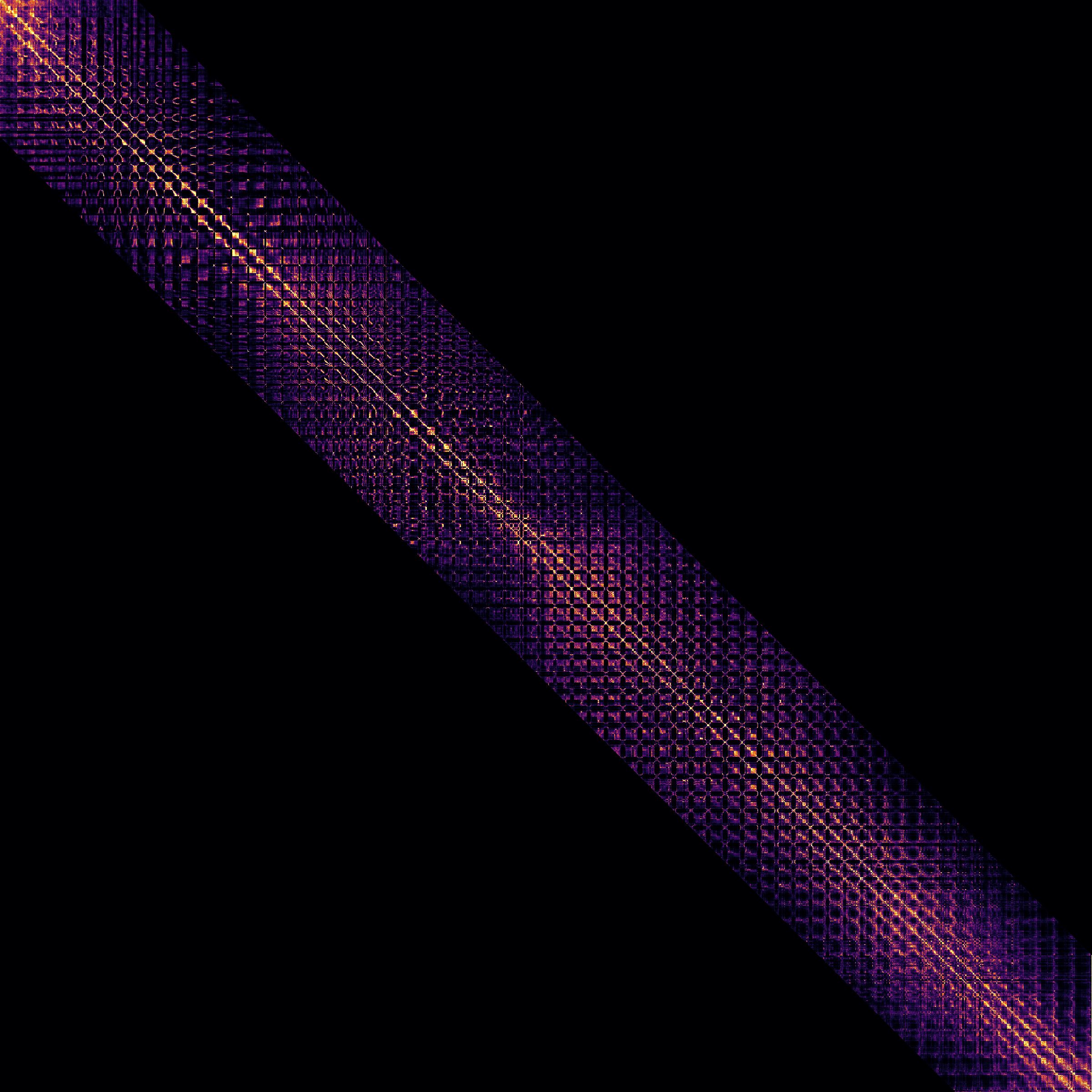} \\

            &
            \storynote{\textbf{Retain exact}} &
            \storynote{High error} &
            \storynote{High error} \\
        \end{tabular}

        \par\vspace{3pt}
        {\footnotesize
        \textit{(a) Complementary attention structures}}
    \end{minipage}
    \hfill
    \begin{minipage}[t]{\storypanel}
        \vspace{0pt}
        \centering

        \begin{tabular}{@{}cccc@{}}
            &
            \storynote{Pretrained\\FlashAttention-2 baseline} &
            \storynote{Low-rank only\\replacement} &
            \storynote{\textbf{LoSpAttn}} \\[2pt]

            \storylabel{\mbox{$N=16{,}384$}} &
            \storyimage{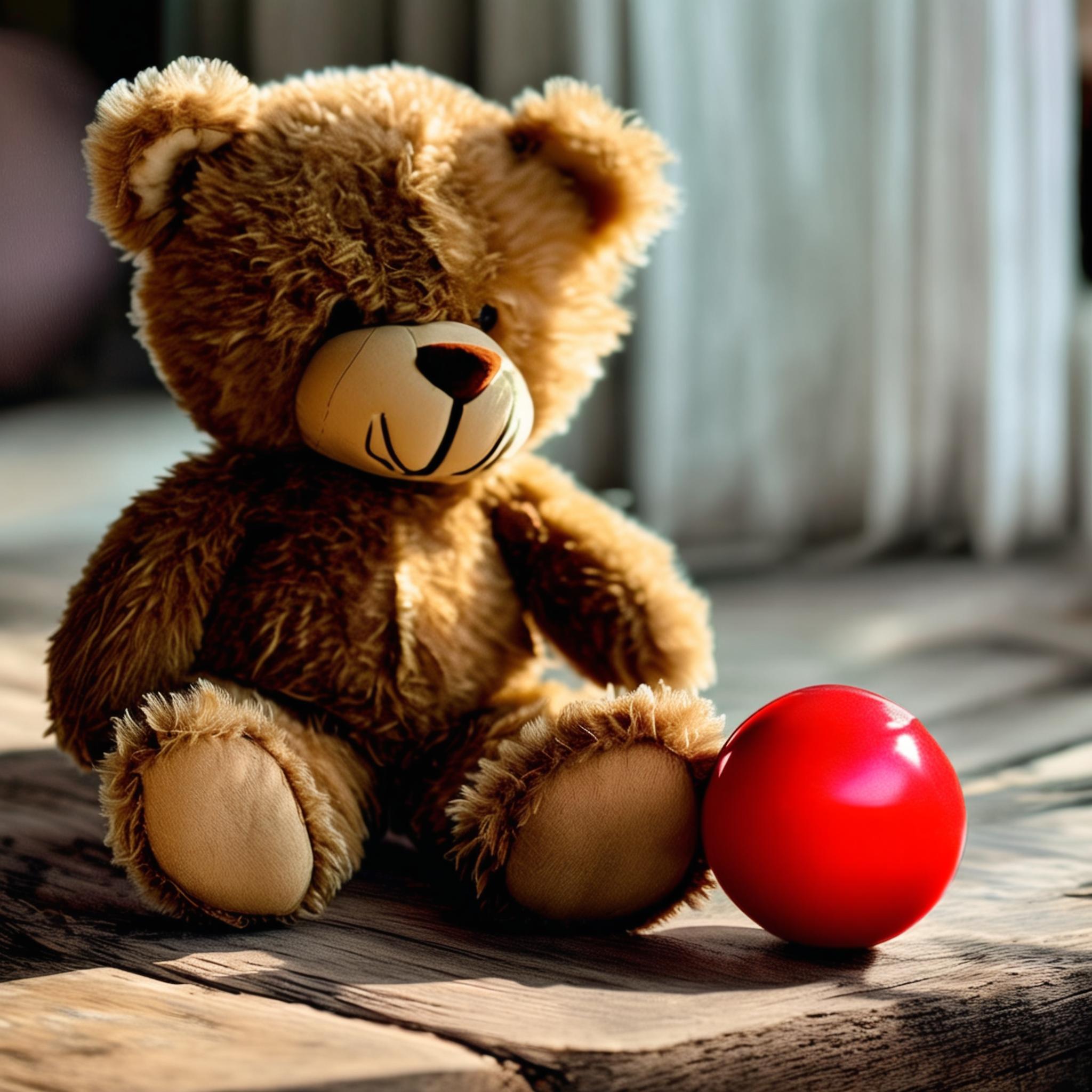} &
            \storyimage{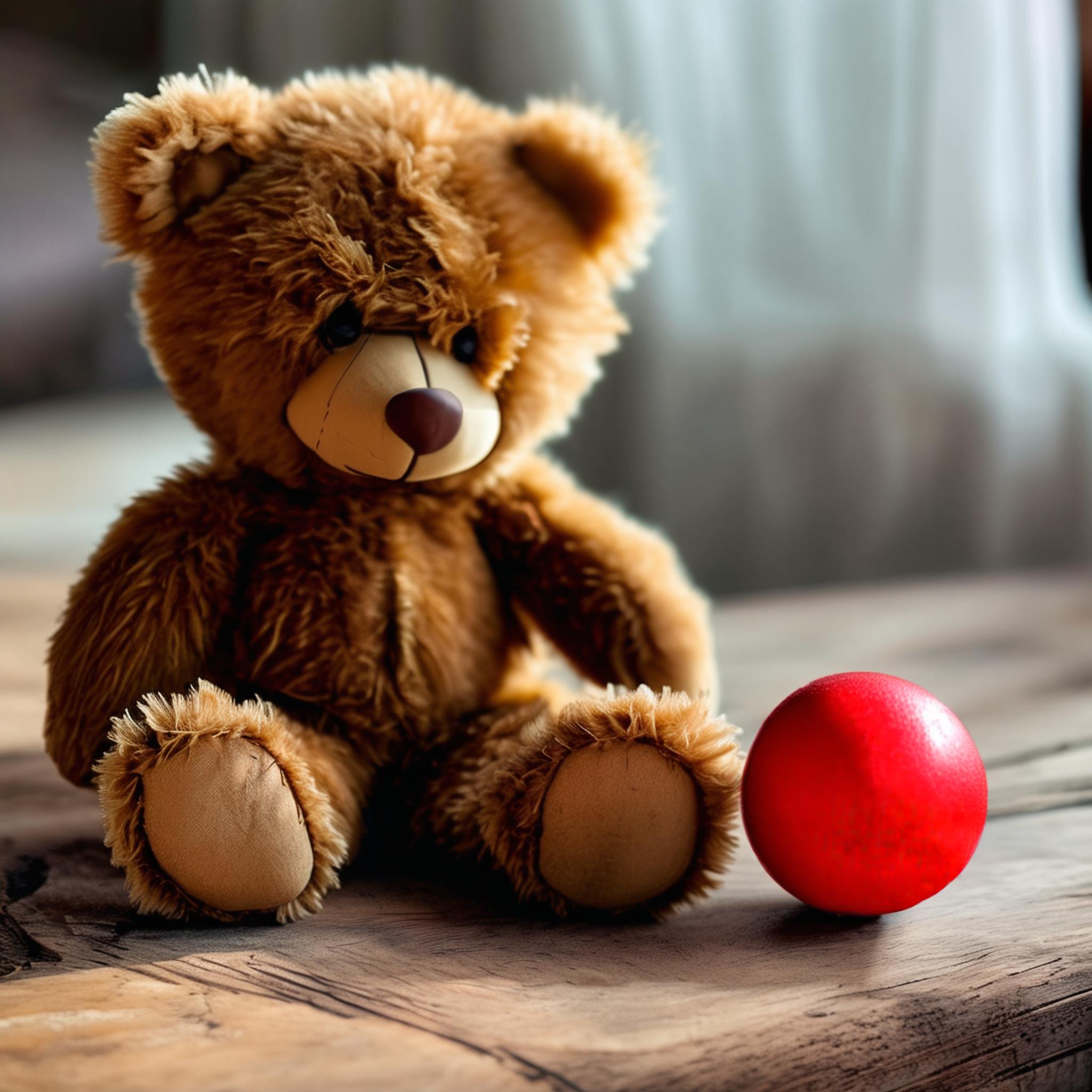} &
            \storyimage{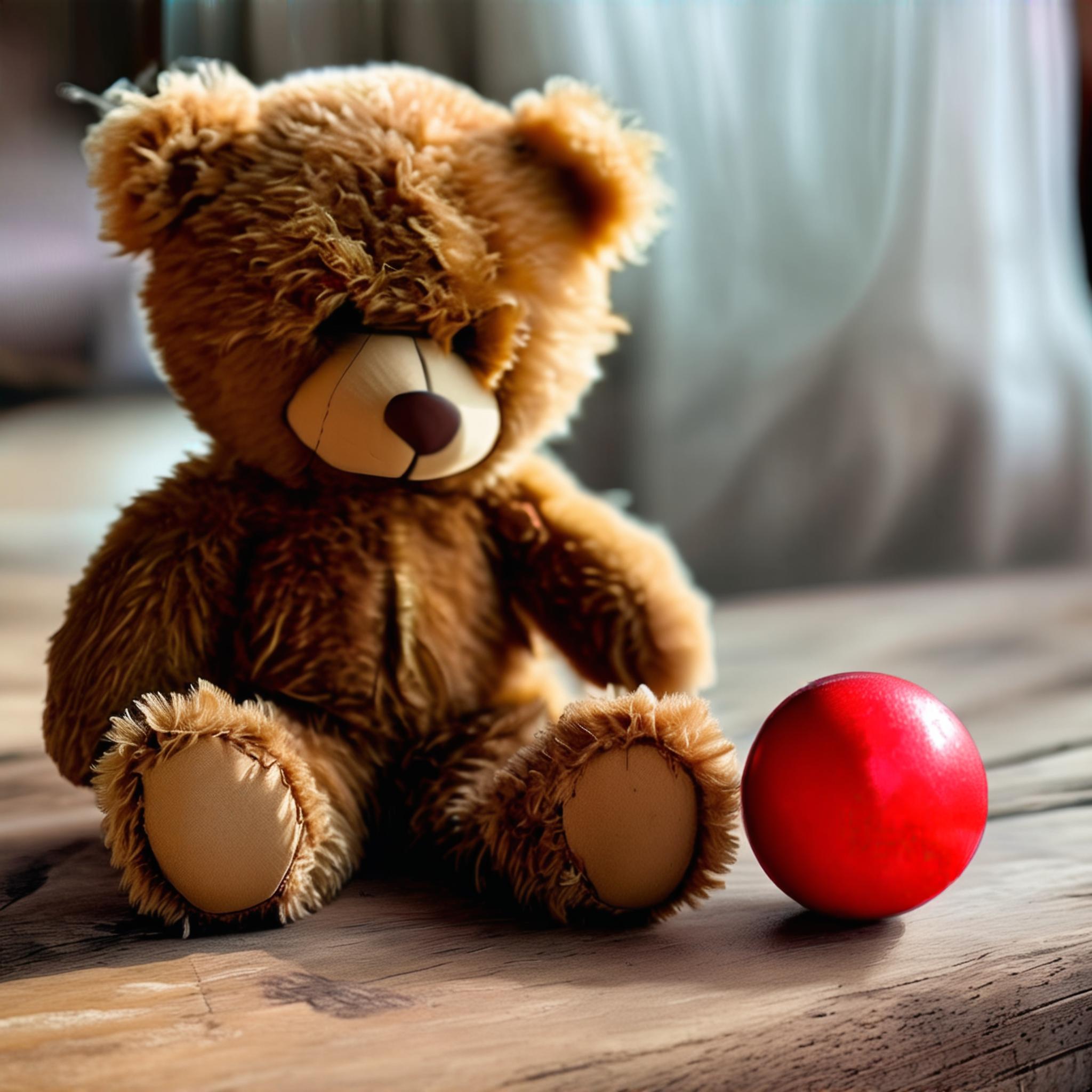} \\

            &
            \storynote{$100\%$ softmax} &
            \storynote{$30\%$ LR } &
            \storynote{$30\% $LR, 
                       $45\%$ SP} \\[2pt]

            \storylabel{\mbox{$N=4{,}096$}} &
            \storyimage{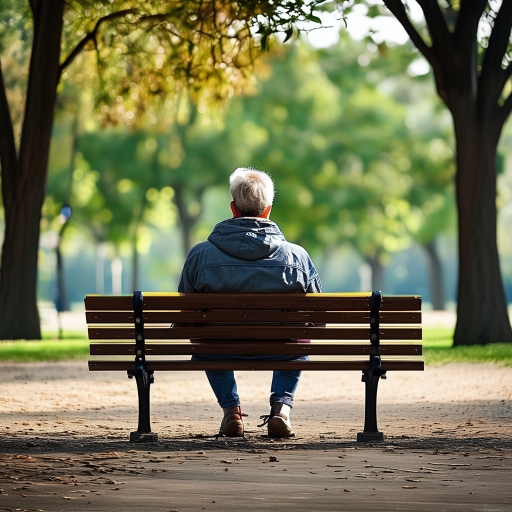} &
            \storyimage{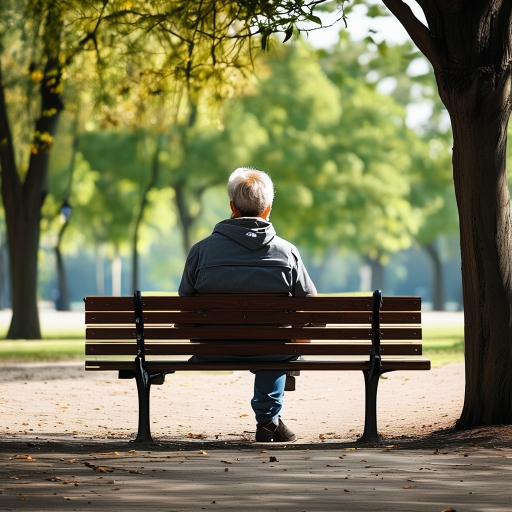} &
            \storyimage{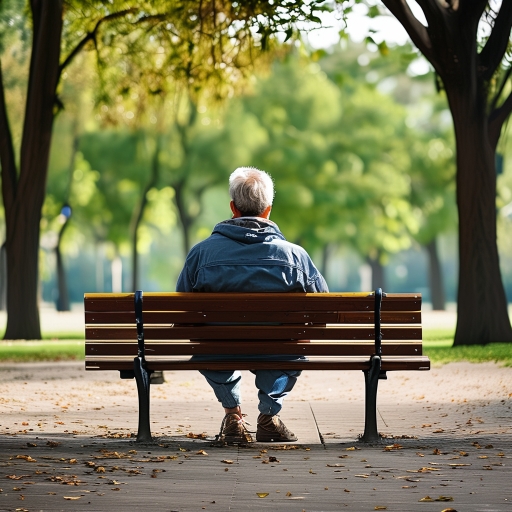} \\
            
            &
            \storynote{$100\%$ softmax} &
            \storynote{$30\%$ LR } &
            \storynote{$30\% $LR, 
                       $43\%$ SP} \\[2pt]

            \storylabel{\mbox{$N=1{,}024$}} &
            \storyimage{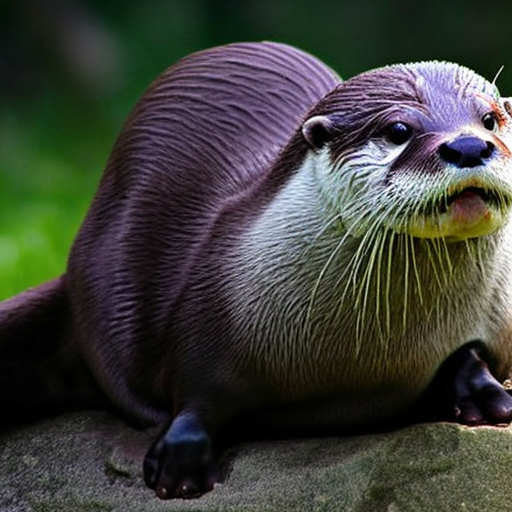} &
            \storyimage{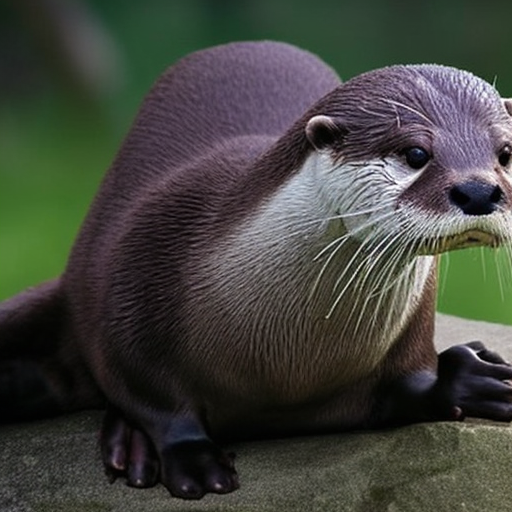} &
            \storyimage{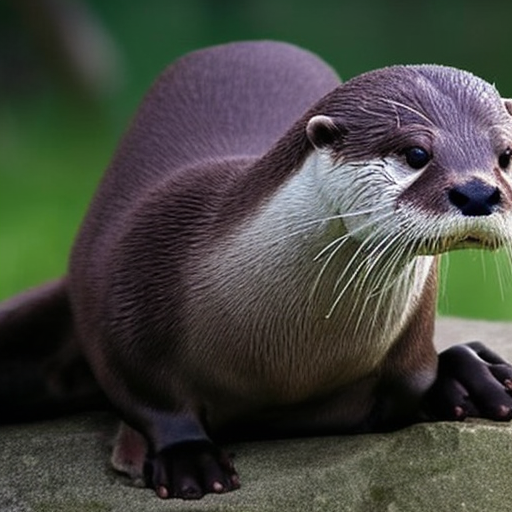} \\

            &
            \storynote{$100\%$ softmax} &
            \storynote{$44\%$ LR } &
            \storynote{$44\% $LR, 
                       $45\%$ SP} \\[2pt]
        \end{tabular}

        \par\vspace{3pt}
        {\footnotesize
        \textit{(b) Generation with selective head replacement}}
    \end{minipage}

    \endgroup

    \caption{\textbf{Structure Guides Attention.} 
        (a) Some Pixart-$\Sigma$ attention heads are accurately captured by our low-rank or sparse approximation, while
        neither approximation accurately captures the third head
        (b) Adding sparse replacement to low-rank replacement increases end-to-end speedup over the pretrained baseline from $1.17\times$ to $1.40\times$ at $N=16{,}384$, while both maintaining comparable visual quality to the baseline. LR/SP denote low-rank/sparse
replacement percentages and the remainder uses exact attention.}
    \label{fig:combined_comparison}
\end{figure*}

Training-free drop-in methods reduce attention computation
without updating model weights.  
KDEformer~\cite{zandieh2023kdeformer} uses importance
sampling to approximate attention, while
Thinformer~\cite{carrell2025lowrank} computes attention
over a representative subset of key--value pairs.
WildCat~\cite{schroder2026wildcat} selects keys and uses
Nystr\"om weights to combine the original values
into a smaller representation. 
MonarchAttention~\cite{yaras2026monarchattention}
approximates $\S_h$ with a structured Monarch matrix,
enabling a subquadratic approximation of the output in \eqref{eq:selfattention}. Scatterbrain~\cite{chen2021scatterbrain} adds sparse corrections
to a low-rank approximation of $\G_h$, while
SpargeAttention~\cite{zhang2025spargeattn} predicts which
attention blocks can be skipped.
For diffusion transformers,
DiTFastAttn~\cite{yuan2024ditfastattn} and
DiTFastAttnV2~\cite{zhang2025ditfastattnv2} combine sparse
attention with attention-output reuse, the latter
selecting strategies per head.
CalibAtt~\cite{yehezkel2026accelerating} reuses calibrated
sparse patterns across inputs, while
HEART~\cite{zheng2026haste} calibrates head-wise sparsity
and adaptively reuses patterns across denoising steps. 
Despite these advances, it remains unclear how effectively
head-wise selection among low-rank, sparse, and exact
attention can accelerate diverse large pretrained models
while retaining competitive task performance.

We address this question with LoSpAttn, a training-free
framework that couples our preconditioned Nystr\"om
approximation with one-time calibration of attention
choices across heads and, in diffusion models, across denoising
timesteps. The attention structure of each head depends on its learned projection matrices and the distribution of input characteristics, making it suitable for a low-rank approximation, a sparse approximation, both, or
none~\cite{chen2021scatterbrain}. For example, in  Fig.~\ref{fig:combined_comparison}(a), the low-rank and sparse DiT attention maps are accurately recovered only by their respective approximations, while neither approximation accurately recovers the third map. These observations motivate adapting the attention computation to individual heads and, additionally, to denoising timesteps in diffusion models. Our main contributions are as follows:
\begin{itemize}
    \item We introduce a fast one-time calibration preprocessing step for each model architecture. This step selects low-rank, block-sparse, or exact softmax attention for individual heads across layers of vision transformers, and additionally across denoising steps for diffusion transformers.
    
    \item We develop a numerically stable Nystr\"om approximation for low-rank heads using symmetric kernel lifting and
    preconditioning, with a bound on the error in approximating the normalized attention matrix.

    \item Experiments on ViT-Large, DiT-XL / 2 and PixArt- $\Sigma$
 demonstrate the applicability of the same calibration
    strategy across classification and generation, with the
    largest inference gains at high resolutions.  Fig.~\ref{fig:combined_comparison}(b) shows a representative result in which selective low-rank and sparse approximations accelerate inference while maintaining generation quality comparable to the FlashAttention-2 baseline.
    \end{itemize}

\section{Proposed Method}
\label{sec:proposed_method}

LoSpAttn adapts the attention computation in each head, selecting our stable Nystr\"om approximation, standard local-window attention~\cite{yuan2024ditfastattn}, or exact softmax attention.
Section~\ref{sec:symm_approx} presents the derivation,
algorithm, and theoretical guarantees of our low-rank
approximation. Section~\ref{sec:rank_analysis} describes
our efficient calibration procedure, performed once per pretrained
model, that assigns low-rank, block-sparse, or exact softmax
attention across heads and layers, with timestep-specific
assignments in diffusion transformers. At inference, selected heads use the low-rank or sparse approximation assigned during calibration, reducing computation while the remaining heads retain exact
softmax attention.

\subsection{Numerically Stable Nystr\"om Attention}
\label{sec:symm_approx}
We first develop a low-rank approximation for softmax attention $\S_h$ in \eqref{eq:selfattention}.  We omit the head index for the sake of clarity. Recall from \eqref{eq:selfattention} that the unnormalized attention matrix is $\mathbf{G}=\exp(\Q\K^\top/\sqrt d)$, where the exponential is applied elementwise. The attention output can be written as
\begin{equation}
    \mathbf{O}
    =
    (\mathbf{G}\V)
    \oslash
    (\mathbf{G}\mathbf{1}_N),
    \label{eq:compact_attention}
\end{equation}
where $\oslash$ denotes row-wise division. Our objective
is to approximate the two matrix products in
\eqref{eq:compact_attention} without computing or storing the full  
$N\times N$ matrix. Since $\mathbf{G}$ is generally asymmetric, we adopt a symmetric lifting strategy ~\cite{chen2021skyformer}, embedding queries
and keys into a joint feature set to construct a symmetric positive-semidefinite matrix,
\begin{equation}
    \Z
    =
    \begin{bmatrix}
        \Q\\
        \K
    \end{bmatrix}
    \in\mathbb{R}^{2N\times d}.
\end{equation}
The corresponding exponential Gram matrix is
\begin{equation*}
\mathbf{H}
    \!=\!
    \exp\!\left(
        \frac{\Z\Z^{\top}}{\sqrt d}
    \right)
\!=\!\left[\begin{matrix}
\exp(\Q\Q^\top/\sqrt d) & \mathbf{G}\\
\mathbf{G}^\top & \exp(\K\K^\top/\sqrt d)
\end{matrix}\right]
\end{equation*}
The exponential kernel $\mathbf H$ is symmetric positive semidefinite~\cite{kar2012random}, with $\mathbf G$ appearing as its upper-right block. We next construct paired landmarks
$\Q_m,\K_m\in\mathbb{R}^{m\times d}$ by average pooling
queries and keys over the same non-overlapping spatial
windows~\cite{lu2021soft}, and define 
\begin{equation*}
    \mathbf{M}
    =
    \begin{bmatrix}
        \Q_m\\
        \K_m
    \end{bmatrix}
    \in\mathbb{R}^{2m\times d},
    \qquad m\ll N.
\end{equation*}
This landmark construction preserves the correspondence between the query and key representations of visual tokens. We then form three smaller exponential-kernel factors:
\begin{equation}
\begin{aligned}
    \mathbf{A}
    &=
    \exp\!\left(
        \frac{\Q\mathbf{M}^{\top}}{\sqrt d}
    \right)
    \in\mathbb{R}^{N\times 2m},\\
    \mathbf{B}
    &=
    \exp\!\left(
        \frac{\mathbf{M}\mathbf{M}^{\top}}{\sqrt d}
    \right)
    \in\mathbb{R}^{2m\times 2m},\\
    \mathbf{C}
    &=
    \exp\!\left(
        \frac{\mathbf{M}\K^{\top}}{\sqrt d}
    \right)
    \in\mathbb{R}^{2m\times N}.
\end{aligned}
\label{eq:nystrom_factors}
\end{equation}
Applying Nystr\"om approximation to the symmetric matrix $\mathbf H$ and extracting its upper-right block yields
\begin{equation}
    \widehat{\mathbf G}
    =\mathbf A\mathbf B^\dagger\mathbf C.
    \label{eq:classical_nystrom}
\end{equation}
Thus, the approximation requires neither the construction of $\mathbf H$ nor its dense blocks $N\times N$.
\begin{algorithm}[t]
\caption{Preconditioned Symmetric Nystr\"om Attention}
\label{alg:pnpnystra}
\begin{algorithmic}[1]
\REQUIRE $\Q,\K\in\mathbb{R}^{N\times d}$,
$\V\in\mathbb{R}^{N\times d_v}$,
landmarks $m$, iterations $t$

\STATE Obtain $\Q_m,\K_m$ by spatial pooling;
$\M=[\Q_m;\K_m]$.

\STATE Compute $\A$ as in \eqref{eq:nystrom_factors}.

\STATE Compute normalized matrices 
$\tilde{\B}$ and
$\tilde{\C}$ as in \eqref{eq:newBandC}.

\STATE $\tilde{\B}_\mathrm{inv}\leftarrow
\operatorname{IterPinv}(\tilde{\B},t)$.

\STATE $\V_{\mathrm{aug}}
\leftarrow[\V,\mathbf{1}_N]$.

\STATE $\mathbf{Y}\leftarrow
\A\big(\tilde{\B}_\mathrm{inv}(\tilde{\C}\V_{\mathrm{aug}})\big)$.

\STATE $\widehat{\mathbf{O}}_{i,:}
\leftarrow
\mathbf{Y}_{i,1:d_v}/\mathbf{Y}_{i,d_v+1}$.

\STATE \textbf{Return} $\widehat{\mathbf{O}}$.
\end{algorithmic}
\end{algorithm}
However, computing $\B^\dagger$ in
\eqref{eq:classical_nystrom} can be numerically unstable,
as the exponential entries of $\B$ may span several orders
of magnitude~\cite{higham2002accuracy}. For example, $\exp(12)\approx1.63\times10^5$
overflows in FP16, while $\exp(-20)\approx2.06\times10^{-9}$
underflows to zero. These representation errors are 
amplified during pseudoinversion, degrading the resulting
attention approximation (see Table.~\ref{tab:vit_head_allocation}).

To mitigate this stability issue, we apply row-wise preconditioning by normalizing $\B$ and
$\C$. Defining
\begin{equation*}
    \D_B=\operatorname{diag}(\B\mathbf{1}_{2m}),
    \qquad
    \D_C=\operatorname{diag}(\C\mathbf{1}_N),
\end{equation*}
we obtain
\begin{equation}
    \tilde{\B}=\D_B^{-1}\B,\qquad
    \tilde{\C}=\D_C^{-1}\C.
    \label{eq:newBandC}
\end{equation}
Independent normalization modifies our approximation as,
\begin{equation}
    \widehat{\G}
    =
    \A\tilde{\B}^{\dagger}\tilde{\C},
    \label{eq:preconditioned_nystrom}
\end{equation}
which generally differs from the classical approximation Nystr\"om
 in \eqref{eq:classical_nystrom}.
We then normalize its rows:
\begin{equation}
    \widehat{\S}
    =
    \operatorname{diag}
    (\widehat{\G}\mathbf{1}_N)^{-1}\widehat{\G}.
    \label{eq:normalized_approx}
\end{equation}
The experiments in Section~\ref{sec:experiments} demonstrate
improved numerical stability and accurate approximation
of $\S$ by $\widehat{\S}$. Algorithm~\ref{alg:pnpnystra} computes
$\widehat{\mathbf O}=\widehat{\S}\V$ without forming
$\widehat{\G}$ or $\widehat{\S}$. 
Step~2 evaluates a row-scaled $\A$ using row-max shifts;
this scaling cancels in the final division.
Step~3 computes $\tilde{\B},\tilde{\C}$ using stable
row-wise softmax, and Step~4 approximates the pseudoinverse
using a fixed number of matrix
iterations~\cite{razavi2014new}, avoiding expensive SVD operations.

Next, we establish conditions under which the preconditioned
approximation $\widehat{\S}$ in \eqref{eq:normalized_approx}
is close to $\S$. For query--key vectors
$\mathcal X=\{\mathbf q_i,\mathbf k_i\}_{i=1}^N$
and landmarks
$\mathcal M=\{\mathbf m_\ell\}_{\ell=1}^{2m}$,
define the maximum landmark approximation error as
\begin{equation}
    h_m =
    \max_{\mathbf x\in\mathcal X}
    \min_{\mathbf z\in\mathcal M}
    \|\mathbf x-\mathbf z\|_2.
    \label{eq:landmark_error}
\end{equation}


\begin{figure*}[t]
    \centering
    \captionsetup[subfigure]{
        font=small,
        justification=centering,
        skip=3pt
    }

    \begin{subfigure}[t]{0.49\textwidth}
        \centering
            \includegraphics[
                width=8cm,
                height=5cm
            ]{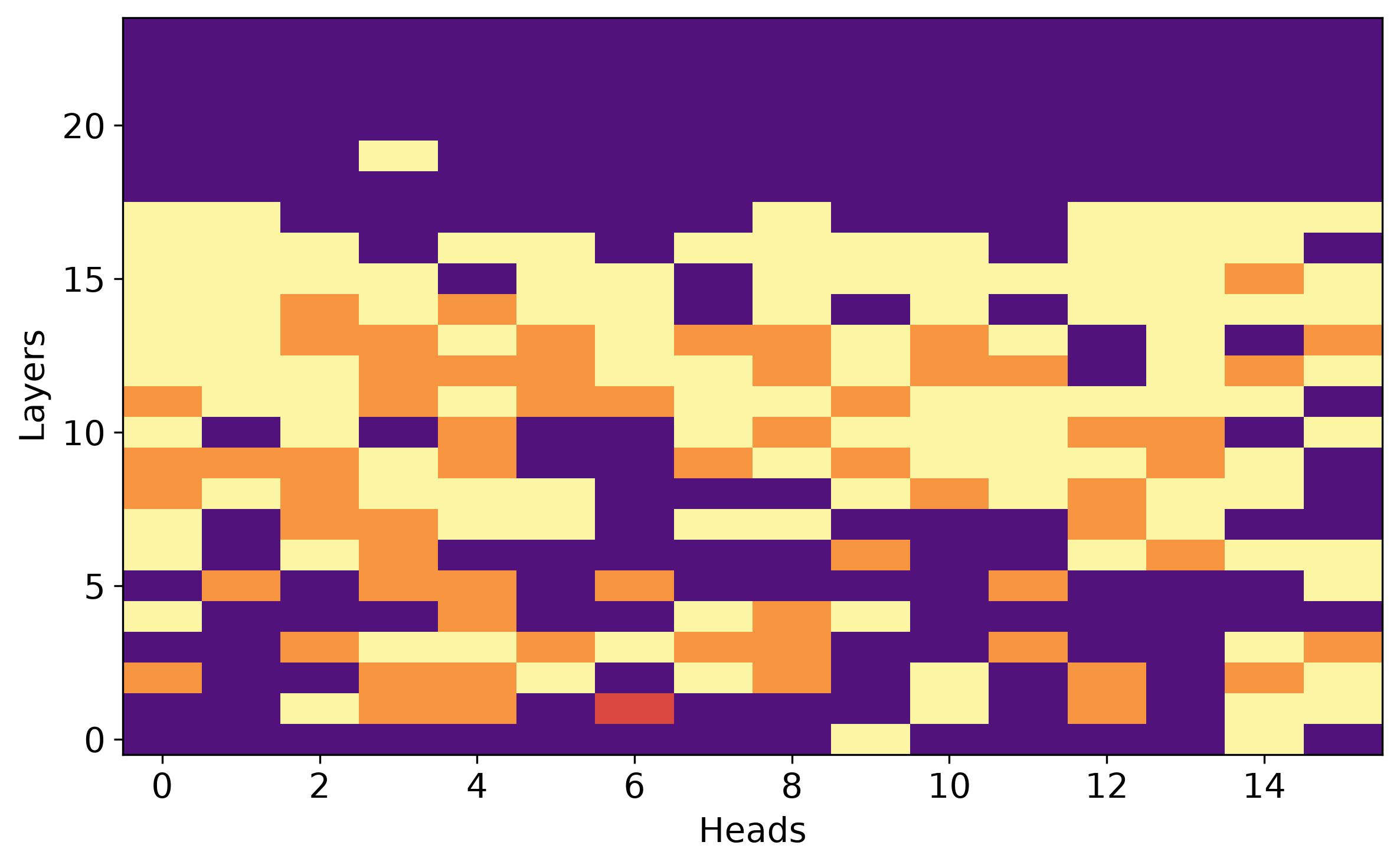}
        \caption{Head structure across ViT-Large layers.}
        \label{fig:coverage_vit}
    \end{subfigure}
    \begin{subfigure}[t]{0.49\textwidth}
        \centering
            \includegraphics[
                width=\linewidth,
                height=5cm
            ]{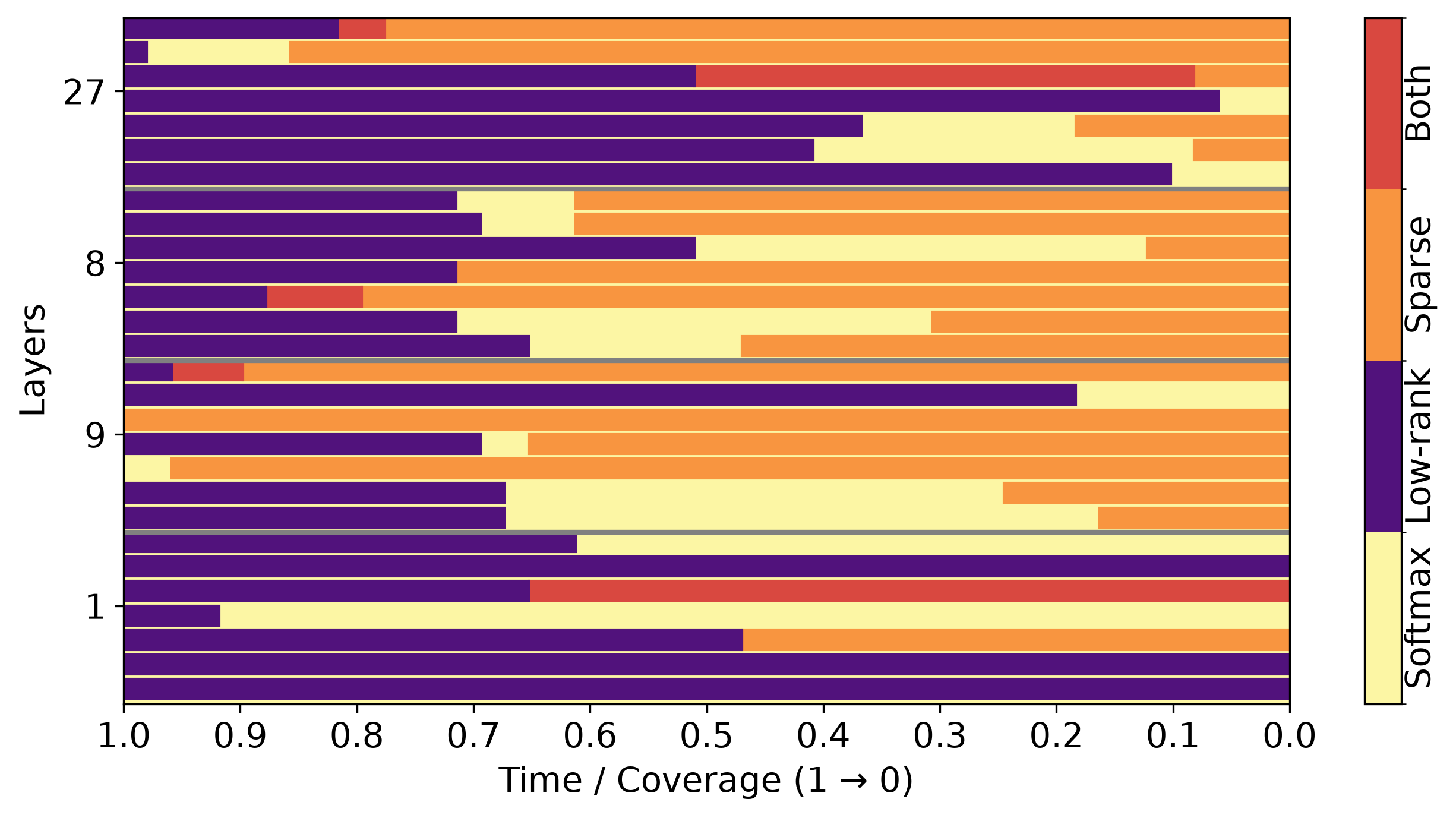}
        \caption{Head structure across DiT-XL/2 timesteps.}
        \label{fig:coverage_dit}
    \end{subfigure}

\caption{\textbf{Complementary attention structures.}
Calibration identifies heads suitable for low-rank
approximation, sparse approximation, both, or neither
(``Softmax'') under the prescribed error tolerances.
Suitability varies across (a) ViT-Large heads and layers
and (b) DiT-XL/2 heads and denoising timesteps.}
    \label{fig:nystrom_rank_heatmaps}
\end{figure*}

\begin{theorem}[Approximation Error]
\label{thm:approximation_error}
Suppose all query, key, and landmark vectors
have Euclidean norm at most $R_0$,
$\tilde{\B}$ is invertible, and
$\|\tilde{\B}^{-1}\tilde{\C}\|_\infty
\le M_0$. Then,
\begin{equation}
    \|\S-\widehat{\S}\|_\infty
    \le
    \frac{R_0}{\sqrt d}(1+M_0)h_m,
    \label{eq:attention_error_bound}
\end{equation}
where $\|\cdot\|_\infty$ denotes the induced 
matrix infinity norm. Moreover, $\widehat{\S}=\S$ when $m=N$
and $\M=[\Q;\K]$.
\end{theorem}
Fig.~\ref{fig:error_bound} in Section~\ref{sec:experiments}
compares measured approximation errors with the theoretical
bound across sampled heads and landmark budgets. The proof for Theorem.~\ref{thm:approximation_error} is provided in the appendix. With full landmarks ($m=N$), $\widehat{\G}$ differs from $\G$ only by a positive row scaling, which cancels upon normalization to recover $\S$ exactly.

\subsection{Structure-Guided Head Selection}
\label{sec:rank_analysis}

We now introduce an efficient one-time calibration strategy
to select heads for low-rank or block-sparse approximation
in a pretrained model. For each head, we measure the errors of the proposed Nystr\"om approximation and the standard local-window attention on calibration samples. A small Nystr\"om error indicates
that the attention matrix is approximately low rank, while
a small local-window error indicates that it is accurately
captured by a diagonally sparse pattern. These errors guide head selection, with exact softmax attention retained for heads not selected for either approximation.

Recall from Section~\ref{sec:symm_approx} that our normalized
Nystr\"om approximation has rank at most $2m$. For each head,
we evaluate low-rank candidates with landmark budgets
$m\in\mathcal M$ and a block-sparse candidate with local-window
radius $b=N/8$~\cite{yuan2024ditfastattn}. Given a calibration set
$\mathcal D=\{\X^{(s)}\}_{s=1}^{S}$, let $\S_h^{(s)}$ denote the exact attention matrix and $\widehat{\S}_{h,a}^{(s)}$ its normalized approximation in branch $a\in\{\mathrm{lr},\mathrm{sp}\}$. We evaluate both branches using the average relative Frobenius error
\begin{equation}
\mathcal E_{h,a}
=
\frac{1}{S}
\sum_{s=1}^{S}
\frac{
\|\S_h^{(s)}-\widehat{\S}_{h,a}^{(s)}\|_F
}{
\|\S_h^{(s)}\|_F
}.
\label{eq:head_approx_error}
\end{equation}
For the low-rank branch, we write
$\mathcal E_{h,\mathrm{lr}}(m)$ to indicate its dependence
on the landmark budget.

\begin{algorithm}[t]
\caption{Structure-Guided Head Selection}
\label{alg:rank_selection}
\begin{algorithmic}[1]
\REQUIRE Network $f_\theta$ with $H$ heads,
calibration set $\mathcal D$,
required low-rank percentage $P$,
candidate landmark budgets $\mathcal M$,
low-rank tolerance grid $\mathcal G_{\epsilon}$,
sparse tolerance $\epsilon_{\mathrm{sp}}$,
window radius $b$
\ENSURE Low-rank heads $\mathcal H_{\mathrm{LR}}$,
sparse heads $\mathcal H_{\mathrm{SP}}$,
landmark budget $m$

\STATE $K\leftarrow\lceil PH/100\rceil$,
$\mathcal H_{\mathrm{LR}}\leftarrow\varnothing$

\STATE Run calibration forward passes on $\mathcal D$;
obtain $\Q_h,\K_h$, and exact attention $\S_h$
for each head and sample

\STATE Compute block-sparse approximations
$\widehat{\S}_{h,\mathrm{sp}}$ with radius $b$
and errors $\mathcal E_{h,\mathrm{sp}}$
for all heads using \eqref{eq:head_approx_error}

\FOR{$m\in\mathcal M$ in increasing order}
    \STATE Compute low-rank approximations
    $\widehat{\S}_{h,\mathrm{lr}}(m)$
    and errors $\mathcal E_{h,\mathrm{lr}}(m)$
    for all heads using \eqref{eq:head_approx_error}

    \STATE Find the smallest
    $\epsilon_{\mathrm{lr}}\in\mathcal G_{\epsilon}$
    such that at least $K$ heads satisfy
    $\mathcal E_{h,\mathrm{lr}}(m)\le\epsilon_{\mathrm{lr}}$;
    set $\epsilon_{\mathrm{lr}}\leftarrow\infty$
    if none exists

    \IF{$\epsilon_{\mathrm{lr}}<\infty$}
        \STATE $\mathcal H_{\mathrm{LR}}\leftarrow
        \{h:\mathcal E_{h,\mathrm{lr}}(m)
        \le\epsilon_{\mathrm{lr}}\}$
        \STATE \textbf{break}
    \ENDIF
\ENDFOR

\STATE $\mathcal H_{\mathrm{SP}}\leftarrow
\{h\notin\mathcal H_{\mathrm{LR}}:
\mathcal E_{h,\mathrm{sp}}\le\epsilon_{\mathrm{sp}}\}$

\STATE Retain exact softmax attention for all remaining heads

\STATE \textbf{return}
$\mathcal H_{\mathrm{LR}},\mathcal H_{\mathrm{SP}},m$
\end{algorithmic}
\end{algorithm}

For a network with $H$ attention heads across all layers, the minimum required low-rank percentage $P$ corresponds to $K=\lceil PH/100\rceil$ heads. Let $\mathcal G_{\epsilon}$ denote the candidate low-rank error tolerances and $\epsilon_{\mathrm{sp}}$ the sparse error tolerance. We examine landmark budgets $m\in\mathcal M$ in increasing order. For each budget, we find the smallest $\epsilon_{\mathrm{lr}}\in\mathcal G_{\epsilon}$ such that
at least $K$ heads satisfy $\mathcal E_{h,\mathrm{lr}}(m)\le\epsilon_{\mathrm{lr}}$. In the first feasible budget, all heads satisfying this tolerance form the low-rank set $\mathcal H_{\mathrm{LR}}$. If no budget is feasible, $\mathcal H_{\mathrm{LR}}=\varnothing$. Among the remaining heads, those satisfying $\mathcal E_{h,\mathrm{sp}}\le\epsilon_{\mathrm{sp}}$
form the sparse set $\mathcal H_{\mathrm{SP}}$ and all others retain exact softmax attention. Algorithm~\ref{alg:rank_selection} summarizes the procedure.

For vision transformers, calibration assigns an attention
branch to each head (Fig.~\ref{fig:nystrom_rank_heatmaps}(a)).
For diffusion transformers, the temporal variation in the rank of the self-attention heads in Fig.~\ref{fig:nystrom_rank_heatmaps}(b) motivates a schedule with low-rank attention followed by exact softmax and then block-sparse attention. We index the sampling steps as $t=1,\ldots,T$ in denoising order. Applying \eqref{eq:head_approx_error} at each step
gives the errors $\mathcal E_{h,t,\mathrm{lr}}(m)$ and
$\mathcal E_{h,t,\mathrm{sp}}$.
For each candidate budget $m$ and tolerance
$\epsilon_{\mathrm{lr}}$, the low-rank phase ends at the
first step exceeding the tolerance, i.e., 
\begin{equation}
\tau_h^{\mathrm{lr}}
=
\min\left\{
t:\mathcal E_{h,t,\mathrm{lr}}(m)>
\epsilon_{\mathrm{lr}}
\right\}.
\label{eq:timestep_threshold}
\end{equation}
We set $\tau_h^{\mathrm{lr}}=T+1$ if the tolerance is
never exceeded. Following the search order in
Algorithm~\ref{alg:rank_selection}, we select
$(m,\epsilon_{\mathrm{lr}})$ such that
$\sum_{h=1}^{H}(\tau_h^{\mathrm{lr}}-1)
\ge\lceil PHT/100\rceil$.
If no feasible configuration exists, we set
$\tau_h^{\mathrm{lr}}=1$ for all heads.

For each head, sparse attention begins at the earliest step from which its error satisfies $\epsilon_{\mathrm{sp}}$ at every subsequent step:
\begin{equation}
\tau_h^{\mathrm{sp}}
=
\min\left\{
t\ge\tau_h^{\mathrm{lr}}:
\max_{u=t,\ldots,T}\mathcal E_{h,u,\mathrm{sp}}
\le\epsilon_{\mathrm{sp}}
\right\}.
\label{eq:sparse_timestep_threshold}
\end{equation}
The search is restricted to $t\le T$, with
$\tau_h^{\mathrm{sp}}=T+1$ if no such step exists.
The head $h$ therefore uses low-rank attention for
$t<\tau_h^{\mathrm{lr}}$, exact softmax for
$\tau_h^{\mathrm{lr}}\le t<\tau_h^{\mathrm{sp}}$,
and block-sparse attention for $t\ge\tau_h^{\mathrm{sp}}$.
Any of these phases may be empty. The complete algorithm for diffusion transformers is provided in the appendix.
One-time calibration for ViT-Large uses $5000$ calibration samples at a batch-size of $128$ taking $1$\,min $40$\,s with $10$\,GB peak GPU memory. Diffusion models are calibrated using $50$ samples, with $50$ and $20$ denoising steps for DiT and PixArt-$\Sigma$, respectively. DiT-XL/2 requires $7$\,min $30$\,s with $14$\,GB peak memory at batch size $32$. 
PixArt-$\Sigma$ at 1K requires $17$\,min with
$24$\,GB peak memory at batch size
16; at 2K, calibration requires
$6$\,h with $38$\,GB peak memory at batch size $5$.
All measurements are reported in an NVIDIA RTX Pro 6000 GPU and include all calibration forward passes and the complete head-selection search.

\paragraph{Inference and computational complexity.}
Given a pretrained vision or diffusion transformer, LoSpAttn
performs a one-time calibration to assign low-rank,
block-sparse, or exact softmax attention to each head,
with timestep-dependent assignments in diffusion models.
For each subsequent input, the heads apply their calibrated
branches to the current queries, keys, and values.
The three branches require $\mathcal O(Nmd+m^3)$,
$\mathcal O(Nwd)$, and $\mathcal O(N^2d)$ time per head,
respectively, where $w$ is the average number of retained
keys per query. 

\section{Experiments}
\label{sec:experiments}
We evaluate LoSpAttn on pretrained ViT-Large for
ImageNet~\cite{deng2009imagenet} classification, DiT-XL/2
for ImageNet class-conditional generation, and
PixArt-$\Sigma$ at 1K/2K for text-to-image generation
using MS-COCO~\cite{lin2014microsoft} captions.
We also benchmark our low-rank approximation on
BigGAN~\cite{brock2018large} using ImageNet, where existing
drop-in methods perform well.
Timings use an NVIDIA RTX PRO 6000 GPU with FP16
unless otherwise specified. Our DiT-XL/2 uses 32-step DDIM with classifier-free guidance
(CFG) scale $1.5$; PixArt-$\Sigma$ uses 50-step DPM-Solver
with CFG scale $4.0$.
We compute Inception Score (IS; mean $\pm$ standard
deviation over 10 splits) ~\cite{salimans2016improved,obukhov2020torchfidelity}
and Fr\'echet Inception Distance (FID)~\cite{heusel2017gans,parmar2022aliased}.

Ablations assess the theoretical error bounds and
head-allocation effects on accuracy, generation quality,
latency, and peak memory. Runtime analyses vary token
count and landmark budget and profile individual components.
Checkpoint identifiers, preprocessing, dataset subsets,
measurement protocols, and further comparisons and
ablations are provided in the appendix.

\begin{table}[t]
    \centering
    \small
    \setlength{\tabcolsep}{2pt}
    \renewcommand{\arraystretch}{1.1}
\caption{PixArt-$\Sigma$ generation at 1K and 2K,
evaluated using 30k and 5k generated images, respectively.
FID uses real MS-COCO reference images.
End-to-end time and peak GPU memory are measured
at batch size eight.}
\label{tab:pixart_results}
    \begin{tabular*}{\linewidth}
        {@{\extracolsep{\fill}}lcccc@{}}
        \toprule
        Method &
        FID $\downarrow$ &
        IS $\uparrow$ &
        \makecell{Inference Time \\(sec) $\downarrow$} &
        \makecell{Peak Memory\\(GB) $\downarrow$} \\
        \midrule

        \multicolumn{5}{l}
        {\textit{PixArt-$\Sigma$, 1K ($N=4096$)}} \\
        \midrule
        Original softmax    & 52.72 & 24.49 & 47.20  & 28.13 \\
        \textbf{LoSpAttn}    & 53.17 & 24.44 & 42.17  & 28.13  \\

        \midrule
        \multicolumn{5}{l}
        {\textit{PixArt-$\Sigma$, 2K ($N=16384$)}} \\
        \midrule
        Original softmax    & 32.43 & 35.14 & 295.68  & 60.17  \\
        \textbf{LoSpAttn}    & 32.98 & 34.97 & 211.27 & 60.17  \\

        \bottomrule
    \end{tabular*}
\end{table}

\begin{figure*}[!t]
    \centering
    \begingroup
    \small
    \setlength{\tabcolsep}{2pt}

    \newcommand{\qualimg}[1]{%
        \raisebox{-0.5\height}{%
            \includegraphics[
                width=\dimexpr(\textwidth-42pt)/6\relax,
                height=\dimexpr(\textwidth-42pt)/6\relax,
                keepaspectratio
            ]{#1}%
        }%
    }

    \newcommand{\qualrowlabel}[1]{%
        \makebox[18pt][c]{%
            \raisebox{-0.5\height}{%
                \rotatebox{90}{%
                    \scriptsize\shortstack{#1}%
                }%
            }%
        }%
    }

    \resizebox{0.95\textwidth}{!}{%
    \begin{tabular}{@{}c*{6}{c}@{}}
        &
        \multicolumn{2}{c}{
            \shortstack{\textbf{PixArt-$\Sigma$ 2K} 
                        $N=16{,}384$}
        } &
        \multicolumn{2}{c}{
            \shortstack{\textbf{PixArt-$\Sigma$ 1K} 
                        $N=4{,}096$}
        } &
        \multicolumn{2}{c}{
            \shortstack{\textbf{DiT-XL/2} 
                        $512\times512$, $N=1{,}024$}
        }
        \\[5pt]

        \qualrowlabel{Native exact\\attention} &
        \qualimg{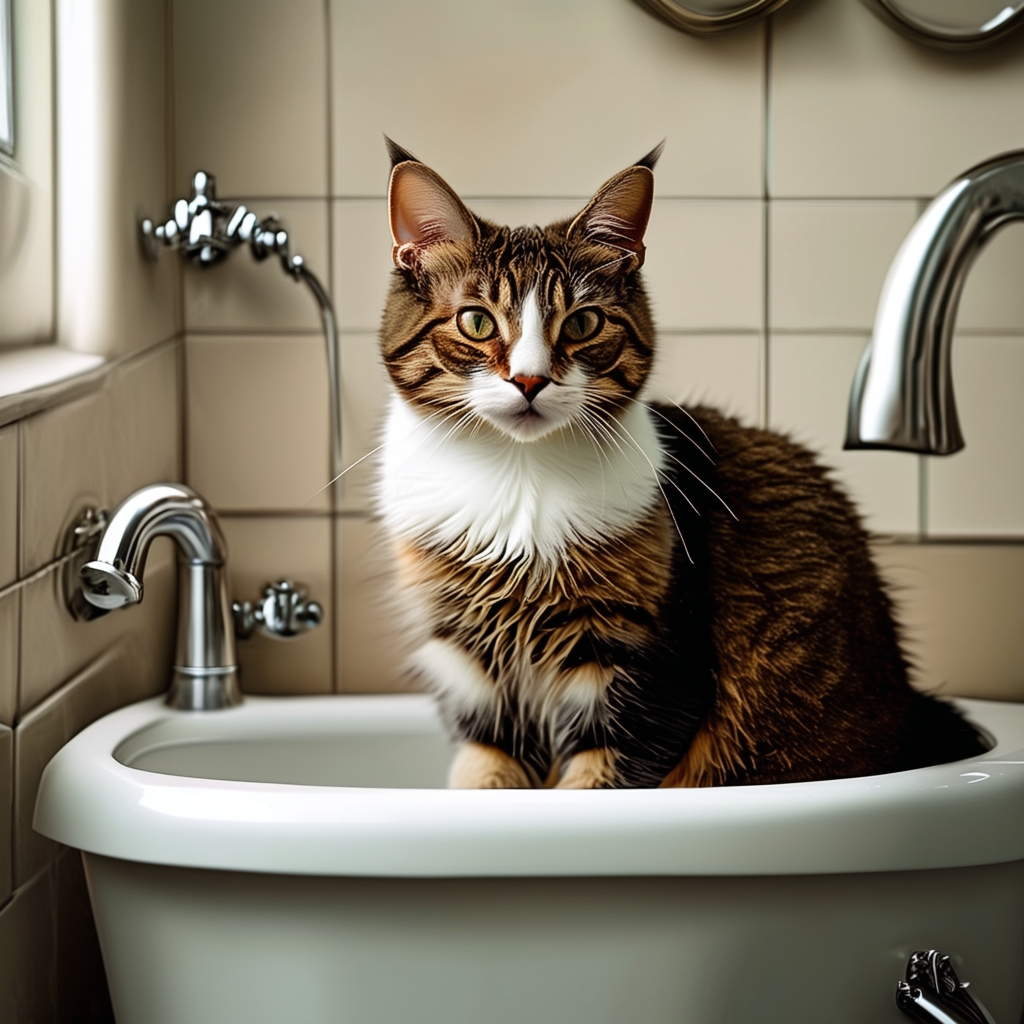} & \qualimg{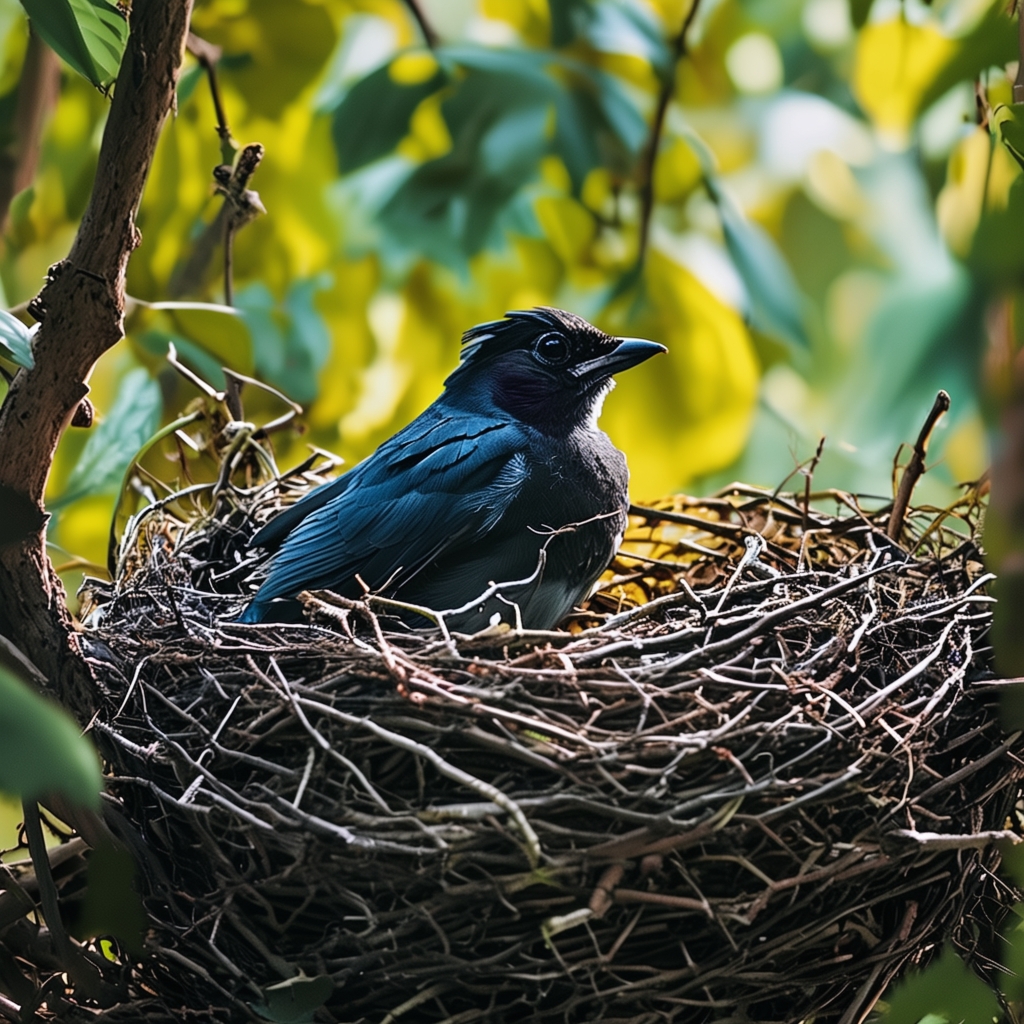} &
        \qualimg{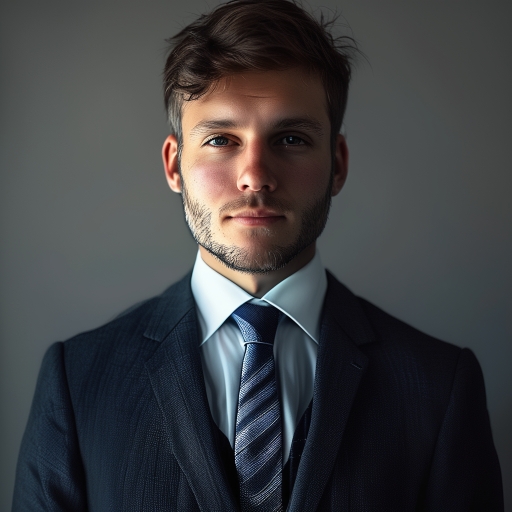} & \qualimg{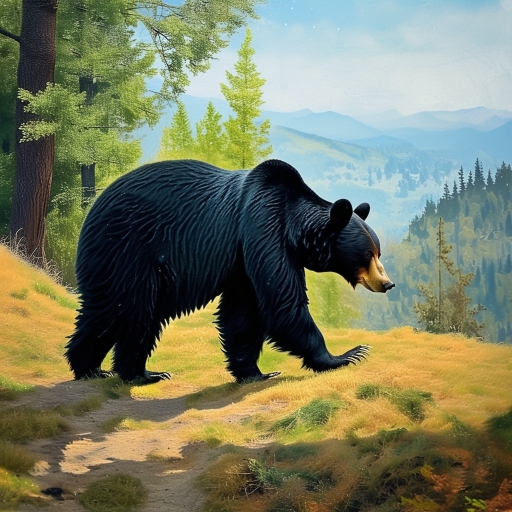} &
        \qualimg{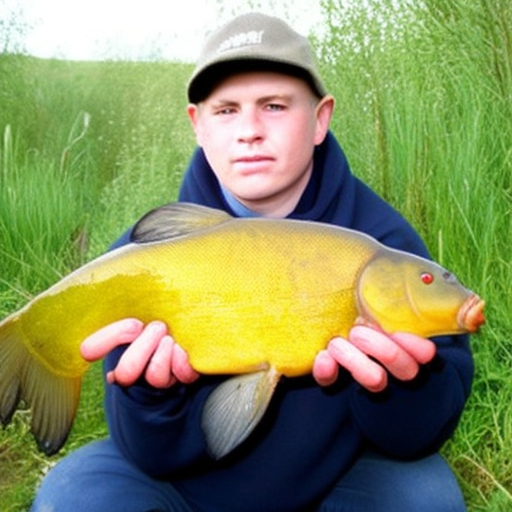} & \qualimg{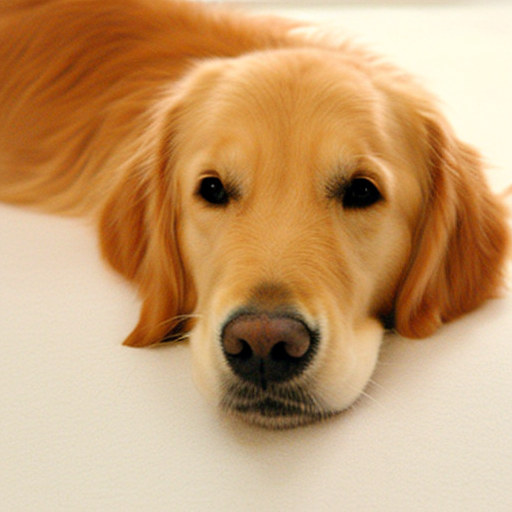}
        \\[4pt]

        \qualrowlabel{MonarchAttn\\First half network} &
        \qualimg{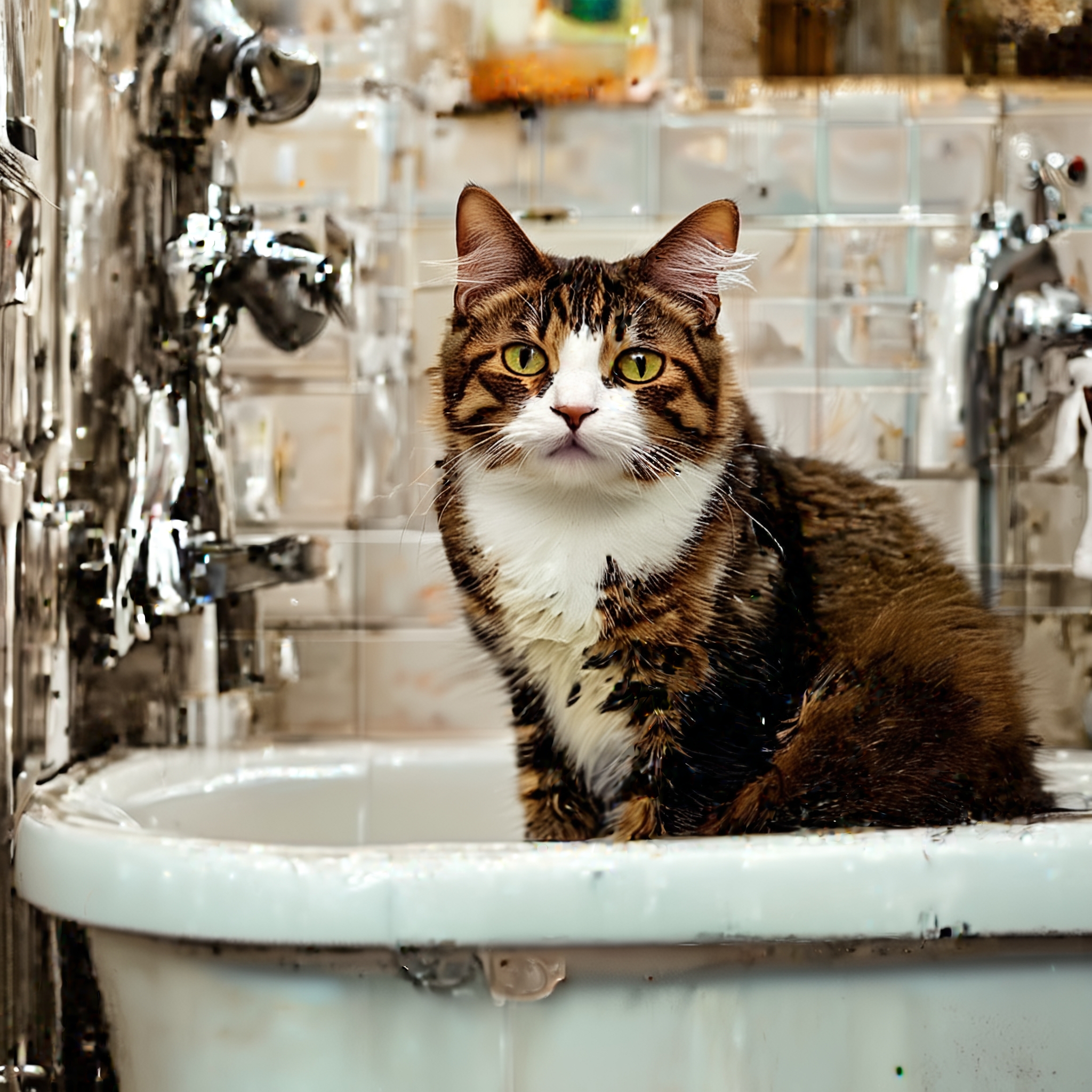} & \qualimg{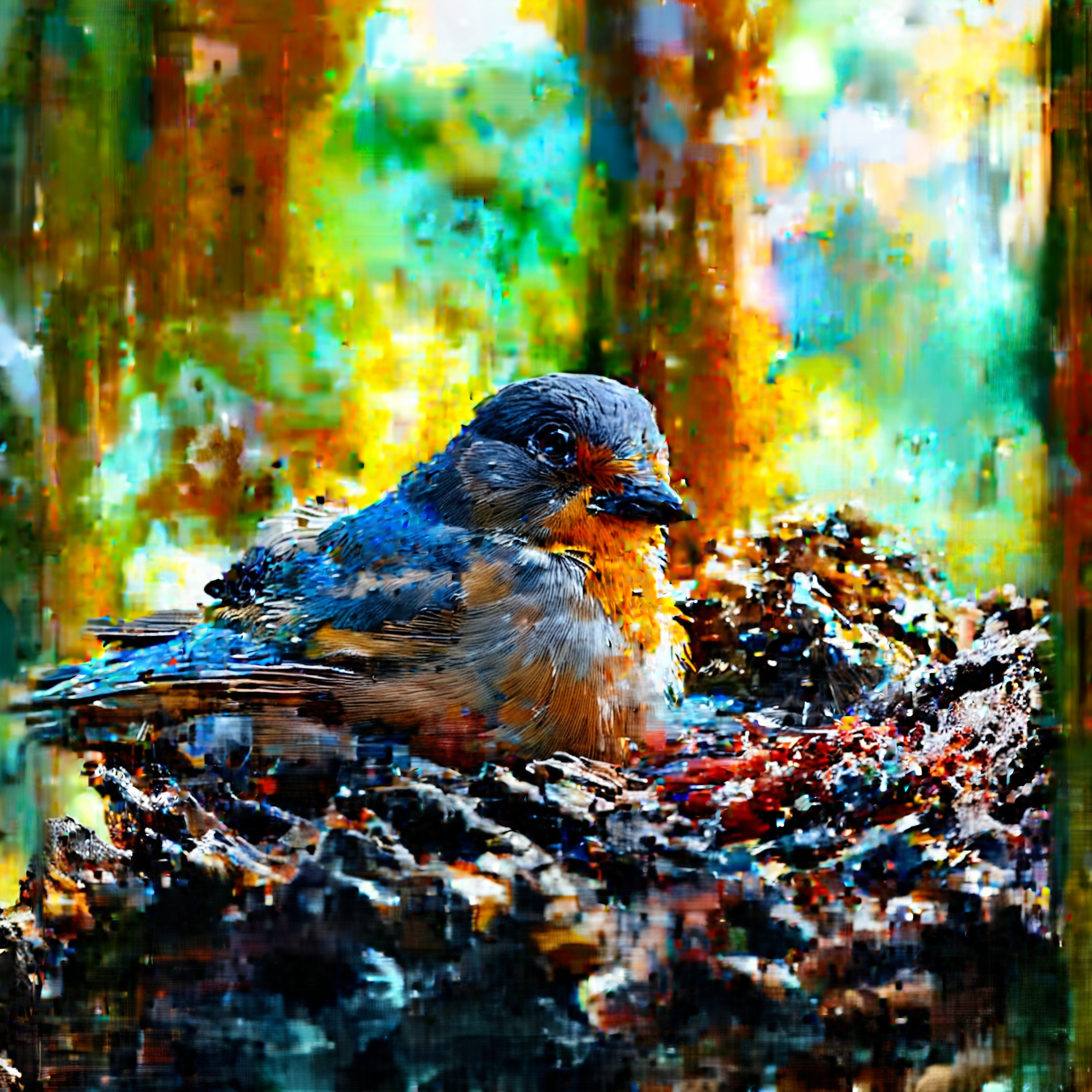} &
        \qualimg{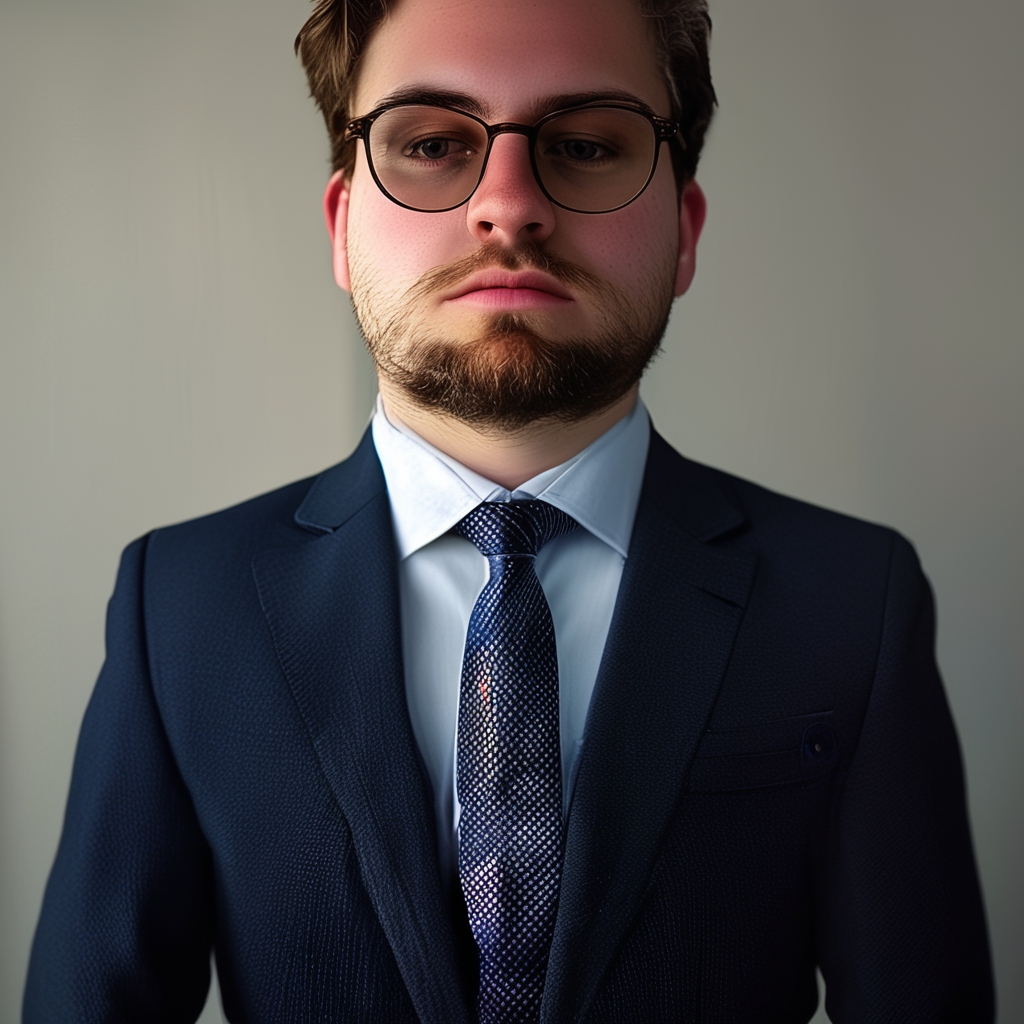} & \qualimg{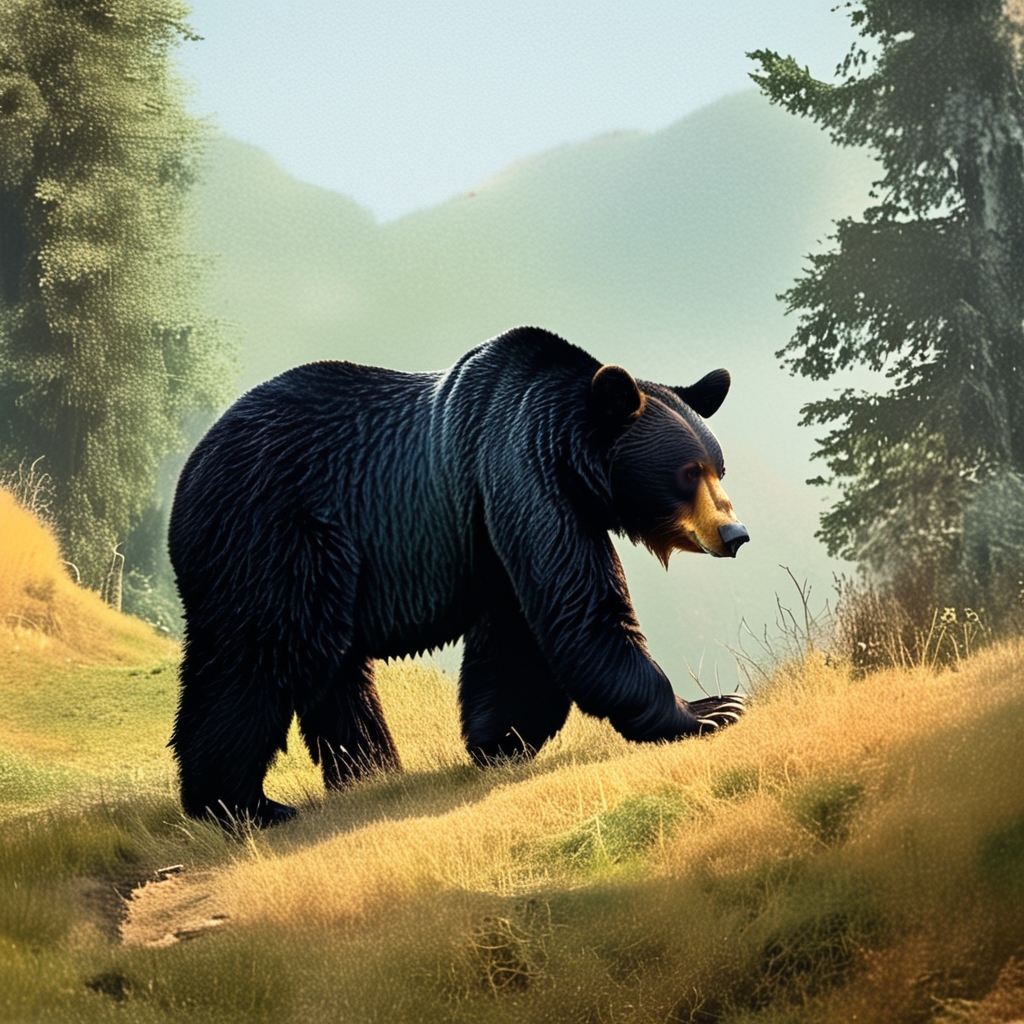} &
        \qualimg{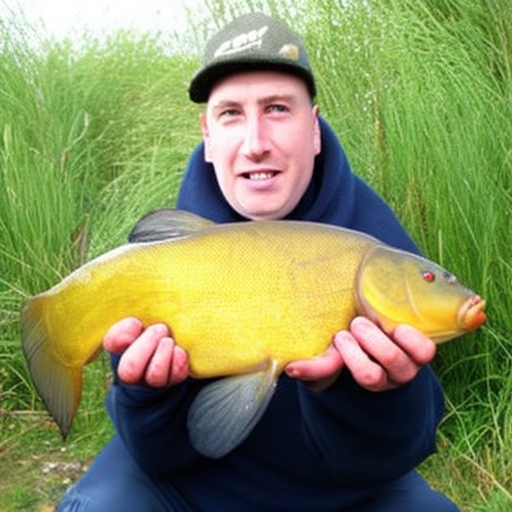} & \qualimg{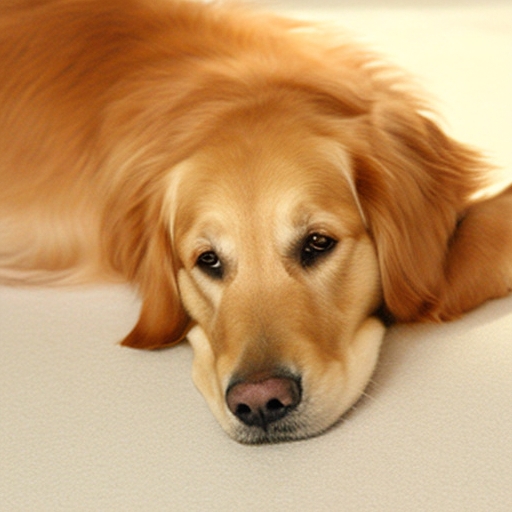}
        \\[4pt]

        \qualrowlabel{DiTFastAttn \\ separate pipeline} &
        \qualimg{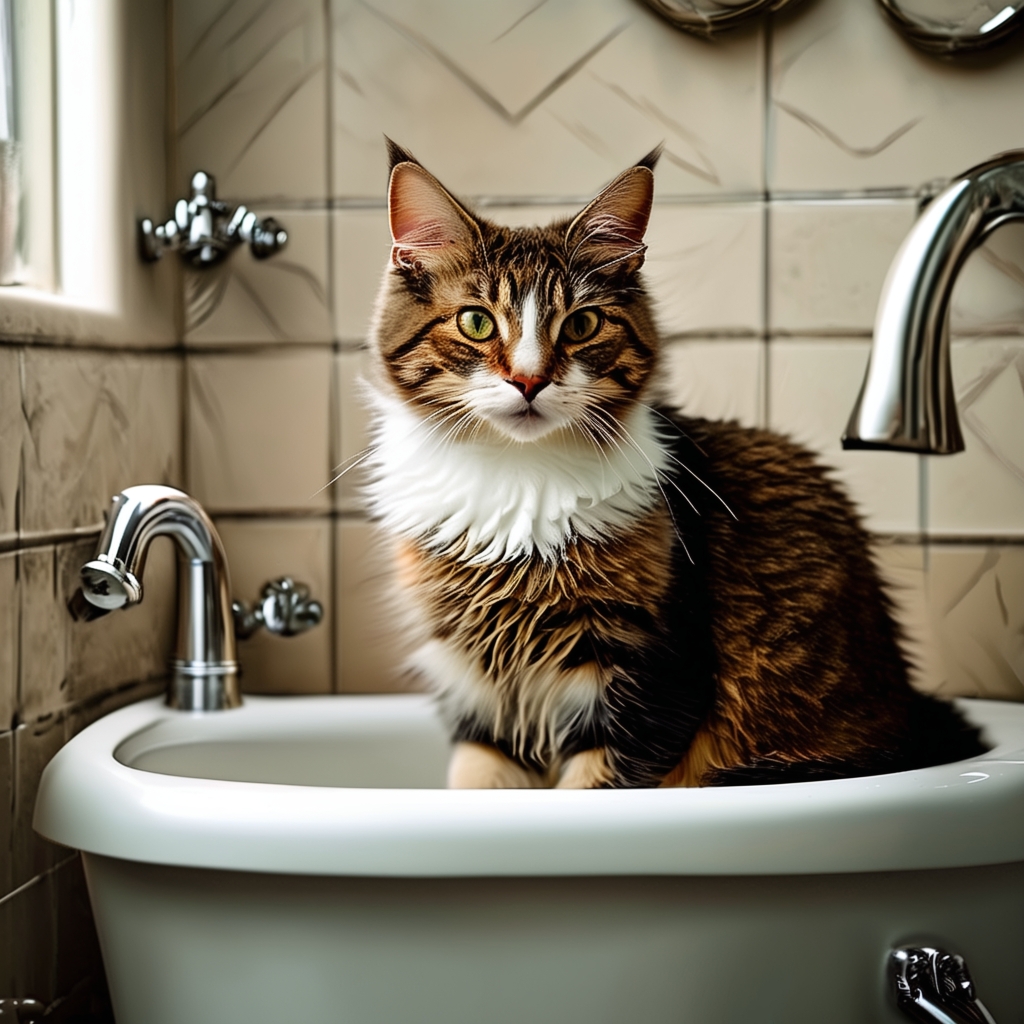} & \qualimg{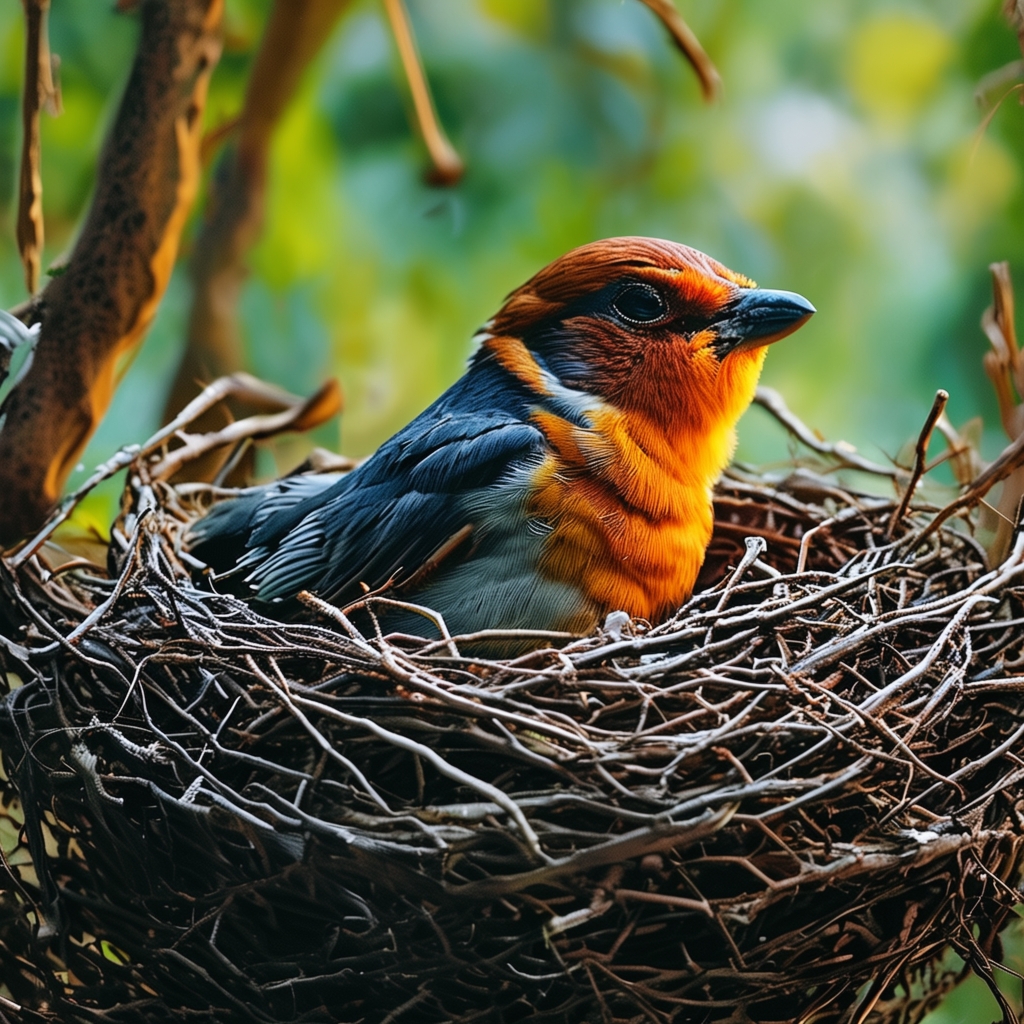} &
        \qualimg{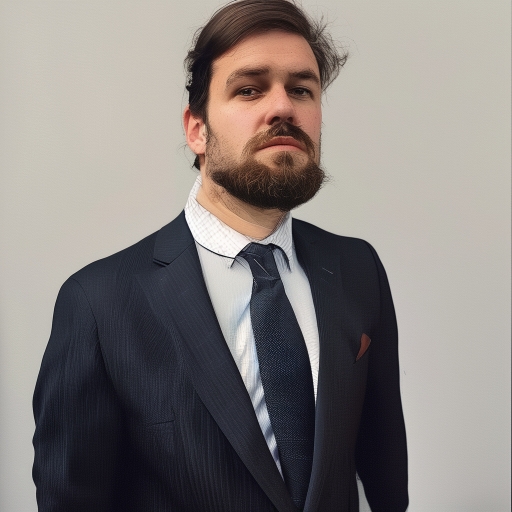} & \qualimg{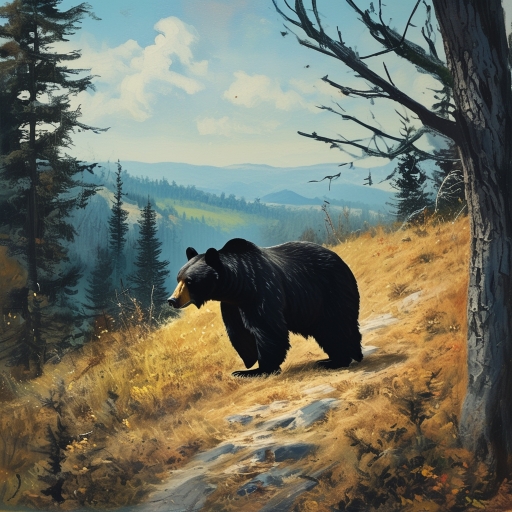} &
        \qualimg{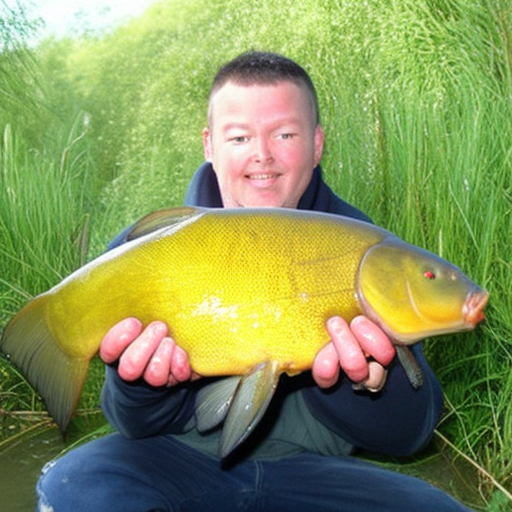} & \qualimg{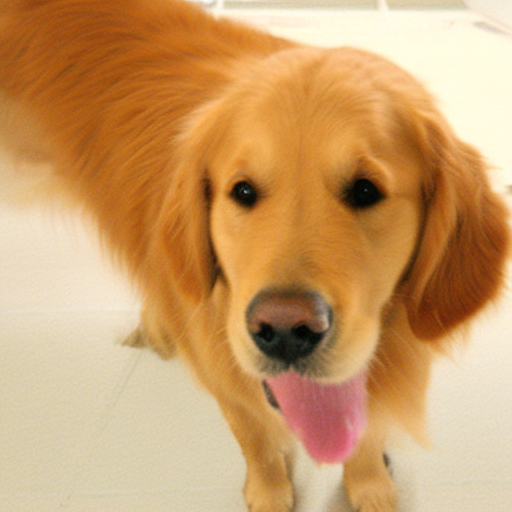}
        \\[4pt]

        \qualrowlabel{\textbf{LoSpAttn}\\Low-Rank} &
        \qualimg{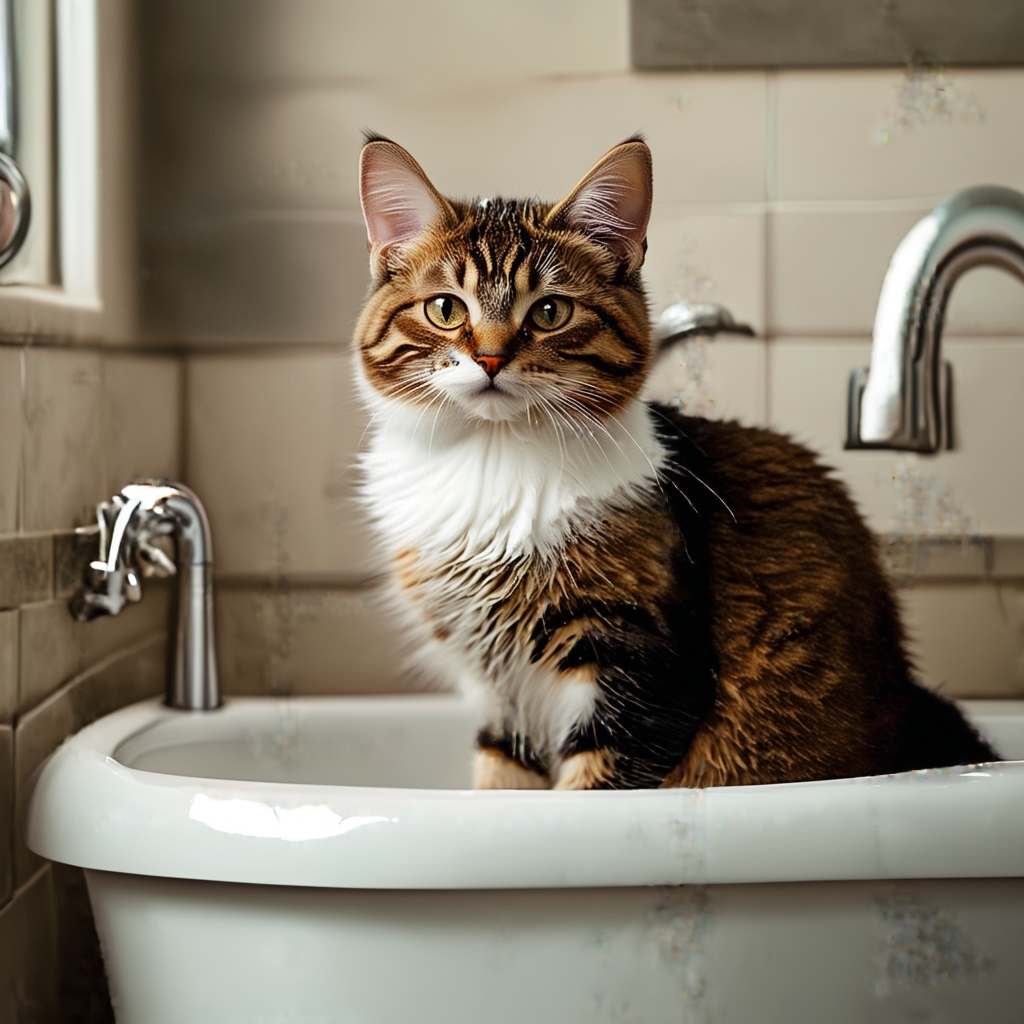} & \qualimg{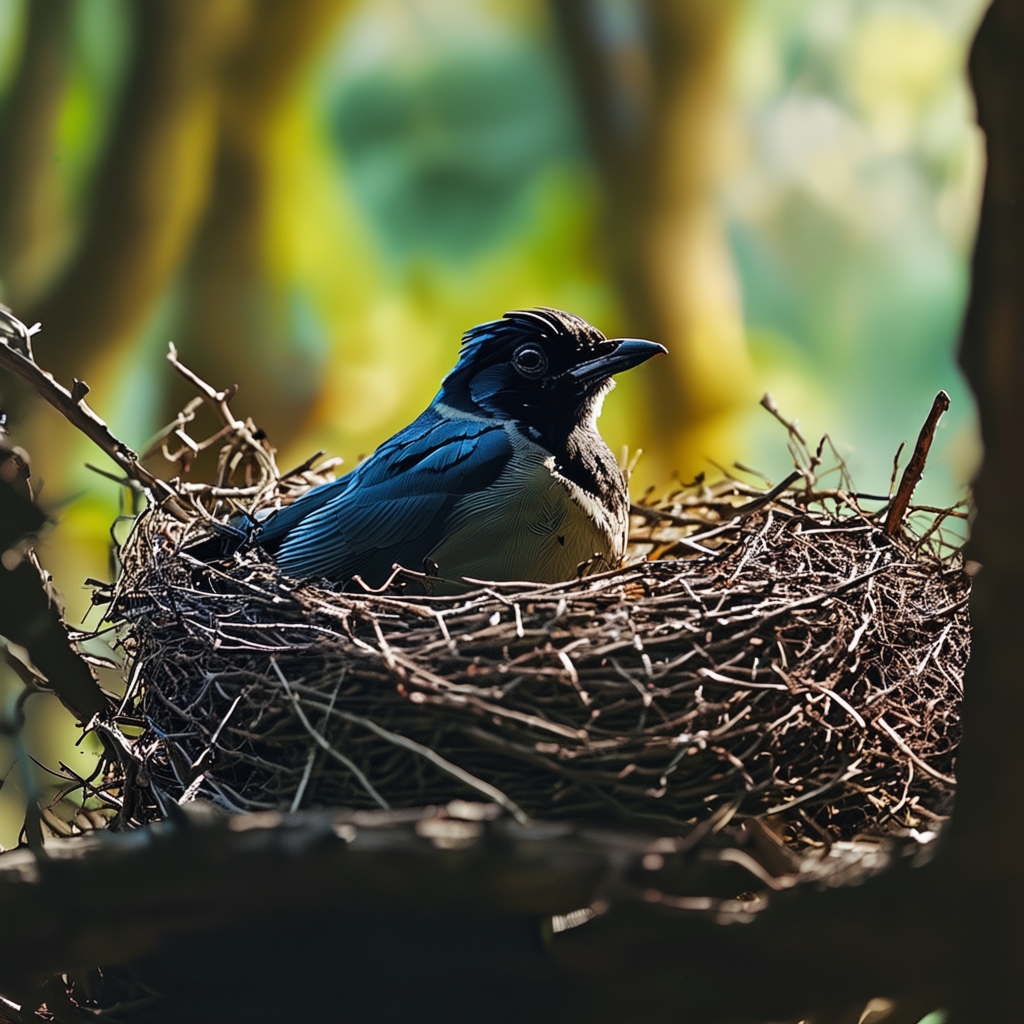} &
        \qualimg{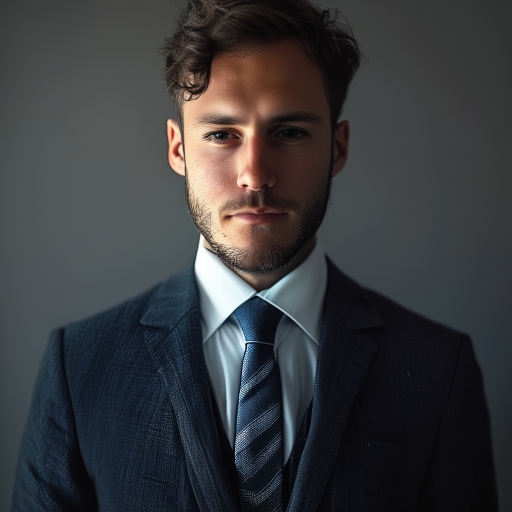} & \qualimg{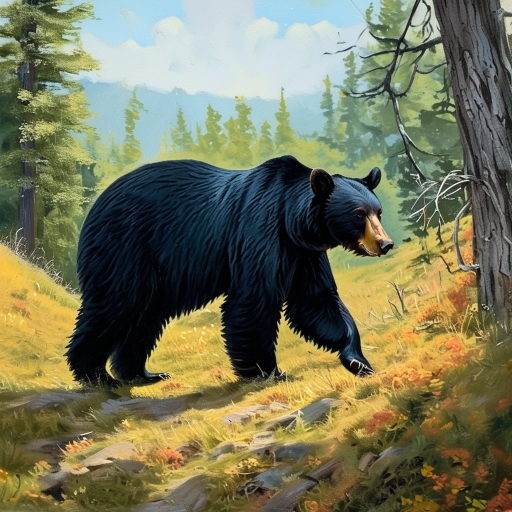} &
        \qualimg{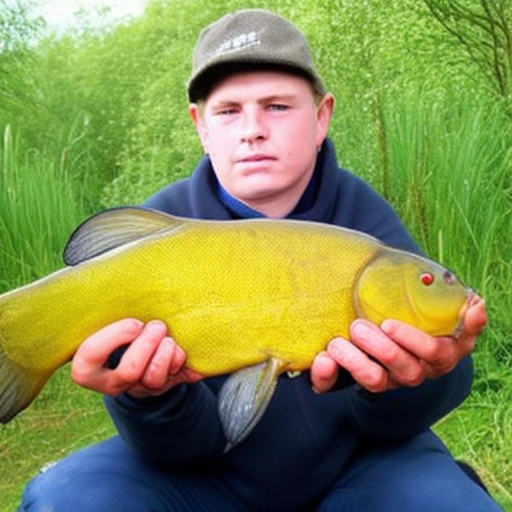} & \qualimg{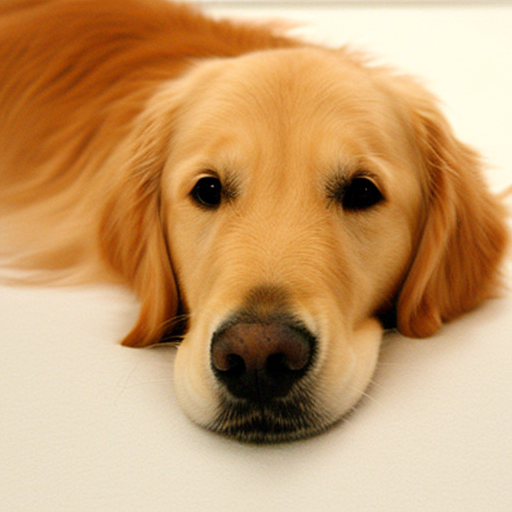}
        \\[4pt]

        \qualrowlabel{\textbf{LoSpAttn}\\Low-Rank + Sparse} &
        \qualimg{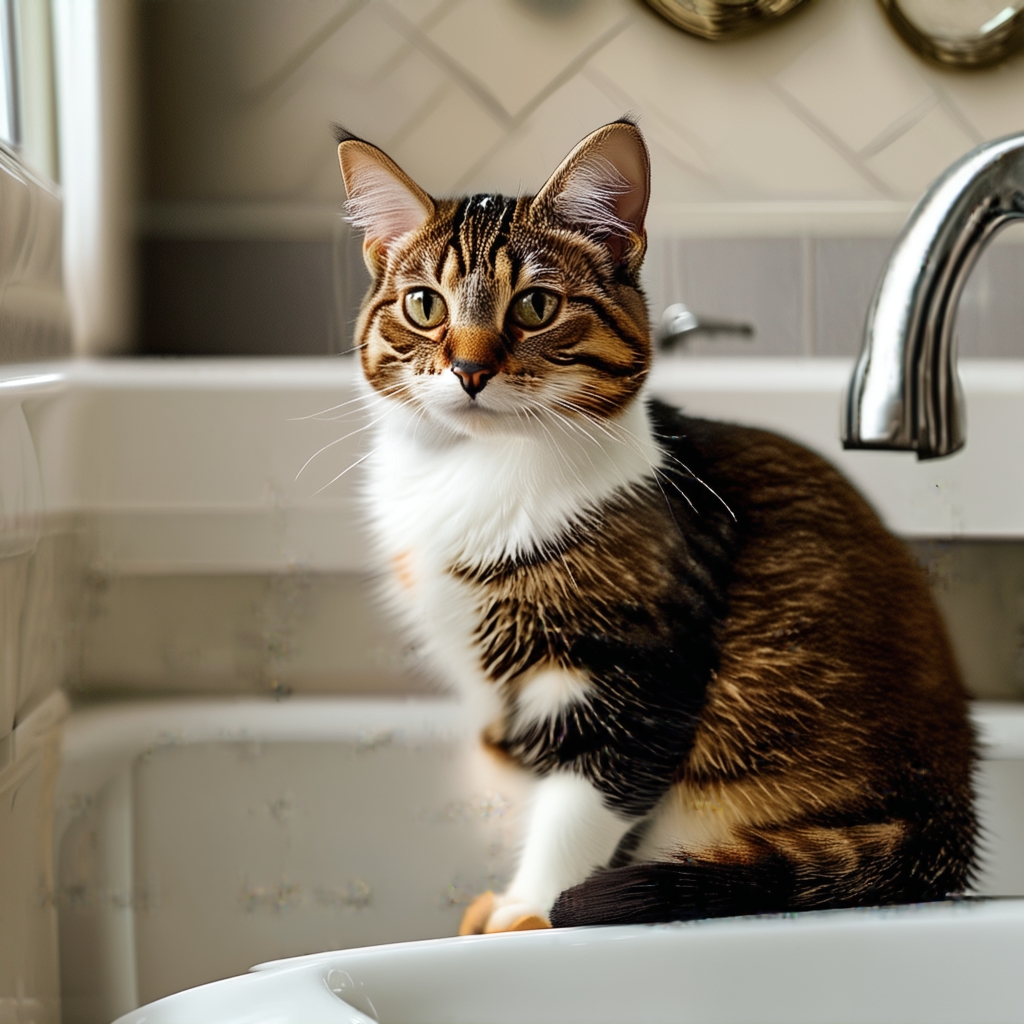} & \qualimg{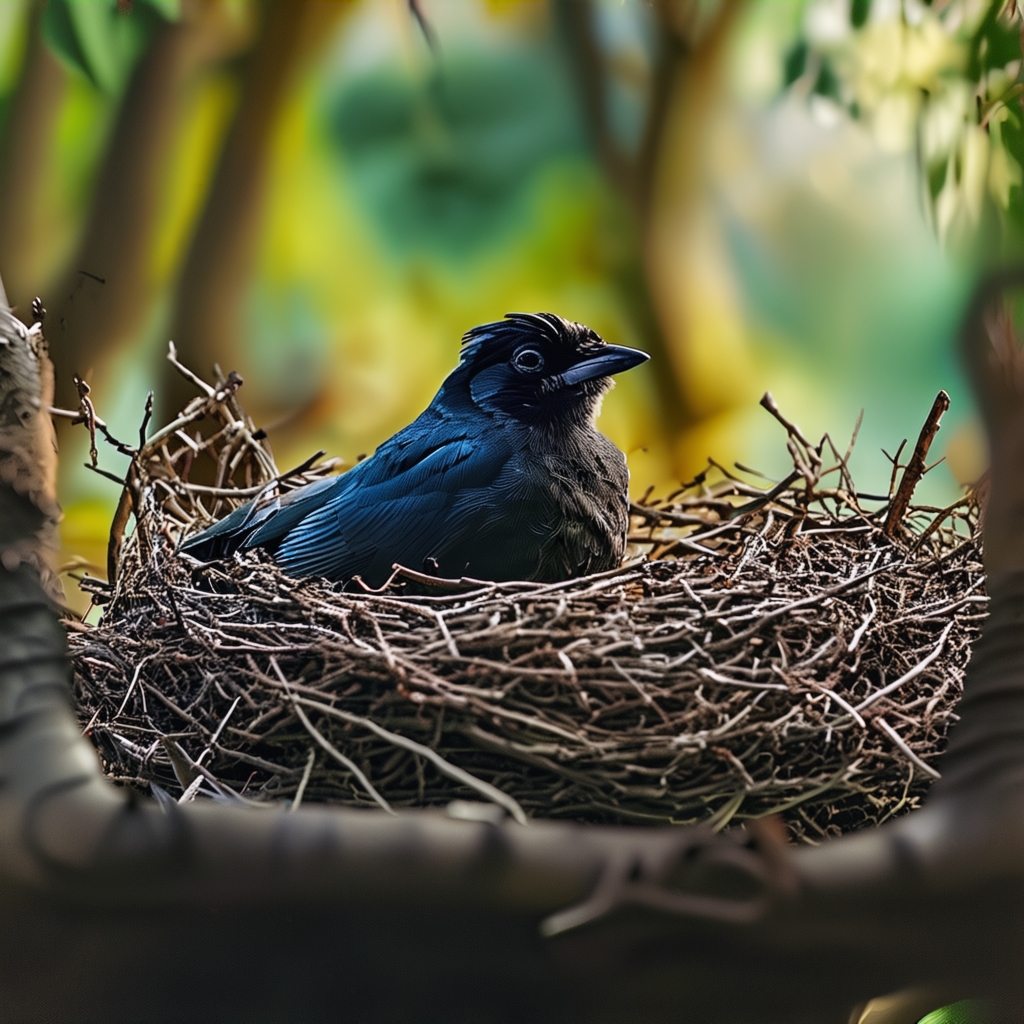} &
        \qualimg{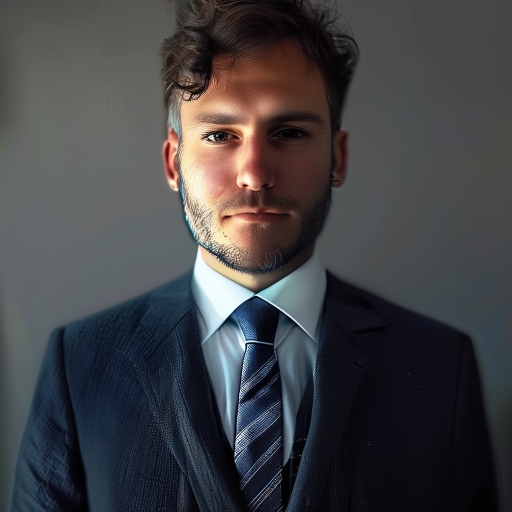} & \qualimg{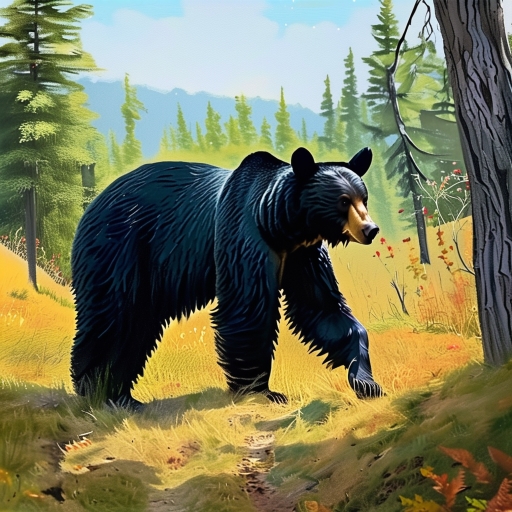} &
        \qualimg{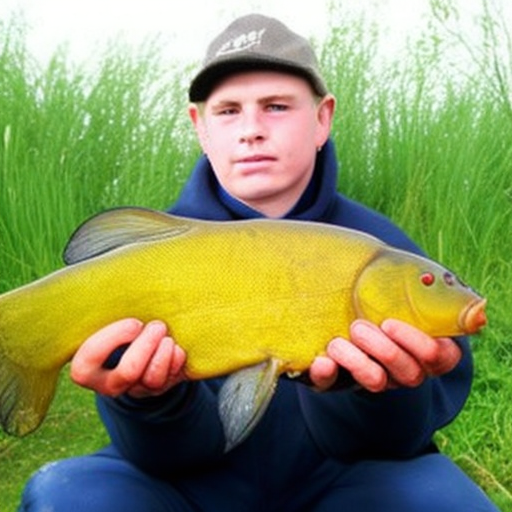} & \qualimg{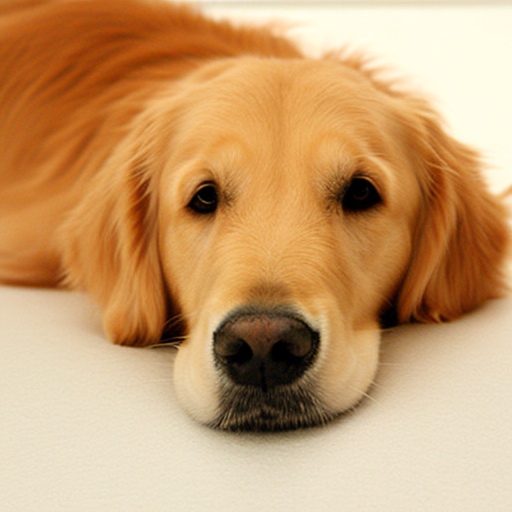}
        \\
    \end{tabular}
    }
    \endgroup

\caption{\textbf{Qualitative comparison across models and
resolutions.}
Two examples each are shown for PixArt-$\Sigma$ at 2K
and 1K and DiT-XL/2 at $512\times512$.
MonarchAttention replaces the first half of attention
layers; DiTFastAttn uses its own inference pipeline.
LoSpAttn is shown with low-rank-only and hybrid replacement.
Hybrid (LR, SP) allocations are $(30\%,45\%)$,
$(30\%,43\%)$, and $(44\%,45\%)$ for the three settings,
respectively, measured over head--timestep pairs;
the remainder uses exact attention.}
    \label{fig:pixart_qualitative}
\end{figure*}

\subsection{Comparison Results}

\noindent \textbf{Text-to-Image Generation.}
Table~\ref{tab:pixart_results} compares LoSpAttn with
the pretrained PixArt-$\Sigma$ models at 1K and 2K resolutions.
LoSpAttn uses $30\%$ low-rank and $43\%$ sparse
head--timestep pairs at 1K, and $30\%$ low-rank and
$45\%$ sparse pairs at 2K; the remainder retains exact
attention. These allocations achieve $1.12\times$ and
$1.40\times$ end-to-end inference speedups, respectively,
with unchanged peak GPU memory. FID increases by $0.45$
at 1K and $0.55$ at 2K, while IS remains nearly unchanged
at 1K. In our evaluation, since exact-attention FID and IS differ between native PixArt-$\Sigma$ and DiTFastAttn,
we report comparisons with matched baselines in the
appendix. The PixArt-$\Sigma$ examples in
Fig.~\ref{fig:pixart_qualitative} shows that our visual quality is 
comparable to FlashAttention-2 at both 1K and 2K.

\noindent \textbf{Class-Conditional Image Generation.}
Table~\ref{tab:dit_xl_512} evaluates DiT-XL/2 at
$512\times512$, using $44\%$ low-rank, $45\%$ sparse,
and $11\%$ exact head--timestep pairs.
Efficiency is measured relative to each framework's exact-attention baseline: SDPA in FP32 and
FlashAttention-2 in FP16. In FP16, LoSpAttn achieves the lowest
$\mathrm{FID}_{\mathrm{ref}}$ among the reported
approximations ($6.54$) and an IS of $158.23\pm6.78$,
compared to $157.93\pm7.09$ for exact attention.
LoSpAttn provides a $1.024\times$ speedup with
$3.9\%$ higher peak memory.
DiTFastAttn achieves $1.049\times$ speedup while using
$192.3\%$ of its baseline memory.
Figure~\ref{fig:pixart_qualitative} illustrates comparable
visual quality with $89\%$ of head--timestep pairs replaced. 
We also evaluate BigGAN, where several existing
drop-in approximations preserve generation quality.
As shown in Table~\ref{tab:biggan}, our low-rank approximation
from Section~\ref{sec:symm_approx} achieves competitive
quality, with FID of $30.74$ and an IS of
$56.17\pm3.28$. Replacing exact attention reduces
the attention-layer cost from $10.737$ to $1.993$ GFLOPs,
a reduction of $81.4\%$.

\noindent \textbf{Image Classification.}
On ViT-Large (Table~\ref{tab:vit_results}), $50\%$
low-rank replacement achieves the highest accuracy
among evaluated approximations ($86.722\%$ top-1,
$98.214\%$ top-5) at $36.5$\,ms.
Adding $10\%$ sparse heads yields $86.530\%$ top-1
accuracy at $37.8$\,ms. Both configurations are more
accurate but slower than MonarchAttention.
The $(65,5,30)$ allocation achieves $85.930\%$ top-1
accuracy with marginal acceleration ($35.3$ versus
$35.7$\,ms for fused exact attention).
Peak memory remains comparable to exact attention.

\begin{table}[t]
    \centering
\caption{DiT-XL/2 generation at $512\times512$.
$\mathrm{FID}_{\mathrm{ref}}$ compares
$10k$ generated images with exact-attention reference samples.
Speedup and peak memory at batch size eight
are relative to each method's matched exact-attention
baseline, separately for FP32 and FP16.}

    \label{tab:dit_xl_512}

    \setlength{\tabcolsep}{3pt}
    \resizebox{\linewidth}{!}{%
    \begin{tabular}{lcccc}
        \toprule
        Method &
        $\mathrm{FID}_{\mathrm{ref}}$ $\downarrow$ &
        IS $\uparrow$ &
        \makecell{Speedup\\($\times$) $\uparrow$} &
        \makecell{Peak memory\\(\% of baseline) $\downarrow$} \\
        \midrule

        \multicolumn{5}{l}{\textit{FP32}} \\
        \midrule
        Exact (fused)
        & -- & $158.05\pm6.84$ & 1.000 & 100.0 \\
        SpargeAttention
        & 79.04 & $22.18\pm0.47$ & 0.885 & 130.3 \\
        Scatterbrain
        & 227.04 & $2.58\pm0.04$ & 0.963 & 127.8 \\
        KDEformer
        & 264.01 & $2.98\pm0.05$ & 0.934 & 128.6 \\
        Thinformer
        & 309.50 & $2.79\pm0.03$ & 0.876 & 107.2 \\
        WildCat
        & 343.92 & $1.58\pm0.01$ & 0.463 & 110.7 \\
        MonarchAttention
        & 12.01 & $111.39\pm3.69$ & 1.066 & 101.9 \\
        LoSpAttn
        & 7.07
        & $155.96\pm7.89$
        & 1.031
        & 90.72 \\

        \midrule
        \multicolumn{5}{l}
        {\textit{FP16: method-specific frameworks}} \\
        \midrule
        Exact (matched baseline)
        & -- & $157.93\pm7.09$ & 1.000 & 100.0 \\
        MonarchAttention
        & 12.14 & $112.28\pm2.44$ & 0.933 & 102.0 \\
        DiTFastAttn
        & 7.42 & $159.35\pm6.11$ & 1.049 & 192.3 \\
        LoSpAttn 
        & \textbf{6.54} & $158.23\pm6.78$
        & 1.024 & 103.9 \\

        \bottomrule
    \end{tabular}%
    }
\end{table}

\begin{table}[t]
    \centering
    \small
    \setlength{\tabcolsep}{3pt}
    \renewcommand{\arraystretch}{1.08}
\caption{BigGAN generation quality on
5k generated images, with FID computed
against exact real ImageNet-1K reference images.
GFLOPs cover only the self-attention layer.}
    
    \label{tab:biggan}

    \begin{tabular*}{\linewidth}
        {@{\extracolsep{\fill}}lrrr@{}}
        \toprule
        Method & FID $\downarrow$ & IS $\uparrow$
        & GFLOPs $\downarrow$ \\
        \midrule
        Exact attention
        & 32.18 & $58.36\pm4.21$ & 10.737 \\
        \midrule
        SpargeAttention
        & 30.60 & $56.60\pm5.21$ & 9.351 \\
        Scatterbrain
        & 40.29 & $35.15\pm3.90$ & 3.164 \\
        KDEformer
        & 30.51 & $57.39\pm3.97$ & 2.596 \\
        Thinformer
        & 30.47 & $56.75\pm3.75$ & 1.846 \\
        WildCat
        & 31.88 & $57.57\pm3.87$ & 1.043 \\
        MonarchAttention
        & 31.61 & $57.72\pm4.65$ & 1.275 \\
        \midrule
        \textbf{LoSpAttn}
        & 30.74 & $56.17\pm3.28$ & 1.993 \\
        \bottomrule
    \end{tabular*}
\end{table}

\begin{table}[t]
    \centering
    \small
    \setlength{\tabcolsep}{2pt}
    \renewcommand{\arraystretch}{1.1}
\caption{ViT-Large/16 on the full 50k-image ImageNet-1K
validation set. Network latency and peak GPU memory
are measured for a forward pass at batch size eight.
LoSpAttn tuples denote percentages of low-rank,
sparse, and exact softmax heads.}
    \label{tab:vit_results}

    \begin{tabular*}{\linewidth}
        {@{\extracolsep{\fill}}lrrrr@{}}
        \toprule
        Method &
        \makecell{Top-1\\(\%) $\uparrow$} &
        \makecell{Top-5\\(\%) $\uparrow$} &
        \makecell{Network\\latency\\(ms) $\downarrow$} &
        \makecell{Peak GPU\\memory\\(MB) $\downarrow$} \\
        \midrule

        Exact (fused)
        & 87.096 & 98.308 & 35.7 & 1425.6 \\

        \midrule
        SpargeAttention
        & 81.042 & 95.200 & 42.7 & 1592.0 \\
        Scatterbrain
        & 80.074 & 95.130 & 42.6 & 1767.4 \\
        KDEformer
        & 81.682 & 95.794 & 45.5 & 1660.7 \\
        Thinformer
        & 74.680 & 91.738 & 70.2 & 1437.6 \\
        MonarchAttention
        & 86.394 & 98.138 & 35.0 & 1446.4 \\

        \midrule
        {LoSpAttn $(50,0,50)$}
        & 86.722 & 98.214 & 36.5 & 1425.9 \\
        {LoSpAttn $(50,10,40)$}
        & 86.530 & 98.150 & 37.8 & 1402.4 \\
        LoSpAttn $(65,5,30)$
        & 85.930 & 97.962 & 35.3 & 1402.8 \\
        \bottomrule
    \end{tabular*}
\end{table}

\begin{figure}[t]
    \centering
    \captionsetup[subfigure]{
        font=small,
        justification=centering,
        skip=3pt
    }
    \begin{subfigure}[t]{0.23\textwidth}
        \centering
            \includegraphics[
                width=\linewidth,
                height=0.6\linewidth
            ]{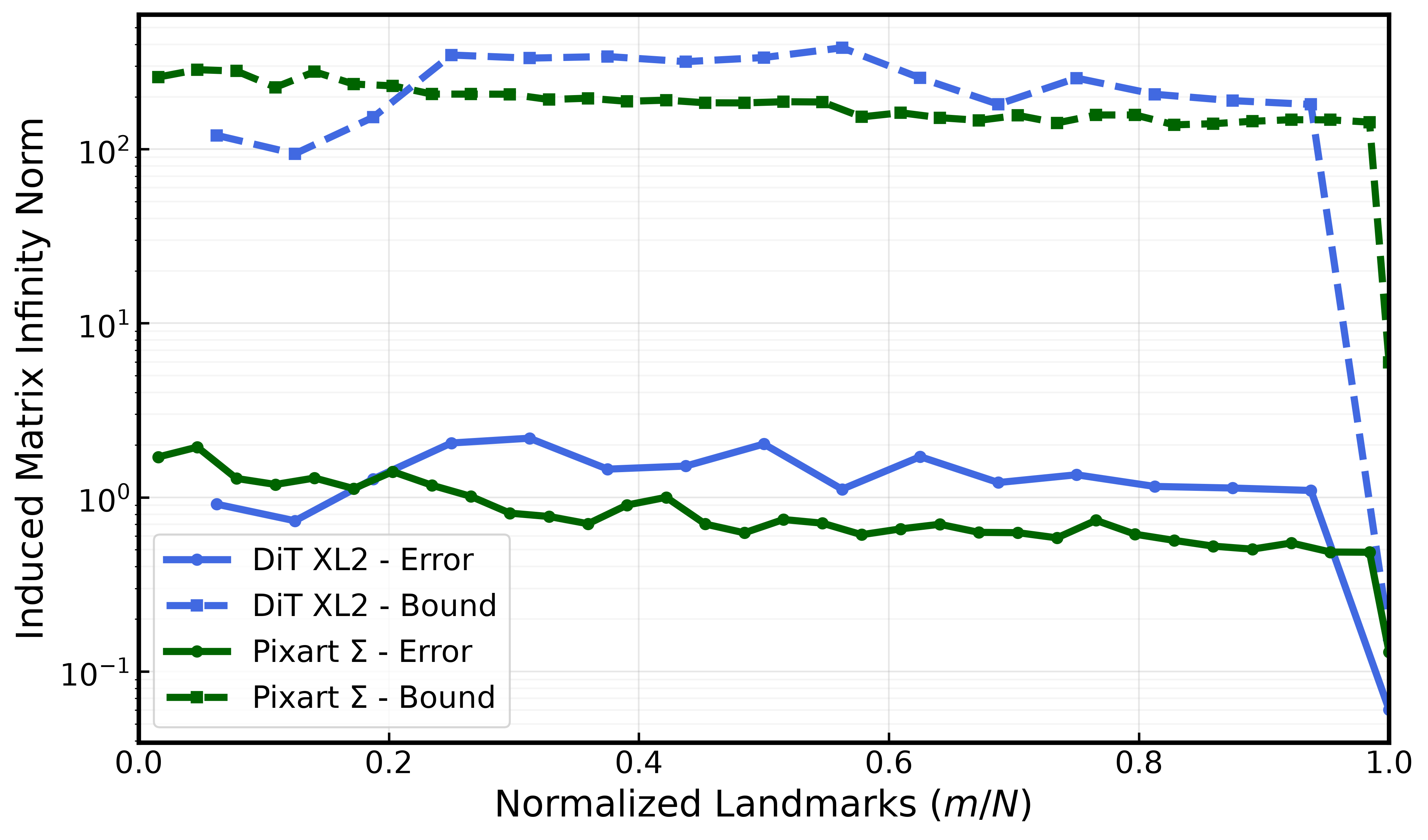}
        \caption{Error plots - DiT and Pixart}
        \label{fig:error_dit}
    \end{subfigure}
    \begin{subfigure}[t]{0.23\textwidth}
        \centering
            \includegraphics[
                width=\linewidth,
                height=0.6\linewidth
            ]{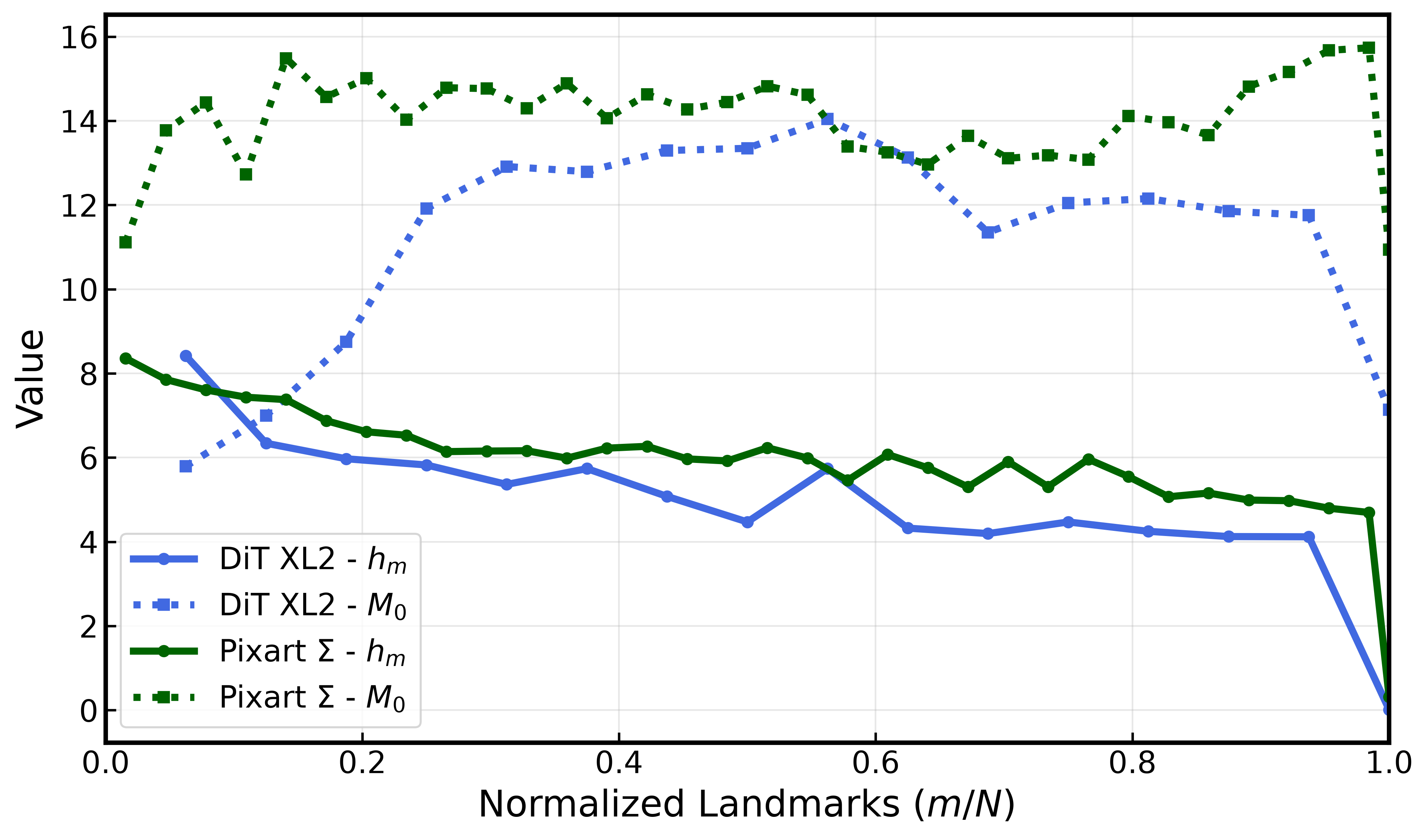}
        \caption{Corresponding $h_m$ and $M_0$}
        \label{fig:error_pixart}
    \end{subfigure}

\caption{\textbf{Theorem.~\ref{thm:approximation_error}}
Averages over five randomly sampled heads per model:
DiT-XL/2 ($512^2$) and PixArt-$\Sigma$ (1K).
(a) Approximation errors and bounds (log scale).
(b) Landmark coverage $h_m$ and stability factor $M_0$.
Both panels vary $m/N$.}
\label{fig:error_bound}
\end{figure}

\subsection{Ablation study}

\begin{figure*}[t]
    \centering

    \begin{subfigure}[t]{0.325\textwidth}
        \centering
        \includegraphics[width=\linewidth]
        {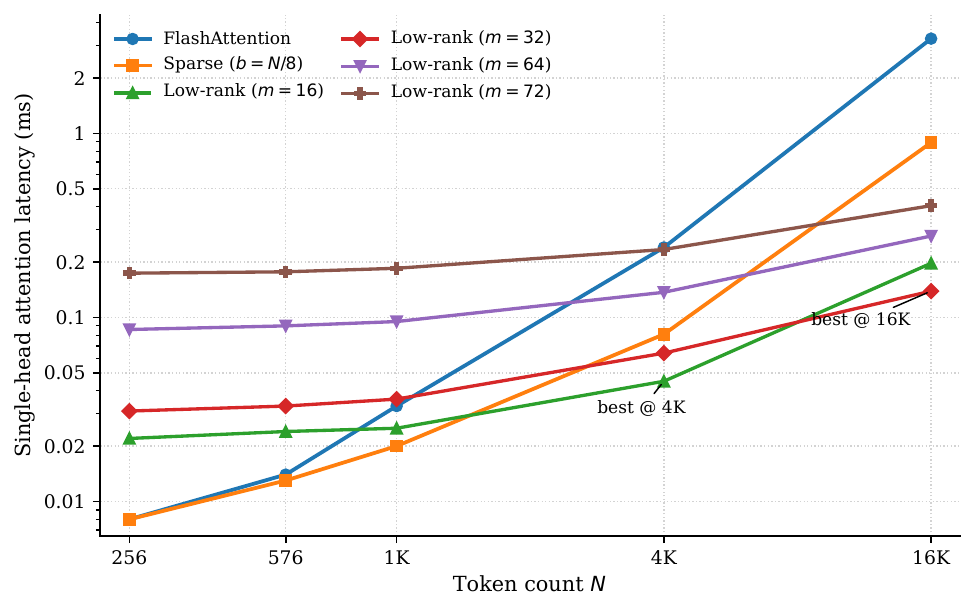}
        \caption{Single-head attention latency.}
        \label{fig:attention_runtime}
    \end{subfigure}
    \hfill
    \begin{subfigure}[t]{0.325\textwidth}
        \centering
        \includegraphics[width=\linewidth]
        {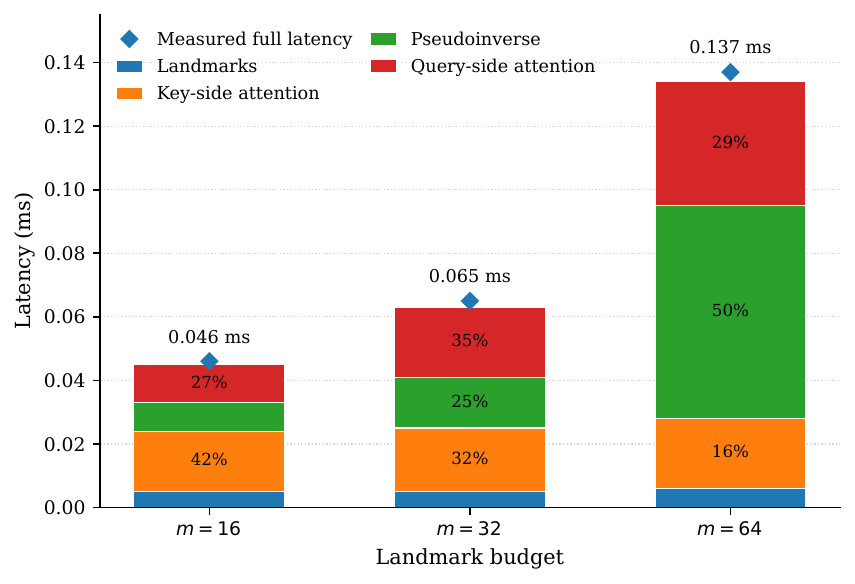}
        \caption{Low-rank runtime breakdown.}
        \label{fig:lowrank_breakdown}
    \end{subfigure}
    \hfill
    \begin{subfigure}[t]{0.325\textwidth}
        \centering
        \includegraphics[width=\linewidth]
        {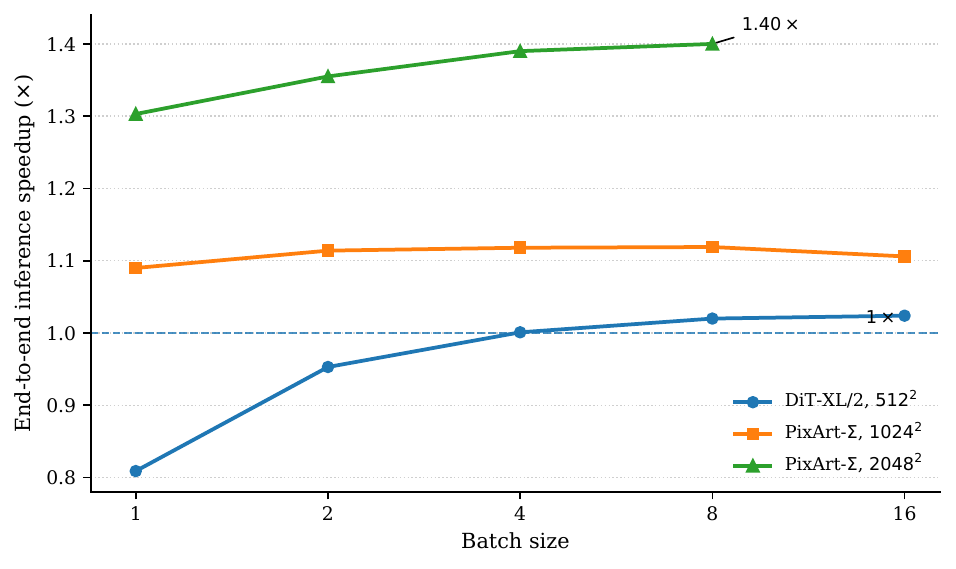}
        \caption{End-to-end speedup across batch sizes.}
        \label{fig:batch_speedup}
    \end{subfigure}
\caption{\textbf{Runtime analysis.}
\textbf{(a)} Single-head latency across token counts:
low-rank attention offers larger gains over FlashAttention
at longer sequences.
\textbf{(b)} Low-rank runtime breakdown at $N=4096$:
pseudoinversion dominates at larger landmark budgets.
\textbf{(c)} End-to-end speedup across batch sizes,
reaching $1.40\times$ on PixArt-$\Sigma$ at 2K resolution
and batch size eight.}
\label{fig:runtime_analysis}
\end{figure*}

\begin{table}[t]
    \centering
    \small
    \setlength{\tabcolsep}{4pt}
    \renewcommand{\arraystretch}{1.08}

\caption{ViT-Large ablations over head allocation and low-rank replacement. LR, SP, and SM denote low-rank, sparse, and softmax attention; LR methods are compared at a fixed allocation. Timings are reported at a batch-size $64$ on RTX 4090 GPU.}
    \label{tab:vit_head_allocation}

    \begin{tabular*}{\linewidth}
        {@{\extracolsep{\fill}}lrrr@{}}
        \toprule
        Configuration &
        Top-1 $\uparrow$ &
        Top-5 $\uparrow$ &
        \makecell{Latency\\(ms) $\downarrow$} \\
        \midrule

        \multicolumn{4}{l}{
        \textit{(a) Head allocation (LR, SP, SM) \%}
        } \\
        \addlinespace[1pt]



        Exact $(0,0,100)$
        & 87.090 & 98.300 & 505.01 \\

        $(0,100,0)$
        & 68.170 & 87.052 & 509.85 \\

        $(100,0,0)$
        & 22.540 & 38.342 & 429.98 \\

        \midrule

        $(26.56,6.00,67.45)$
        & 86.870 & 98.270 & 487.69 \\

        $(41.67,10.42,47.92)$
        & 86.630 & 98.210 & 476.80 \\

        $(55.73,17.19,27.08)$
        & 86.210 & 97.990 & 468.76 \\

        \midrule

        \multicolumn{4}{l}{
        \textit{(b) LR approximation at
        $(41.67,10.42,47.92)$}
        } \\
        \addlinespace[1pt]

        Ours
        & 86.630 & 98.210 & 476.80 \\

        Nystr\"omformer
        & 86.630 & 98.142 & 495.38 \\

        Performer
        & 76.794 & 93.306 & 534.02 \\

        WildCat
        & 86.732 & 98.156 & 640.73 \\

        Ours w/o precond.
        & 64.418 & 79.394 & 480.66 \\

        \bottomrule
    \end{tabular*}
\end{table}

\noindent \textbf{ViT-Large Ablations.}
Table~\ref{tab:vit_head_allocation} examines the allocation of the heads and the choice of the low-rank approximation. 
Uniform low-rank or sparse replacement substantially
degrades accuracy, whereas hybrid allocations retain
$86.21\%$--$86.87\%$ top-1 accuracy.
Increasing replacement coverage reduces hybrid latency
from $487.69$ to $468.76$\,ms, approaching the original
baseline ($505.01$\,ms).
In the fixed allocation $(41.67,10.42,47.92)$, our
approximation achieves the lowest latency among the
 low-rank methods tested, with $86.63\%$ top-1 accuracy;
WildCat is slightly more accurate but slower.
Removing preconditioning reduces top-1 accuracy to
$64.42\%$ due to numerical instability, demonstrating the importance of preconditioning.

\noindent \textbf{Head Allocation in Diffusion Transformers.}
Table~\ref{tab:dit_head_allocation} evaluates DiT-XL/2
allocations on head--timestep pairs.
Uniform low-rank and sparse replacement yield FID scores
of $224.00$ and $113.62$, respectively, whereas hybrid
allocations achieve $5.63$--$10.00$.
Increasing replacement coverage reduces hybrid latency
from $1201.67$ to $1184.47$\,ms but increases FID,
revealing an accuracy--efficiency trade-off.
The fastest hybrid provides a modest $1.9\%$ latency
reduction relative to the exact attention 
($1207.22$\,ms).

\noindent \textbf{Empirical Error Bounds.}
Figure~\ref{fig:error_bound} evaluates the bound in 
Theorem~\ref{thm:approximation_error} across landmark
budgets. Mean approximation errors remain below the
corresponding mean bounds, although the bounds are
conservative. Increasing the landmark budget generally
reduces the coverage error $h_m$, while the variation in
the stability factor $M_0$ means that the bound does not need
to decrease monotonically.

\noindent \textbf{Attention Runtime.}
Figure~\ref{fig:attention_runtime} compares the
single-head latency of exact, sparse, and low-rank
attention branches across token counts. At
$N=16{,}384$, low-rank attention with $m=32$ reduces
latency from $3.274$ to $0.139$\,ms, corresponding to a
$23.6\times$ speedup over FlashAttention-2, while sparse
attention achieves a $3.7\times$ speedup. Low-rank attention offers increasing speedups over FlashAttention-2 at longer sequences but remains slower at $N=256$ and $576$ due to landmark construction and inversion overheads.

\noindent \textbf{Low-Rank Runtime Breakdown.}
Figure~\ref{fig:lowrank_breakdown} decomposes the
low-rank branch in $N=4096$ into different steps in Algorithm.~\ref{alg:pnpnystra}. Increasing the landmark budget
from $m=16$ to $m=64$ increases the attention
latency from $0.046$ to $0.137$\,ms. At $m=16$,
key-side attention (computing $\tilde{\C}\V$) dominates the latency. At $m=64$, inversion accounts for approximately $50\%$ of
the total latency. This identifies inversion as the main computational bottleneck at larger $m$. 

\noindent \textbf{Effect of Batch Size and Resolution.}
Figure~\ref{fig:batch_speedup} reports end-to-end
inference speedup over the FlashAttention-2 
across batch sizes. PixArt-$\Sigma$ benefits
more at higher resolution: in $2048^2$,
the speedup increases from $1.30\times$ at batch size
$1$ to $1.40\times$ at batch size $8$, whereas the speedup at 
$1024^2$ remains close to $1.1\times$ in batch sizes. DiT-XL/2 incurs overhead in small batches and provides
modest gains from batch size four.  These results
show the benefit of LoSpAttn with both the sequence length and
batch size.

\begin{table}[t]
    \centering
    \small
    \setlength{\tabcolsep}{3pt}
    \renewcommand{\arraystretch}{1.1}
\caption{Head-allocation ablation on DiT-XL/2. LR, SP, and SM denote low-rank, sparse, and softmax attention. Timings are reported at a batch-size of 8 on RTX 4090 GPU.}
    \label{tab:dit_head_allocation}

    \begin{tabular*}{\linewidth}
        {@{\extracolsep{\fill}}lrrr@{}}
        \toprule
        \makecell{Head allocation (\%)\\(LR, SP, SM)} &
        $\mathrm{FID}_{\mathrm{ref}}$ $\downarrow$ &
        IS $\uparrow$ &
        \makecell{Latency\\(ms) $\downarrow$} \\
        \midrule
        Exact $(0,0,100)$ 
        & -- & 125.63 & 1207.22 \\
        $(0,100,0)$
        & 113.617 & 15.30 & 1234.72 \\
        $(100,0,0)$
        & 224.000 & 3.37 & 1132.14 \\

        \midrule
        $(26.1,21.9,52.0)$
        & 5.630 & 123.07 & 1201.67 \\
        $(35.2,36.2,28.6)$
        & 6.720 & 123.59 & 1196.47 \\
        $(44.2,44.7,11.1)$
        & 8.470 & 125.99 & 1190.62 \\
        $(44.2,52.9,2.9)$
        & 10.003 & 116.81 & 1184.47 \\

        \bottomrule
    \end{tabular*}
\end{table}

\section{Conclusion}
LoSpAttn shows that selectively exploiting low-rank structure
and sparsity results in accurate, accelerated drop-in variants
of large pretrained vision and diffusion transformers.
Our fast, general calibration strategy selects low-rank,
sparse, or exact attention for each head. The low-rank branch uses our efficient, preconditioned Nystr\"om approximation with theoretical error guarantees for normalized attention.  Current limitations include a one-time calibration cost and limited acceleration at short sequence lengths. Future work is to develop an efficient fused kernel for hybrid attention to further improve inference speed.











{
    \small
    \bibliographystyle{ieeenat_fullname}
    \bibliography{main}
}

\clearpage
\appendix
\section*{Appendix}

In the appendix, we show the proof of Theorem $1$, description of structure-guided head selection for DiTs, empirical results supporting our theorem, further descriptions of our proposed method, and more visual comparison results for image generation.

\section{Proof of Theorem 1}

\noindent Define the exponential kernel
\[
    \kappa(\mathbf x,\mathbf y)
    =
    \exp(\mathbf x^\top\mathbf y/\sqrt d).
\]
Let $i^{\text{th}}$ row of landmark matrix $\M$ be denoted as $\m_i$. 
Then the normalized exponential-kernel
distributions over the landmarks and keys has entries
\[
p_M(\mathbf x)
=    \frac{
        \kappa(\mathbf x,\mathbf m_j)
    }{
        \sum_{\ell=1}^{2m}
        \kappa(\mathbf x,\mathbf m_\ell)
    }
\quad
p_K(\mathbf x)
= \frac{
        \kappa(\mathbf x,\mathbf k_j)
    }{
        \sum_{\ell=1}^{N}
        \kappa(\mathbf x,\mathbf k_\ell)
    }.
\]
Since $\tilde{\B}$ and $\tilde{\C}$ are
row-stochastic, we have
\[
\tilde{\B}^{-1}\tilde{\C}\mathbf1_N
=\mathbf1_{2m}, 
\]
where $\mathbf1_n$ denotes all ones vector of dimension $n$.
Consequently,
\begin{equation}
    \widehat{\S}_{i,:}
    =p_M(\q_i)\tilde{\B}^{-1}\tilde{\C},
    \qquad
    \S_{i,:}=p_K(\q_i).
    \label{eq:proof_normalized}
\end{equation}
For any landmark $\mathbf m_\ell$,
$p_M(\mathbf m_\ell)=\tilde{\B}_{\ell,:}$
and $p_K(\mathbf m_\ell)=\tilde{\C}_{\ell,:}$.
Thus,
\begin{equation}
\begin{aligned}
    p_M(\mathbf m_\ell)
    \tilde{\B}^{-1}\tilde{\C}
    &=
    \tilde{\B}_{\ell,:}
    \tilde{\B}^{-1}\tilde{\C} =
    \mathbf e_\ell^\top\tilde{\C}\\
    &=
    \tilde{\C}_{\ell,:}
    =
    p_K(\mathbf m_\ell),
\end{aligned}
\label{eq:proof_interpolation}
\end{equation}
where $\mathbf e_\ell$ is the $\ell$-th standard
basis vector. Thus, the normalized Nystr\"om
approximation exactly recovers the attention
distribution at every landmark. For each query $\q_i$, choose a nearest
landmark $\mathbf m_\ell$ satisfying
$\|\q_i-\mathbf m_\ell\|_2\le h_m$.
Combining \eqref{eq:proof_normalized}
and \eqref{eq:proof_interpolation},
\begin{align*}
&\|\widehat{\S}_{i,:}-\S_{i,:}\|_1 \\
&\le
\|[p_M(\q_i)-p_M(\mathbf m_\ell)]
\tilde{\B}^{-1}\tilde{\C}\|_1
 + 
\|p_K(\mathbf m_\ell)-p_K(\q_i)\|_1\\
&\le
\|p_M(\q_i)-p_M(\mathbf m_\ell)\|_1 
\|\tilde{\B}^{-1}\tilde{\C}\|_\infty
 + 
\|p_K(\mathbf m_\ell)-p_K(\q_i)\|_1\\
&\le
M_0\|p_M(\q_i)-p_M(\mathbf m_\ell)\|_1
 + 
\|p_K(\mathbf m_\ell)-p_K(\q_i)\|_1.
\end{align*}
Since softmax $\sigma$ is $1$-Lipschitz
from $\ell_\infty$ to $\ell_1$
\cite{gao2026fame}, 
\begin{align}
    \|p_K(\x)-p_K(\y)\|_1
    &\le
    \left\|
    \frac{\K(\x-\y)}{\sqrt d}
    \right\|_\infty \nonumber\\
    &=
    \frac{1}{\sqrt d}
    \max_j |\boldsymbol{k}_j^\top(\x-\y)| \nonumber\\
    &\le
    \frac{1}{\sqrt d}
    \max_j \|\boldsymbol{k}_j\|_2\|\x-\y\|_2 \nonumber\\
    &\le
    \frac{R_0}{\sqrt d}\|\x-\y\|_2,
    \label{eq:softmax_lipschitz}
\end{align}
where the last two inequalities follow from
Cauchy--Schwarz and $\|\boldsymbol{k}_j\|_2\le R_0$.
Similarly, since $\|\m_\ell\|_2\le R_0$
for all landmarks, we obtain
\begin{equation}
    \|p_M(\x)-p_M(\y)\|_1
    \le
    \frac{R_0}{\sqrt d}\|\x-\y\|_2.
\end{equation}

Therefore,
\[
\|\widehat{\S}_{i,:}-\S_{i,:}\|_1
\le
\frac{R_0}{\sqrt d}(1+M_0)h_m.
\]
Taking the maximum over $i$ proves
\eqref{eq:attention_error_bound}.

When $m=N$ and $\M=[\Q;\K]$, let
$\mathbf E=[\I_N\ \mathbf 0]$ denote the
matrix selecting the first $N$ rows.
Then $\A=\mathbf E\B$ and $\mathbf E\C=\G$.
Consequently,
\begin{align}
    \widehat{\G}
    &=\A\B^{-1}\D_B\D_C^{-1}\C \nonumber\\
    &=\mathbf E\D_B\D_C^{-1}\C
    =\D_B^{(Q)}(\D_C^{(Q)})^{-1}\G,
\end{align}
where $\D_B^{(Q)}$ and $\D_C^{(Q)}$
are the leading $N\times N$ diagonal blocks.
Thus,
\[
    \widehat G_{ij}
    =\frac{(\D_B)_{ii}}{(\D_C)_{ii}}G_{ij}.
\]
Since the scaling is positive and
constant within each row,
\[
\widehat S_{ij}
=
\frac{
(\D_B)_{ii}G_{ij}/(\D_C)_{ii}
}{
(\D_B)_{ii}(\G\mathbf1_N)_i/(\D_C)_{ii}
}
=S_{ij}.
\]
Thus $\widehat{\S}=\S$, completing
the proof. \hfill$\square$

\section{Algorithm for head selection for DiT}
\label{sec:dit_head_selection}

Algorithm~\ref{alg:dit_rank_selection} calibrates attention
choices over $HT$ head--timestep pairs, where $H$ counts
attention heads across all layers and $T$ denotes sampling
steps indexed in denoising order. The average relative
errors of low-rank and local-window approximations are computed using \eqref{eq:head_approx_error}.
The target $P$ specifies the minimum percentage of
head--timestep pairs assigned to low-rank attention.
For each landmark budget and tolerance, we identify the
first step at which each head's low-rank error exceeds the tolerance. For each head, the sparse phase begins at the earliest
step after its low-rank phase from which the sparse
error remains within $\epsilon_{\mathrm{sp}}$ at every
remaining step. The steps in between retain exact softmax
attention. The resulting schedule therefore
consists of low-rank, exact, and sparse phases,
any of which may be empty, and is reused for
subsequent inputs without repeating calibration.
\begin{algorithm}[!htp]
\caption{Structure-Guided Head Selection for DiTs}
\label{alg:dit_rank_selection}
\begin{algorithmic}[1]
\REQUIRE Pretrained DiT $f_\theta$ with $H$ heads
and $T$ sampling steps,
calibration set $\mathcal D$,
required low-rank percentage $P$,
candidate landmark budgets $\mathcal M$,
low-rank tolerance grid $\mathcal G_{\epsilon}$,
sparse tolerance $\epsilon_{\mathrm{sp}}$,
window radius $b$
\ENSURE Low-rank and sparse head--timestep sets
$\mathcal H_{\mathrm{LR}},\mathcal H_{\mathrm{SP}}$,
switching steps
$\{(\tau_h^{\mathrm{lr}},\tau_h^{\mathrm{sp}})\}_{h=1}^{H}$,
landmark budget $m$

\STATE $K\leftarrow\lceil PHT/100\rceil$,
$\mathrm{found}\leftarrow\mathrm{false}$

\STATE Run calibration denoising trajectories on $\mathcal D$;
obtain $\Q_{h,t},\K_{h,t}$ and exact attention $\S_{h,t}$
for each head, step, and sample

\STATE Compute local-window approximation errors
$\mathcal E_{h,t,\mathrm{sp}}$ with radius $b$
using \eqref{eq:head_approx_error}

\FOR{$m\in\mathcal M$ in increasing order}
    \STATE Compute low-rank approximation errors
    $\mathcal E_{h,t,\mathrm{lr}}(m)$
    using \eqref{eq:head_approx_error}

    \FOR{$\epsilon_{\mathrm{lr}}\in\mathcal G_{\epsilon}$
    in increasing order}
        \STATE Compute $\tau_h^{\mathrm{lr}}$ for all heads
        using \eqref{eq:timestep_threshold},
        with $T+1$ for an empty defining set

        \IF{$\sum_{h=1}^{H}
        (\tau_h^{\mathrm{lr}}-1)\ge K$}
            \STATE $\mathrm{found}\leftarrow\mathrm{true}$
            \STATE \textbf{break}
        \ENDIF
    \ENDFOR

    \IF{$\mathrm{found}$}
        \STATE \textbf{break}
    \ENDIF
\ENDFOR

\IF{$\mathrm{found}=\mathrm{false}$}
    \STATE $\tau_h^{\mathrm{lr}}\leftarrow1$
    for all heads; $m\leftarrow\varnothing$
\ENDIF

\STATE Compute $\tau_h^{\mathrm{sp}}$ for all heads
using \eqref{eq:sparse_timestep_threshold},
with $T+1$ for an empty defining set

\STATE $\mathcal H_{\mathrm{LR}}\leftarrow
\{(h,t):1\le h\le H,\;1\le t<\tau_h^{\mathrm{lr}}\}$

\STATE $\mathcal H_{\mathrm{SP}}\leftarrow
\{(h,t):1\le h\le H,\;
\tau_h^{\mathrm{sp}}\le t\le T\}$

\STATE Retain exact softmax attention for all remaining pairs

\STATE \textbf{return}
$\mathcal H_{\mathrm{LR}},\mathcal H_{\mathrm{SP}},
\{(\tau_h^{\mathrm{lr}},\tau_h^{\mathrm{sp}})\}_{h=1}^{H},m$
\end{algorithmic}
\end{algorithm} 






\section{More Quantitative Ablations and Results}

\noindent\textbf{Comparison with DiTFastAttn.}
We extend the qualitative comparison in
Fig.~\ref{fig:pixart_qualitative} with a quantitative
evaluation of LoSpAttn and DiTFastAttn.
For a comparison within the same inference pipeline,
we integrate LoSpAttn into Hugging Face Diffusers.
All experiments use a single NVIDIA RTX PRO 6000 GPU.
LoSpAttn calibration uses 15 images per resolution,
50 denoising steps, and landmark budgets
$\mathcal M=\{32,64,128\}$, with local-window sizes
of 512 and 2048 tokens at 1K and 2K, respectively.

Table~\ref{tab:pixart_results_extended} shows that both
methods retain FID and IS values close to the exact
baseline. DiTFastAttn achieves $1.09\times$ and
$1.34\times$ end-to-end speedups at 1K and 2K,
compared with $1.06\times$ and $1.30\times$ for
LoSpAttn. LoSpAttn uses $15.7\%$ and $27.3\%$
less peak GPU memory than DiTFastAttn, respectively.

LoSpAttn calibration takes $9$\,min $40$\,s at 1K
and $41$\,min $54$\,s at 2K. The corresponding
DiTFastAttn calibration times are $1$\,h $43$\,min
and $4$\,h $6$\,min, respectively.
\begin{table}[!htbp]
    \centering
    \scriptsize
    \setlength{\tabcolsep}{2pt}
    \renewcommand{\arraystretch}{1.0}
    \caption{PixArt-$\Sigma$ generation at 1K and 2K,
    evaluated using 10k and 5k generated images, respectively.
    DiTFastAttention replacement uses D3 compression
    $(\delta=0.075)$. FID uses real MS-COCO reference images.
    End-to-end time and peak GPU memory are measured at batch size 4.}
    \label{tab:pixart_results_extended}

    \resizebox{\columnwidth}{!}{%
    \begin{tabular}{lcccc}
        \toprule
        Method & FID $\downarrow$ & IS $\uparrow$ &
        \makecell{Inference Time\\(sec) $\downarrow$} &
        \makecell{Peak Memory\\(GB) $\downarrow$} \\
        \midrule

        \multicolumn{5}{l}{\textit{PixArt-$\Sigma$, 1K ($N=4096$)}} \\
        \midrule
        Original softmax & 56.93 & 23.43 $\pm$ 0.82& 17.70 & 22.07 \\
        DiTFastAttn(D3) & 56.84 & 23.24 $\pm$ 1.02& 16.27 & 24.12 \\
        LoSpAttn(29\%, 44\%) & 57.46 & 23.53 $\pm$ 1.04& 16.71 & 20.33 \\

        \midrule
        \multicolumn{5}{l}{\textit{PixArt-$\Sigma$, 2K ($N=16384$)}} \\
        \midrule
        Original softmax & 59.18 & $21.52 \pm 0.94$ & 129.11 & 48.02 \\
        DiTFastAttn(D3) & 58.93 & $21.63 \pm 0.91$ & 96.37 & 55.55 \\
        LoSpAttn(18\%, 46\%) & 59.56 & $21.59 \pm 0.89$ & 99.10 & 40.36 \\
        \bottomrule
    \end{tabular}%
    }
\end{table}



\noindent\textbf{Head Selection and Timestep Adaptation.}
Table~\ref{tab:dit_head_allocation_extension} extends
Table~\ref{tab:dit_head_allocation} with low-rank-only,
sparse-only, static, and random assignments.
Low-rank-only and sparse-only configurations replace
$44.2\%$ and $44.7\%$ of head--timestep pairs,
achieving $\mathrm{FID}_{\mathrm{ref}}$ scores of
$7.80$ and $7.799$, respectively.

Static calibration assigns one attention branch to
each head and reuses it across all denoising steps.
Compared with static assignments, timestep-dependent
selection reduces $\mathrm{FID}_{\mathrm{ref}}$
from $14.33$ to $7.80$ for low-rank replacement,
with a smaller change from $7.91$ to $7.799$ for
sparse replacement.
Random assignment at the same $(44.2,44.7,11.1)$
allocation substantially degrades generation,
yielding $\mathrm{FID}_{\mathrm{ref}}=204.28$
and IS $3.34$, compared with $8.47$ and $125.99$
for calibrated selection.

\begin{table}[!htbp]
    \centering
    \small
    \setlength{\tabcolsep}{3pt}
    \renewcommand{\arraystretch}{1.1}
\caption{Extension of \ref{tab:dit_head_allocation}, Head-allocation ablation on DiT-XL/2. LR, SP, and SM denote low-rank, sparse, and softmax attention. FID calculated with 5000 images.}
    \label{tab:dit_head_allocation_extension}

    \begin{tabular*}{\linewidth}
        {@{\extracolsep{\fill}}lrrr@{}}
        \toprule
        \makecell{Head allocation (\%)\\(LR, SP, SM)} &
        $\mathrm{FID}_{\mathrm{ref}}$ $\downarrow$ &
        IS $\uparrow$ \\
        \midrule
        Exact $(0,0,100)$ 
        & -- & 125.63 $\pm$ 6.1 \\
        $(0,100,0)$
        & 113.617 & 15.30 $\pm$ 0.6 \\
        $(100,0,0)$
        & 224.000 & 3.37 $\pm$ 0.1 \\

        \midrule
        $(26.1,21.9,52.0)$
        & 5.630 & 123.07 $\pm$ 4.1 \\
        $(35.2,36.2,28.6)$
        & 6.720 & 123.59 $\pm$ 5.5 \\
        $(44.2,44.7,11.1)$
        & 8.470 & 125.99 $\pm$ 5.6 \\
        $(44.2,52.9,2.9)$
        & 10.003 & 116.81 $\pm$ 5.1 \\

        \midrule
        $(44.2, 0, 55.8)$
        & 7.80 & 129.8 $\pm$ 7.9 \\
        $(0, 44.7, 55.3)$
        & 7.799 & 116.30 $\pm$ 5.7 \\
        Static $(44.2, 0, 55.8)$
        & 14.33 & 117.77 $\pm$ 4.4 \\
        Static $(0, 44.7, 55.3)$
        & 7.91 & 112.27 $\pm$ 5.7 \\
        Random $(44.2, 44.7, 11.1)$
        & 204.28 & 3.34 $\pm$ 0.07 \\

        \bottomrule
    \end{tabular*}
\end{table}



\begin{figure}[!htbp]
    \centering
    \captionsetup[subfigure]{
        font=small,
        justification=centering,
        skip=3pt
    }
    \begin{subfigure}[!htbp]{0.23\textwidth}
        \centering
            \includegraphics[
                width=\linewidth,
                height=0.6\linewidth
            ]{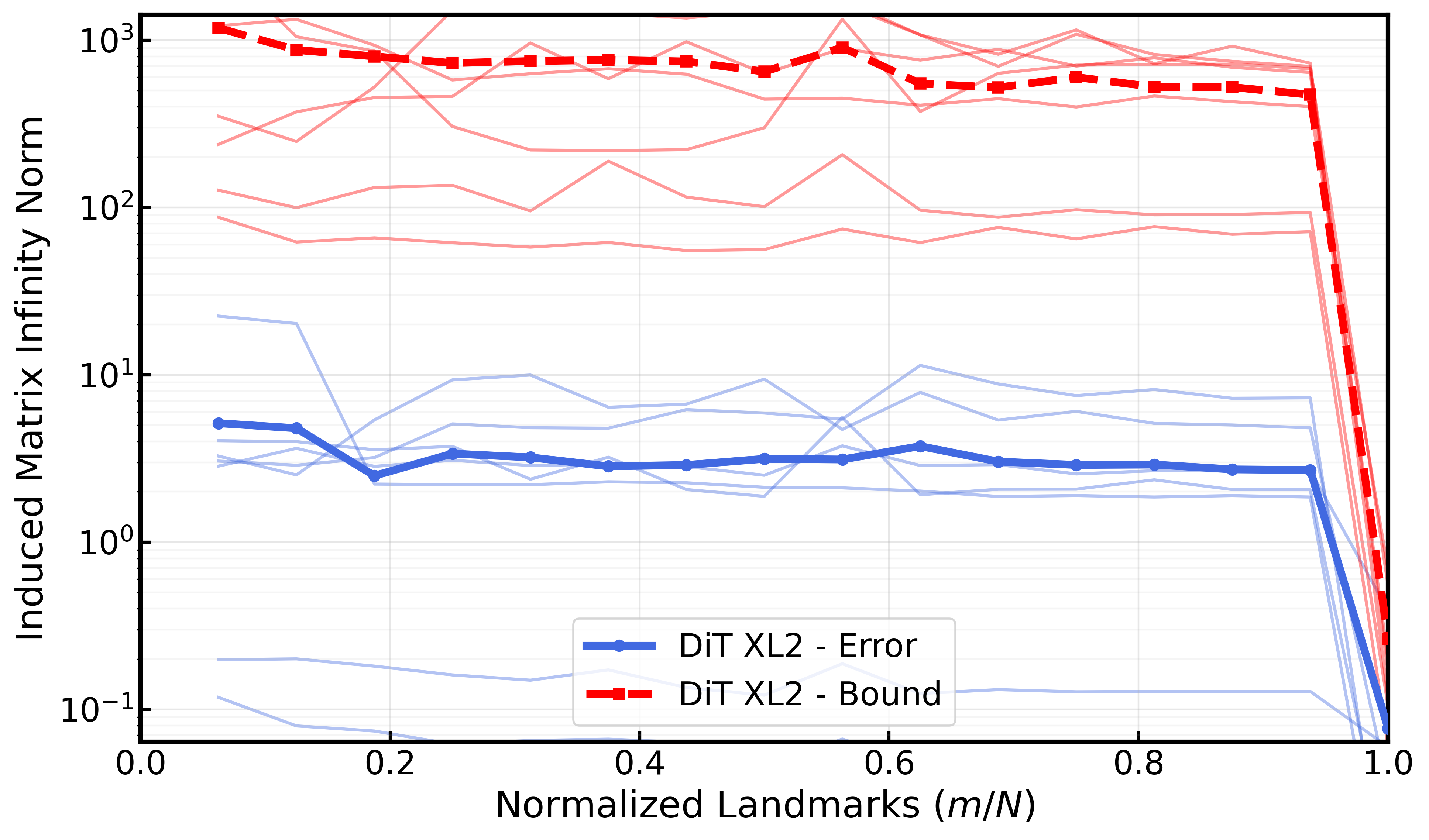}
        \caption{Error Bound plot - DiT XL2 512}
        \label{fig:error_dit_multi}
    \end{subfigure}
    \begin{subfigure}[!htbp]{0.23\textwidth}
        \centering
            \includegraphics[
                width=\linewidth,
                height=0.6\linewidth
            ]{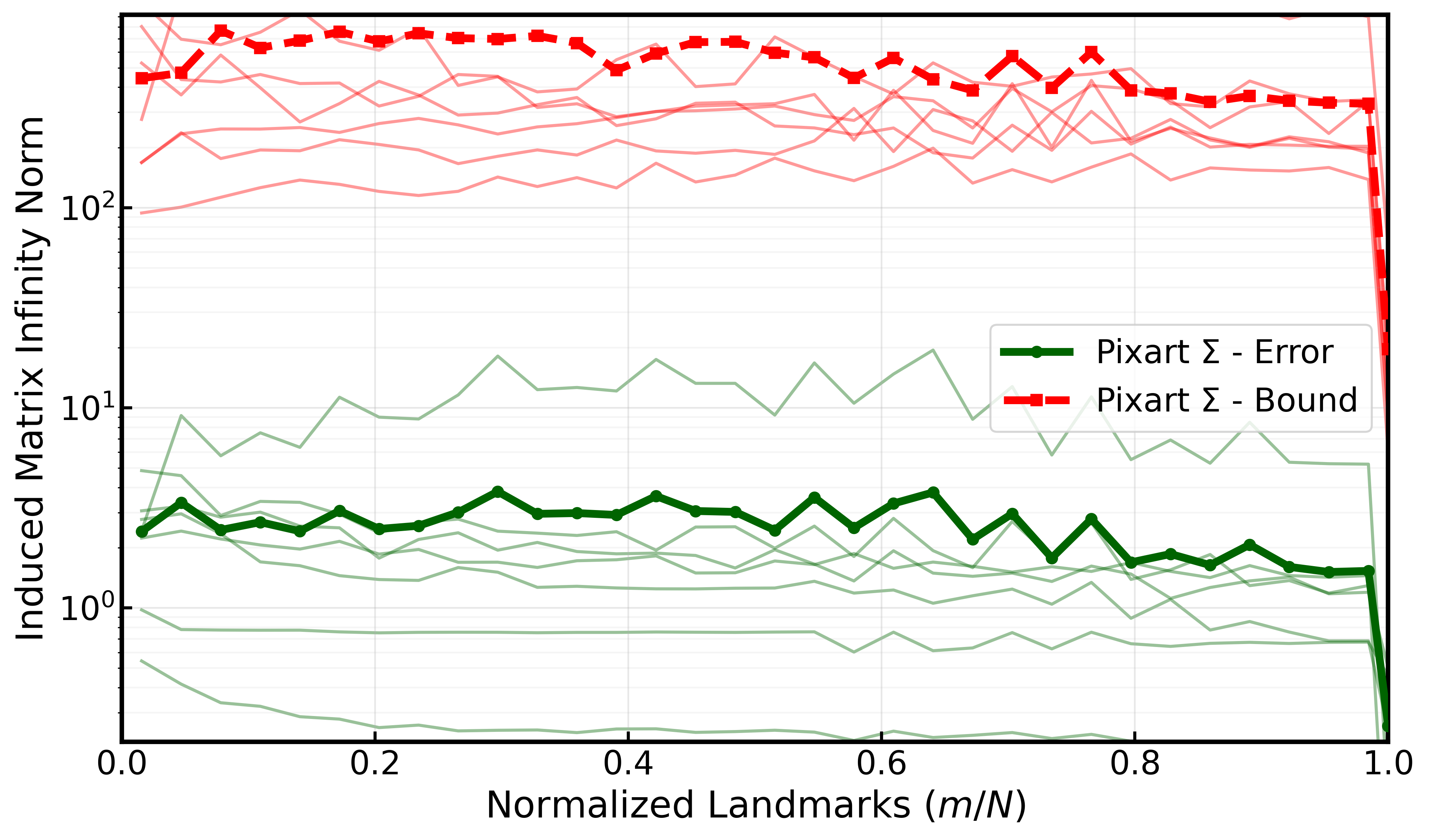}
        \caption{PixArt-$\Sigma$ 1K}
        \label{fig:error_pixart_multi}
    \end{subfigure}

\caption{\textbf{Per-head error bounds.}
Extension of Fig.~\ref{fig:error_bound} for five randomly
sampled heads per model: (a) DiT-XL/2 at $512\times512$
and (b) PixArt-$\Sigma$ at 1K.
Solid curves show approximation errors and dashed curves
show theoretical bounds; thin curves represent individual
heads and thick curves their means.
Errors use the induced matrix infinity norm on a logarithmic
scale, with normalized landmark budget $m/N$ on the
horizontal axis.}
\label{fig:error_bound_heads}
\end{figure}

\noindent\textbf{Per-Head Error Analysis.}
Figure~\ref{fig:error_bound_heads} extends the averaged
analysis in Fig.~\ref{fig:error_bound} by showing
individual-head errors and their corresponding bounds
from Theorem~\ref{thm:approximation_error}.
For both models, every sampled head's error remains
below its corresponding bound across the tested landmark
budgets. Thus, the empirical agreement holds for
individual heads as well as their averages, although
the bounds remain conservative.

\noindent\textbf{Preconditioning Ablation.}
Our formulation in \eqref{eq:newBandC} normalizes
$\B$ and $\C$ independently using
$\D_B=\operatorname{diag}(\B\mathbf1_{2m})$ and
$\D_C=\operatorname{diag}(\C\mathbf1_N)$.
As a control, we use $\D_C$ to normalize both factors.
For invertible $\B$, exact inversion gives
\[
\A(\D_C^{-1}\B)^{-1}(\D_C^{-1}\C)
=
\A\B^{-1}\C,
\]
recovering the classical Nystr\"om approximation.

Table~\ref{tab:precond_error} shows that independent
normalization yields lower or matching attention errors
and lower output errors across all reported heads.
For DiT head $(27,0)$, the attention error decreases
from $40.539$ to $0.731$, and the output error from
$14.256$ to $0.275$.
The improvements vary across heads, with the largest
gains occurring where shared normalization produces
large approximation errors.

\begin{table}[!htbp]
    \centering
    \scriptsize
    \setlength{\tabcolsep}{2pt}
    \renewcommand{\arraystretch}{1.0}
\caption{Relative attention and output errors for randomly
sampled heads using independent normalization (ours)
or shared normalization of the Nystr\"om factors.
Both variants use $m=32$ and six pseudoinverse iterations.}
    \label{tab:precond_error}
    \resizebox{\columnwidth}{!}{%
    \begin{tabular}{lcccc}
        \toprule
        (Layer, Head) & \makecell{Independent \\ normalization \\Attention error $\downarrow$} & 
        \makecell{Independent \\ normalization \\Output error $\downarrow$} &
        \makecell{Shared \\ Normalization \\Attention error $\downarrow$} &
        \makecell{Shared \\ Normalization \\output error $\downarrow$} \\
        \midrule

        \multicolumn{5}{l}{\textit{DiT XL2($N=512$)}} \\
        \midrule
        (27, 0) & 0.731 & 0.275 & 40.539 & 14.256 \\
        (24, 4) & 0.857 & 0.336 & 2.086 & 0.970 \\
        (13, 7) & 0.633 & 0.420 & 0.685 & 0.466 \\
        (1, 4) & 0.362 & 0.172 & 0.362 & 0.178 \\

        \midrule
        \multicolumn{5}{l}{\textit{PixArt-$\Sigma$, 1K ($N=4096$)}} \\
        \midrule
        (7, 8) & 0.966 & 0.656 & 0.970 & 0.668 \\
        (3, 4) & 0.829 & 0.288 & 0.844 & 0.331 \\
        (23, 1) & 0.822 & 0.312 & 0.856 & 0.379 \\
        (12, 10) & 0.693 & 0.351 & 1.753 & 1.441 \\
        \bottomrule
    \end{tabular}%
    }
\end{table}

\begin{table}[!htbp]
    \centering
    \scriptsize
    \setlength{\tabcolsep}{2pt}
    \renewcommand{\arraystretch}{1.0}
    \caption{Effect of landmark selection on relative
    attention and output errors for randomly sampled heads,
    using $m=64$ and eight pseudoinverse iterations.
    Bold rows report landmark-selection time in milliseconds.}
    \label{tab:pooling}

    \resizebox{\columnwidth}{!}{%
    \begin{tabular}{lcccc}
        \toprule
        & \multicolumn{2}{c}{2D average pooling}
        & \multicolumn{2}{c}{$k$-means clustering} \\
        \cmidrule(lr){2-3}
        \cmidrule(lr){4-5}
        (Layer, Head)
        & \makecell{Attention\\error $\downarrow$}
        & \makecell{Output\\error $\downarrow$}
        & \makecell{Attention\\error $\downarrow$}
        & \makecell{Output\\error $\downarrow$} \\
        \midrule

        \multicolumn{5}{l}{\textit{DiT-XL/2 ($N=512$)}} \\
        \midrule
        (2, 3)  & 0.021 & 0.064 & 0.010 & 0.023 \\
        (11, 7) & 0.385 & 0.230 & 0.299 & 0.173 \\
        (16, 9) & 0.962 & 0.537 & 0.984 & 0.482 \\
        (25, 8) & 0.988 & 0.728 & 1.026 & 0.741 \\
        \addlinespace[3pt]
        \textbf{Landmark time (ms)} $\downarrow$
        & \multicolumn{2}{c}{\textbf{0.62}}
        & \multicolumn{2}{c}{\textbf{2.29}} \\

        \midrule
        \multicolumn{5}{l}{
            \textit{PixArt-$\Sigma$, 1K ($N=4096$)}
        } \\
        \midrule
        (0, 1)   & 0.029 & 0.008 & 0.021 & 0.006 \\
        (12, 9)  & 0.603 & 0.273 & 0.391 & 0.152 \\
        (15, 15) & 0.775 & 0.295 & 0.416 & 0.173 \\
        (27, 11) & 0.964 & 0.364 & 0.894 & 0.278 \\
        \addlinespace[3pt]
        \textbf{Landmark time (ms)} $\downarrow$
        & \multicolumn{2}{c}{\textbf{0.97}}
        & \multicolumn{2}{c}{\textbf{2.35}} \\
        \bottomrule
    \end{tabular}%
    }
\end{table}

\begin{table*}[htp]
\centering
\caption{Inference efficiency of LoSpAttn (LoSP) and
DiTFastAttn (DFA) in shared Diffusers pipelines with
50 sampling steps, using FP16 on a single NVIDIA
RTX PRO 6000 Blackwell Max-Q GPU.
Original uses DFA's processor with exact attention
at every step; all methods share the \texttt{flash\_attn}
backend for exact attention.
Throughput is batch size divided by end-to-end time.
Per-step time is end-to-end time divided by 50 and
therefore includes pipeline overheads.
Peak memory covers the full pipeline; OOM denotes
out of memory.}
\label{tab:e2e_latency_dfa_losp}
\setlength{\tabcolsep}{4pt}
\resizebox{\textwidth}{!}{%
\begin{tabular}{ll ccc ccc ccc ccc}
\toprule
& & \multicolumn{3}{c}{End-to-end Time (s) $\downarrow$}
  & \multicolumn{3}{c}{Throughput (img/s) $\uparrow$}
  & \multicolumn{3}{c}{Time per Step (ms) $\downarrow$}
  & \multicolumn{3}{c}{Peak Memory (GB) $\downarrow$} \\
\cmidrule(lr){3-5} \cmidrule(lr){6-8} \cmidrule(lr){9-11} \cmidrule(lr){12-14}
Model & Batch & Original & LoSP & DFA & Original & LoSP & DFA & Original & LoSP & DFA & Original & LoSP & DFA \\
\midrule
\multirow{7}{*}{DiT-XL/2, $512^2$}
 & 1  & 0.943  & 1.570  & 0.942  & 1.060 & 0.637 & 1.061 & 18.87  & 31.40  & 18.84  & 2.31  & 2.41  & 2.65  \\
 & 4  & 2.842  & 3.093  & 2.766  & 1.407 & 1.293 & 1.446 & 56.84  & 61.85  & 55.33  & 4.09  & 3.81  & 4.78  \\
 & 8  & 5.517  & 5.986  & 5.321  & 1.450 & 1.336 & 1.503 & 110.35 & 119.72 & 106.43 & 6.58  & 5.81  & 7.75  \\
 & 16 & 11.248 & 12.218 & 10.961 & 1.423 & 1.310 & 1.460 & 224.95 & 244.35 & 219.23 & 11.57 & 9.82  & 13.68 \\
& 64 & 51.958 & 56.264 & 49.598 & 1.232 & 1.137 & 1.290 & 1039.16 & 1125.28 & 991.95 & 39.50 & 31.84 & 47.31 \\
\midrule
\multirow{7}{*}{PixArt-$\Sigma$, $1024^2$}
 & 1  & 4.409  & 4.213  & 4.031  & 0.227 & 0.237 & 0.248 & 88.18   & 84.25   & 80.61   & 14.94 & 14.68 & 15.63 \\
 & 4  & 17.695 & 16.714 & 16.269 & 0.226 & 0.239 & 0.246 & 353.90  & 334.28  & 325.38  & 22.07 & 20.33 & 24.12 \\
 & 8  & 38.860 & 36.717 & 35.644 & 0.206 & 0.218 & 0.224 & 777.20  & 734.34  & 712.89  & 32.07 & 28.36 & 35.95 \\
 & 16 & 77.785 & 74.531 & 71.631 & 0.206 & 0.215 & 0.223 & 1555.70 & 1490.62 & 1432.61 & 48.08 & 40.43 & 55.61 \\
\midrule
\multirow{4}{*}{PixArt-$\Sigma$, $2048^2$}
 & 1  & 30.863  & 23.326  & 22.674 & 0.032 & 0.043 & 0.044 & 617.25  & 466.52  & 453.48  & 23.57 & 21.82 & 25.62 \\
 & 4  & 129.113 & 99.097  & 96.368 & 0.031 & 0.040 & 0.042 & 2582.27 & 1981.93 & 1927.37 & 48.02 & 40.36 & 55.55 \\
 & 8  & 258.713 & 202.303 & OOM    & 0.031 & 0.040 & OOM   & 5174.26 & 4046.05 & OOM     & 75.95 & 60.41 & OOM   \\
\bottomrule
\end{tabular}%
}
\end{table*}

\noindent\textbf{Landmark Selection.}
We construct landmarks using adaptive 2D average pooling,
which exploits the spatial layout of image tokens.
The dependence of our error bound
\eqref{eq:attention_error_bound} on landmark coverage
$h_m$ motivates comparison with $k$-means clustering.
Table~\ref{tab:pooling} shows that $k$-means reduces
attention and output errors for most sampled heads.
However, it increases landmark-selection time from
$0.62$ to $2.29$\,ms on DiT-XL/2 and from
$0.97$ to $2.35$\,ms on PixArt-$\Sigma$ at 1K.
We therefore use pooling to limit landmark-construction
overhead.

\noindent\textbf{End-to-End Efficiency in Diffusers.}
Table~\ref{tab:e2e_latency_dfa_losp} evaluates efficiency
across models, resolutions, and batch sizes.
Relative to the matched exact-attention baseline,
LoSpAttn achieves $1.04$--$1.06\times$ speedups on
PixArt-$\Sigma$ at 1K and $1.28$--$1.32\times$ at 2K.
DiTFastAttn has lower latency at all mutually feasible
configurations, while LoSpAttn consistently uses less
peak memory.

At 2K and batch size eight, DiTFastAttn runs out of
memory, whereas LoSpAttn completes generation using
$60.41$\,GB, compared with $75.95$\,GB for exact
attention, and reduces runtime from $258.71$ to
$202.30$\,s.
On DiT-XL/2, LoSpAttn remains slower than exact
attention across all tested batch sizes, although
it reduces peak memory from $39.50$ to $31.84$\,GB
at batch size 64.

\begin{table}[!htbp]
    \centering
\caption{Effect of pseudoinverse iterations on the symmetric
Nystr\"om approximation for PixArt-$\Sigma$ at 1K,
with $m=64$. Errors are averaged over the lowest-error
50\% of heads. Runtime covers pseudoinverse computation only.}
    \label{tab:moore}
    \begin{tabular}{lccc}
        \toprule
        Iters & Attention error $\downarrow$ & Output error $\downarrow$ & time (ms) \\
        \midrule
        2  & 0.618371 & 0.372898 & 0.1619 \\
        4  & 0.554779 & 0.300066 & 0.2674 \\
        6  & 0.527682 & 0.273288 & 0.3739 \\
        8  & 0.520001 & 0.263001 & 0.4775 \\
        10 & 0.529456 & 0.262264 & 0.5889 \\
        12 & 0.558936 & 0.266088 & 0.6888 \\
        \bottomrule
    \end{tabular}
\end{table}

\noindent\textbf{Pseudoinverse Iterations.}
Table~\ref{tab:moore} examines the accuracy--runtime
trade-off as the iteration count increases.
From two to eight iterations, attention error decreases
from $0.618$ to $0.520$ and output error from $0.373$
to $0.263$, while inversion time increases from
$0.162$ to $0.478$\,ms.
Eight iterations achieve the lowest attention error;
ten yield only a marginal further reduction in output
error ($0.262$). At twelve iterations, both errors
increase, indicating that additional computation does
not consistently improve approximation accuracy.

\begin{table}[!htbp]
    \centering
\caption{T2T-ViT classification accuracy (\%) on the
50k-image ImageNet-1K validation set.
GFLOPs are reported separately for the first and
second attention layers in the T2T module (L1/L2).}
    \label{tab:t2tvit}
    \resizebox{\columnwidth}{!}{%
        \begin{tabular}{lccc}
            \toprule
            Method & Top-1 $\uparrow$ & Top-5 $\uparrow$ & GFLOPS (L1/L2) \\
            \midrule
            Original & 82.55  & 96.06 \ & 161.12 / 10.07 \\
            SpargeAttention & 81.73 & 95.85 & 161.14 / 10.09 \\
            Scatterbrain & 82.12 & 95.93 & 5.49 / 0.34 \\
            KDEformer & 82.07 & 95.86 & 5.05 / 0.30 \\
            WildCat & 82.27 & 95.94 & 11.65 / 2.55 \\
            Thinformer & 82.20 & 95.87 & 9.79 / 3.28 \\
            MonarchAttention & 82.38 & 96.03  & 18.70 / 2.33 \\
            LoSpAttn & 81.24 & 95.56 & 19.96 / 3.09 \\
            \bottomrule
        \end{tabular}
    }
\end{table}

\noindent\textbf{T2T-ViT Classification.}
Similar to the BigGAN evaluation in the main paper,
we compare LoSpAttn with existing drop-in approximations
on T2T-ViT, a setting where these methods perform well.
We replace the two attention layers in the T2T module
while leaving the ViT backbone unchanged.
Table~\ref{tab:t2tvit} shows that LoSpAttn reduces
GFLOPs from $161.12/10.07$ to $19.96/3.09$, with top-1 accuracy decreasing
from $82.55\%$ to $81.24\%$.
The competing approximations retain higher classification
accuracy in this setting.

\section{Additional qualitative comparisons}

\begin{figure*}
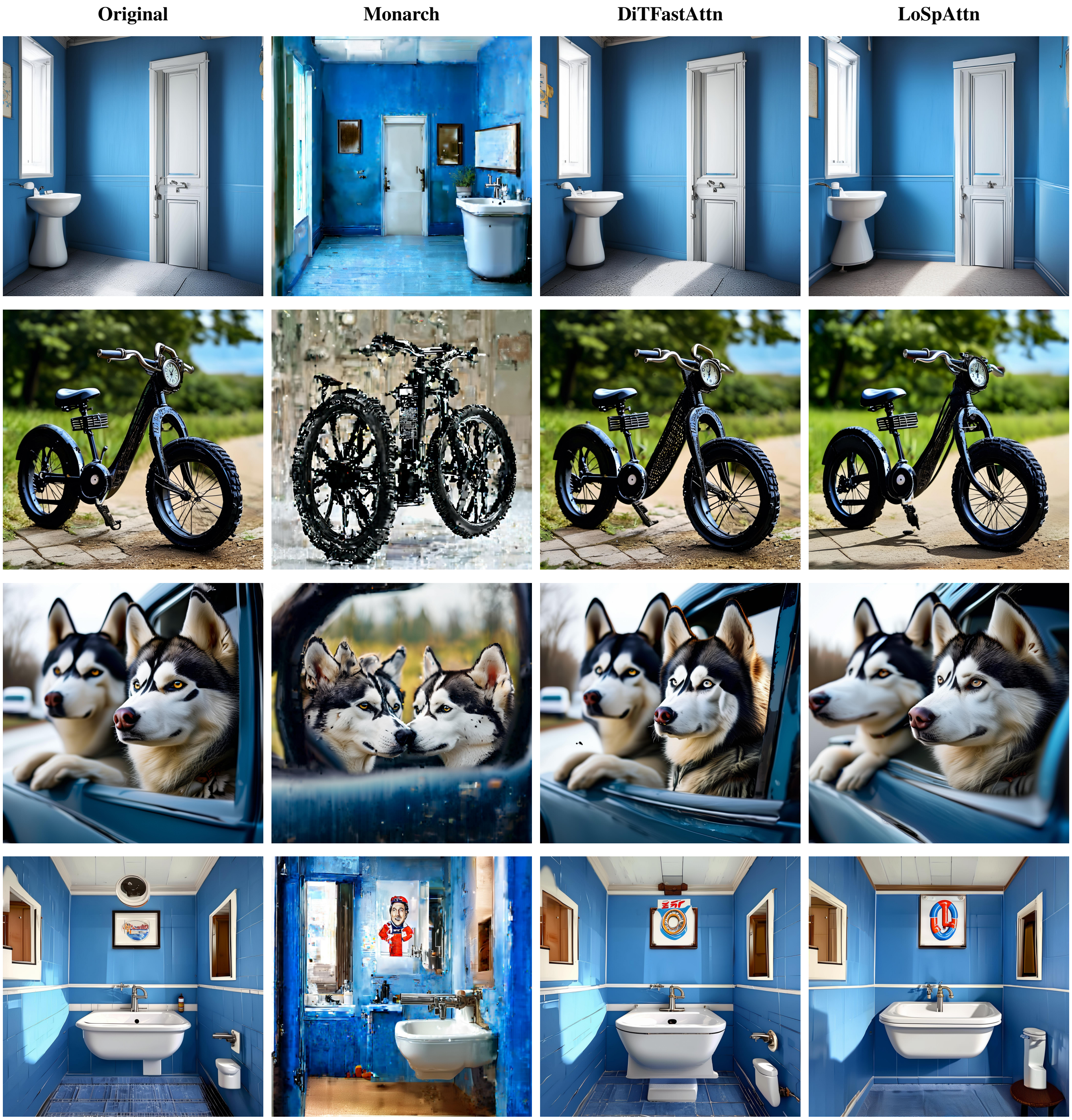

    \suppqualsetup

    \begin{tabular}{@{}cccc@{}}
        \suppqualheader
        \suppqualrow{pixart2k}{000000000049}
        \suppqualrow{pixart2k}{000000000478}
        \suppqualrow{pixart2k}{000000001113}
        \suppqualrow{pixart2k}{000000000121}
    \end{tabular}

    \caption{Qualitative comparisons on PixArt-$\Sigma$
    at 2K resolution (examples 1--4).
    Columns show original softmax, MonarchAttention,
DiTFastAttn, and LoSpAttn.
Prompts, from top to bottom:
(1) A room with blue walls and a white sink and door.;
(2) A black metal bicycle with a clock inside the front wheel.;
(3) Two husky's hanging out of the car windows.;
(4) This is a blue and white bathroom with a wall sink and a lifesaver on the wall..}    
    \label{fig:qualitative_pixart2k_a}
\end{figure*}

\begin{figure*}
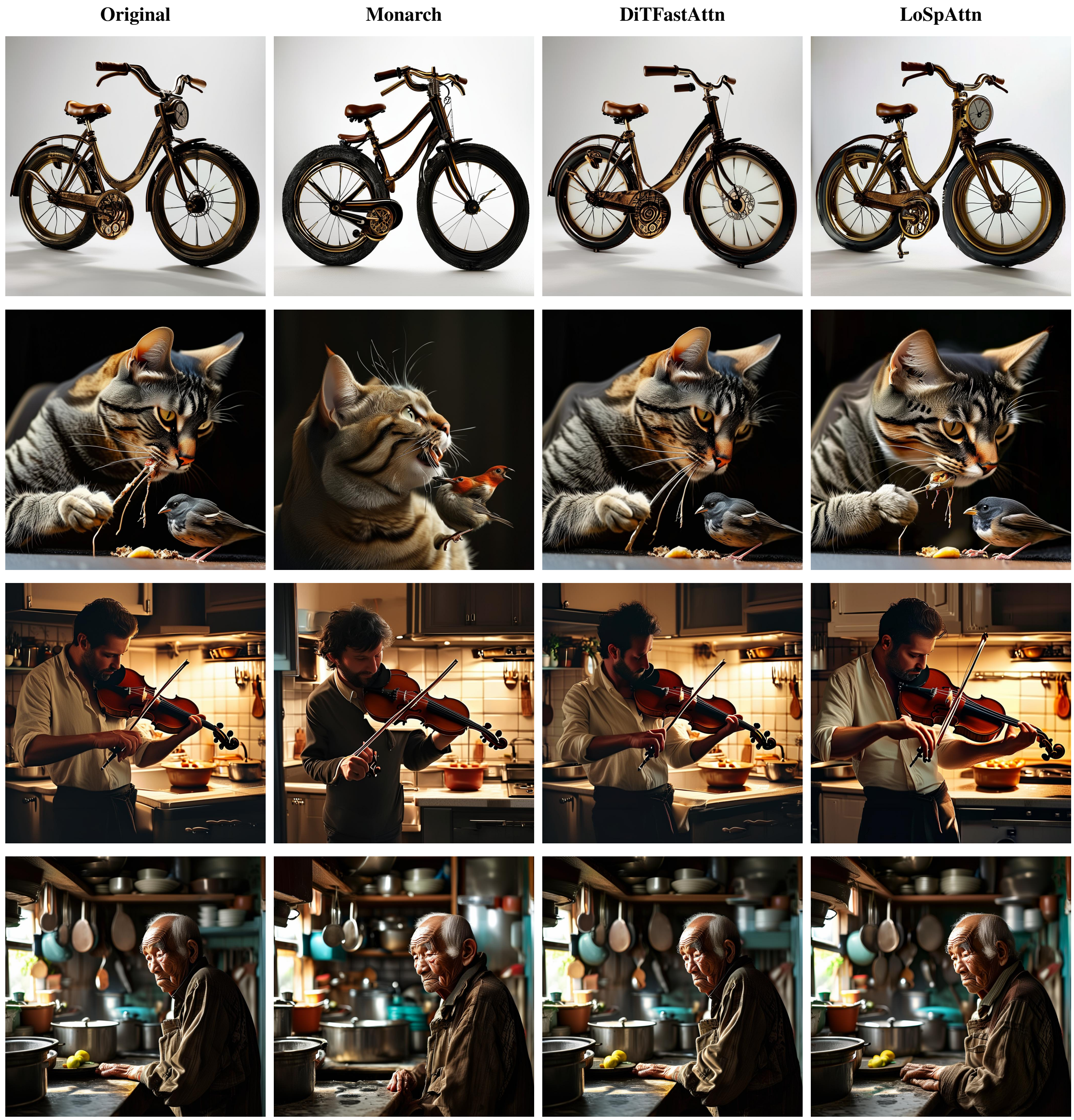

    \suppqualsetup

    \begin{tabular}{@{}cccc@{}}
        \suppqualheader
        \suppqualrow{pixart1k}{000000000037}
        \suppqualrow{pixart1k}{000000001158}
        \suppqualrow{pixart1k}{000000001436}
        \suppqualrow{pixart1k}{000000001437}
    \end{tabular}

\caption{Qualitative comparisons on PixArt-$\Sigma$
at 1K resolution (examples 1--4).
Columns show original softmax, MonarchAttention,
DiTFastAttn, and LoSpAttn.
Prompts, from top to bottom:
(1) A bicycle replica with a clock as the front wheel
(2) A cat eating a bird it has caught
(3) A picture of a man playing a violin in a kitchen
(4) A shot of an elderly man inside a kitchen.
}    
\label{fig:qualitative_pixart1k_a}
\end{figure*}

\begin{figure*}
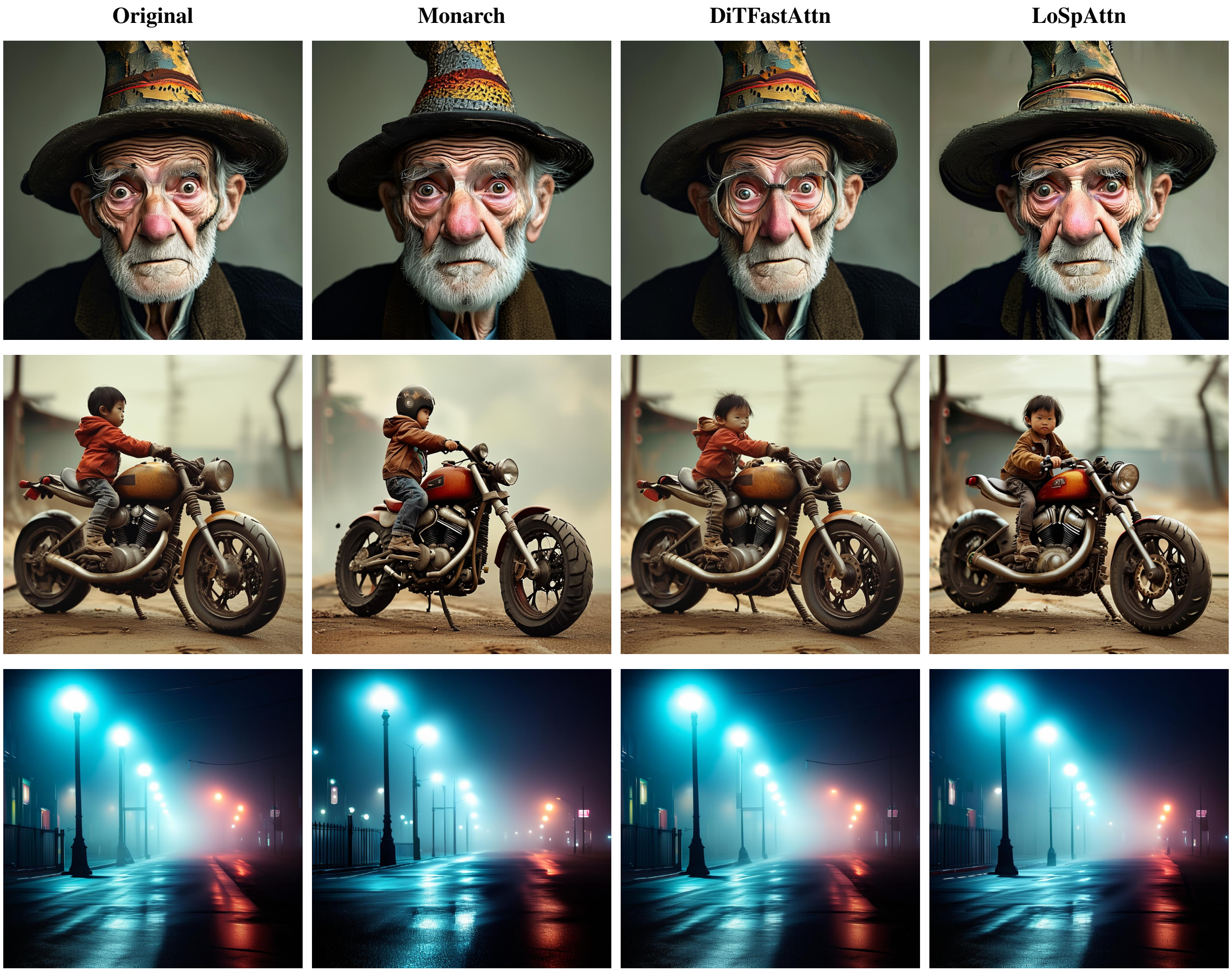

\suppqualsetup

    \begin{tabular}{@{}cccc@{}}
        \suppqualheader
        \suppqualrow{pixart1k}{000000001707}
        \suppqualrow{pixart1k}{000000002405}
        \suppqualrow{pixart1k}{000000004372}
    \end{tabular}

    \caption{Qualitative comparisons on PixArt-$\Sigma$
    at 1K resolution (examples 5--7).
Columns show original softmax, MonarchAttention,
DiTFastAttn, and LoSpAttn.
(5) An old man is wearing an odd hat
(6) A small child climbs atop a large motorcycle
(7) Empty fog covered streets in the night amongst traffic lights.
}    
    \label{fig:qualitative_pixart1k_b}
\end{figure*}

\noindent\textbf{Additional Qualitative Comparisons.}
Figures~\ref{fig:qualitative_pixart2k_a} and
Figures~\ref{fig:qualitative_pixart1k_a}--%
\ref{fig:qualitative_pixart1k_b}
present seven examples each for PixArt-$\Sigma$
at 2K and 1K, respectively.
Columns show original softmax, MonarchAttention,
DiTFastAttn, and LoSpAttn, with each row corresponding
to the same prompt.
In the 2K examples, MonarchAttention introduces
pronounced texture artifacts and structural changes
in the bathrooms and bicycle.
DiTFastAttn and LoSpAttn more closely retain the original
scene layout, subject appearance, and lighting.
At 1K, all methods produce recognizable scenes,
with differences in object proportions, poses, and
fine details. LoSpAttn retains the facial detail
of the elderly portrait and the lighting and reflections
of the night street, although subject poses vary,
particularly in the motorcycle example.




\end{document}